 \documentclass[mnsc,nonblindrev]{informs4} 

\OneAndAHalfSpacedXI

\usepackage{algorithm,algorithmic}
\usepackage{subfigure}
\usepackage{mathrsfs}
\usepackage{amsmath,bm}
\usepackage{graphicx}
\usepackage{url}
\definecolor{DarkBlue}{rgb}{0,0.08,0.45}
\usepackage[backref = false, bookmarks, colorlinks = true, plainpages = false, citecolor = DarkBlue , urlcolor = DarkBlue, filecolor = DarkBlue, linkcolor = DarkBlue]{hyperref}
\usepackage{newunicodechar}
\usepackage[utf8]{inputenc}
\usepackage{booktabs}

\usepackage{natbib}
 \bibpunct[, ]{(}{)}{,}{a}{}{,}%
 \def\bibfont{\small}%
 \def\bibsep{\smallskipamount}%

 \usepackage{pgfplots}
\usepgfplotslibrary{fillbetween}
\pgfplotsset{compat=1.13}
\TheoremsNumberedThrough     
\ECRepeatTheorems

\EquationsNumberedThrough    

\begin{document}
\newcommand{\abs}[1]{\left|  #1 \right| }
\newcommand{\brak}[1]{\left(#1\right)}    
\newcommand{\crl}[1]{\left\{#1\right\}}   
\newcommand{\edg}[1]{\left[#1\right]}     
\newcommand{\norm}[1]{\|#1\|}
\newcommand{\floor}[1]{\lfloor #1 \rfloor}

\newcommand{\cA}{{\mathcal A}}
\newcommand{\cB}{{\mathcal B}}
\newcommand{\cD}{{\mathcal D}}
\newcommand{\cF}{{\mathcal F}}
\newcommand{\cG}{{\mathcal G}}
\newcommand{\cH}{{\mathcal H}}
\newcommand{\cK}{{\mathcal K}}
\newcommand{\cL}{{\mathcal L}}
\newcommand{\cM}{{\mathcal M}}
\newcommand{\cR}{{\mathcal R}}
\newcommand{\cS}{{\mathcal S}}
\newcommand{\cT}{{\mathcal T}}
\newcommand{\cX}{{\mathcal X}}
\newcommand{\cP}{{\mathcal P}}
\newcommand{\cV}{{\mathcal V}}

\newcommand{\mA}{{\mathbb A}}
\newcommand{\mV}{{\mathbb V}}
\newcommand{\mC}{{\mathbb C}}
\newcommand{\mR}{{\mathbb R}}
\newcommand{\mE}{{\mathbb E}}
\newcommand{\mw}{{\mathbb w}}
\newcommand{\mT}{{\mathbb T}}

\newcommand{\bb}{{\mathbf b}}
\newcommand{\bd}{{\mathbf d}}
\newcommand{\by}{{\mathbf y}}
\newcommand{\bp}{{\mathbf p}}
\newcommand{\bc}{{\mathbf c}}
\newcommand{\bg}{{\mathbf g}}
\newcommand{\bl}{{\mathbf l}}
\newcommand{\bbf}{{\mathbf f}}

\newcommand{\bx}{{\mathbf x}}
\newcommand{\bA}{{\mathbf A}}
\newcommand{\bB}{{\mathbf B}}
\newcommand{\bC}{{\mathbf C}}
\newcommand{\bD}{{\mathbf D}}
\newcommand{\bG}{{\mathbf G}}
\newcommand{\bL}{{\mathbf L}}
\newcommand{\bS}{{\mathbf S}}
\newcommand{\bQ}{{\mathbf Q}}
\newcommand{\bU}{{\mathbf U}}
\newcommand{\bV}{{\mathbf V}}
\newcommand{\bX}{{\mathbf X}}
\newcommand{\bZ}{{\mathbf Z}}
\newcommand{\bF}{{\mathbf F}}

\newcommand{\bmu}{{\boldsymbol \mu}}
\newcommand{\bw}{{\boldsymbol w}}
\newcommand{\balpha}{{\boldsymbol \alpha}}
\newcommand{\blambda}{{\boldsymbol \lambda}}

\newcommand{\btheta}{\boldsymbol{\theta}}
\newcommand{\bsigma}{\boldsymbol{\sigma}}
\newcommand{\bnu}{\boldsymbol{\nu}}
\newcommand{\bSigma}{\boldsymbol{\Sigma}}
\newcommand{\bgamma}{\boldsymbol{\gamma}}
\newcommand{\bs}{\boldsymbol{s}}
\newcommand{\ba}{\boldsymbol{a}}

\newcommand{\F}{\mathbb{F}}
\newcommand{\p}{\mathbb{P}}
\newcommand{\Q}{\mathbb{Q}}

\newcommand{\R}{\mathbb{R}}
\newcommand{\q}{\mathbb{Q}}

\newcommand{\tr}{{\rm tr}}

\newcommand{\id}{{\mathbbm 1}}

\newcommand{\expect}{\mathbb{E}}



\RUNAUTHOR{Jiang, Han, Peng, and Zhang}

\RUNTITLE{Closing the Loop: Coordinating Inventory and Recommendation via DRL}

\TITLE{Closing the Loop: Coordinating Inventory and Recommendation via Deep Reinforcement Learning on Multiple Timescales}

\ARTICLEAUTHORS{%
\AUTHOR{Jinyang Jiang, Jinhui Han, Yijie Peng, and Ying Zhang}
\AFF{Guanghua School of Management, Peking University, Beijing 100871, CHINA\\
\EMAIL{jinyang.jiang@stu.pku.edu.cn}, \EMAIL{jinhui.han@gsm.pku.edu.cn}, 
\EMAIL{pengyijie@pku.edu.cn},
\EMAIL{zhang@gsm.pku.edu.cn}}
} 

\ABSTRACT{Effective cross-functional coordination is essential for enhancing firm-wide profitability, particularly in the face of growing organizational complexity and scale. Recent advances in artificial intelligence, especially in reinforcement learning (RL), offer promising avenues to address this fundamental challenge.  This paper proposes a unified multi-agent RL framework tailored for joint optimization across distinct functional modules, exemplified via coordinating inventory replenishment and personalized product recommendation. We first develop an integrated theoretical model to capture the intricate interplay between these functions and derive analytical benchmarks that characterize optimal coordination. The analysis reveals synchronized adjustment patterns across products and over time, highlighting the importance of coordinated decision-making. Leveraging these insights, we design a novel multi-timescale multi-agent RL architecture that decomposes policy components according to departmental functions and assigns distinct learning speeds based on task complexity and responsiveness.
Our model-free multi-agent design improves scalability and deployment flexibility, while multi-timescale updates enhance convergence stability and adaptability across heterogeneous decisions.
We further establish the asymptotic convergence of the proposed algorithm. Extensive simulation experiments demonstrate that the proposed approach significantly improves profitability relative to siloed decision-making frameworks, while the behaviors of the trained RL agents align closely with the managerial insights from our theoretical model. 
Taken together, this work provides a scalable, interpretable RL-based solution to enable effective cross-functional coordination in complex business settings.
}%


\maketitle

\section{Introduction}

Coordinating decision-making across functional units remains one of the fundamental challenges in complex organizations. In modern enterprises, responsibilities are distributed across specialized departments, such as operations, marketing, finance, and sales, each pursuing its own objectives under distinct information sets, incentives, and constraints. While this specialization facilitates managerial focus and domain expertise, it often leads to suboptimal system-wide outcomes when the interdependencies among functions are not properly managed. Substantial value can be unlocked through integrated cross-functional decision-making; yet, the dynamic, nonlinear, and uncertain nature of these interactions makes such coordination difficult to achieve using traditional analytical or rule-based methods. Even in large technology-driven firms and digital platforms such as Amazon and JD.com, where the scale and complexity of decision-making are particularly pronounced, fine-grained, real-time coordination across functions remains elusive. Recent advances in artificial intelligence (AI), particularly reinforcement learning (RL), offer new opportunities to revisit this longstanding challenge by enabling adaptive policy learning in complex, high-dimensional environments with interacting agents and sequential decisions. Building on these developments, this paper proposes a new RL framework to facilitate effective cross-functional coordination, with a specific application to the integration of inventory and marketing decisions.

To align with realistic organizational structures while preserving scalability, we adopt a \emph{heterogeneous multi-agent RL architecture} built on direct policy optimization. In this framework, each functional unit, representing distinct departments or decision channels, operates its own structurally distinct sub-policy. These sub-policies, implemented as deep neural networks mapping system states to actions, execute independently during deployment but are trained jointly, allowing the agents to gradually learn highly coordinated and synchronized behaviors. This modular decomposition reduces the overall number of parameters to be learned, substantially improving training efficiency, stability, and scalability as organizational complexity grows. 
Distinct from conventional value-based RL approaches frequently employed in the literature, such as deep Q-learning (DQN; \citealt{mnih2015human}), our method directly optimizes policies rather than approximating value functions over exponentially expanding state-action spaces. This direct policy optimization offers stronger robustness and better scalability when faced with complex observation and decision structures. 
Another widely used RL baseline is proximal policy optimization (PPO; \citealt{schulman2017proximal}). 
However, its single-agent formulation often creates substantial parameter redundancy and limits coordination efficiency in multi-agent contexts.
Moreover, our approach is fully \emph{model-free}, bypassing the need to estimate underlying system dynamics. This not only simplifies implementation in high-dimensional settings but also improves robustness against model misspecification---an issue commonly faced by model-based RL methods in real-world applications.

To further enhance learning precision and coordination quality, we introduce a \emph{multi-timescale policy update mechanism} that assigns distinct update speeds to individual agents according to their decision complexity and learning responsiveness. In doing so, it allows simpler operational decisions to adapt rapidly, while more complex policy components undergo gradual and stable refinement. Notably, this multi-timescale update mechanism aligns with emerging perspectives from both neuroscience and AI. For example, Geoffrey Hinton, recipient of the Nobel Price in Physics and Turing Award, has emphasized that different regions of the brain likely learn at different rates, and has advocated for machine learning algorithms to reflect such biologically inspired structures.\footnote{This perspective was shared by Geoffrey Hinton during his keynote at the ACM Federated Computing Research Conference (FCRC) in 2019 and in a fireside chat with Andrew Chi-Chih Yao at the World Artificial Intelligence Conference (WAIC) in 2025.}
Incorporating it into policy updates is therefore not only a practical design choice, but also a timely methodological advancement in contemporary AI research. Compared to conventional training schedules with uniform step sizes, the multi-timescale design improves stability, convergence efficiency, and responsiveness, which are critical properties for enabling cross-functional coordination in complex operational environments.

The proposed multi-timescale multi-agent RL framework is specifically tailored to the coordination problem under consideration. To motivate its design and establish a foundation for evaluation, we first analyze simplified theoretical settings that not only highlight the necessity of this approach but also generate managerial insights to assess the plausibility of the learned RL solutions. In particular, when the behaviors exhibited by the trained RL agents align with these analytical insights, it offers an interpretable basis for understanding and validating the otherwise black-box decision processes. Specifically, we examine coordination along two orthogonal dimensions: (i) cross-product (horizontal) coordination, which explores how inventory and marketing decisions should interact across products in a static setting, and (ii) intertemporal (vertical) coordination, which characterizes how decisions should dynamically evolve over time to achieve optimal system performance. 

At the cross-product level, analytical results reveal that inventory replenishment and product recommendation should exhibit tightly synchronized adjustment patterns. When inventory levels are high, recommendation efforts should intensify to stimulate demand; conversely, when customer purchase intentions rise due to active recommendations, inventory replenishment must respond accordingly to secure sufficient supply. Furthermore, recommendation decisions should prioritize products with higher relative marketing efficiency and profitability, allocating limited promotional resources toward these items. The stronger these metrics, the more aggressive the recommendation efforts should be. At the intertemporal level, two fundamental coordination mechanisms are identified. The first is \emph{demand smoothing}, where recommendation intensity is strategically managed over time to stabilize demand in line with inventory availability, thereby improving system controllability and profitability. The second is \emph{adaptive ordering}, where inventory decisions proactively respond to shifts in customer purchasing intentions driven by evolving recommendation strategies.


Connecting these theoretical insights to algorithm design, the departmental structure naturally motivates a multi-agent architecture, wherein each agent operates its own sub-policy while jointly optimizing toward a common profit objective. It preserves coordination while reducing the dimensionality of the action space, thereby improving scalability, training efficiency, and real-world implementability. In addition, the differing stability and complexity of decision components motivate a multi-timescale stochastic approximation framework: more stable decisions (e.g., inventory control) adopt larger step sizes, while more sensitive policies (e.g., recommendation decisions) are updated more conservatively. We then assign smaller neural networks and faster updates to inventory agents, allowing rapid adjustment to recommendation policies within the multi-timescale structure. 

We evaluate the proposed multi-timescale multi-agent approach through both theoretical analysis and numerical experiments. Theoretically, we establish convergence guarantees for the multi-timescale method in a stylized one-period setting, showing that the fast-timescale component first tracks its conditional optimum given the slow component, which subsequently converges toward a stationary point of the joint objective. In the general setting, we further characterize the asymptotic behavior of both fast- and slow-timescale agents as evolving within invariant sets associated with the underlying differential inclusions. While non-convex problems may admit multiple equilibria, these results broadly capture the algorithm’s long-run dynamics. Convergence to a unique solution would require strong assumptions rarely satisfied in practical large-scale systems.

Numerically, we conduct extensive simulation studies to assess the learning efficiency, managerial validity, and performance of the trained policies. The multi-timescale design consistently outperforms single-timescale baselines, achieving faster convergence and greater training stability. Simultaneously, the multi-agent structure reduces computational overhead relative to canonical single-agent RL methods, which may fail entirely as problem complexity grows. These benefits are expected to scale favorably with the number of functional modules involved. We further validate that the learned RL policies exhibit behaviors aligned with our theoretical insights: replenishment and recommendation decisions are highly synchronized; recommendation intensity adapts systematically to products’ relative marketing efficiency and profitability; and both demand smoothing and adaptive ordering patterns emerge under exogenous perturbations. Finally, we demonstrate that coordinated decision-making delivers substantial system-level profit improvements over decentralized benchmarks, which is also expected to grow with the problem scale. Additional robustness checks in the online appendix further confirm these findings.


To summarize, the main contributions of the present paper are threefold. Methodologically, we propose a model-free multi-timescale multi-agent RL framework broadly applicable to asymmetric cross-functional coordination problems that are common and fundamental in large organizations. Using a coordinated inventory and marketing case as a representative application, we demonstrate the implementation of our framework and highlight its scalability and computational efficiency, particularly as the number of interacting decision agents increases. Theoretically, we analyze the algorithm’s asymptotic behavior and establish convergence guarantees for the multi-timescale learning mechanism using stochastic approximation theory. We examine stylized yet analytically tractable models, deriving key structural insights into optimal coordination strategies. These insights not only justify our algorithmic design but also offer important managerial intuition.
Finally, we conduct extensive numerical experiments that validate the effectiveness of the proposed approach and illustrate its clear advantages over standard RL alternatives. The observed RL agents' behaviors align closely with the theoretical insights, offering rare interpretability for what are often perceived as black-box systems. Collectively, the proposed method is shown to enable effective cross-functional coordination while preserving the modularity and operational autonomy required for practical deployment in complex organizations.

The remainder of this paper is organized as follows. Section \ref{section: Related Work} reviews the
literature on related topics. Section \ref{Theoretical Model} presents a theoretical model and derives analytical insights into the coordination of inventory and recommendation decisions, emphasizing both cross-product and intertemporal synergies that inform algorithm design. Section \ref{sec:multi_scale_RL} introduces a multi-agent RL framework with multi-timescale updates, translating these insights into a practical solution for cooperative decision-making. Section \ref{sec:Numerical Experiments} validates the proposed method through extensive numerical experiments. Section \ref{sec:conclusion} concludes the paper. All proofs and supplementary materials are provided in the online appendix.

\section{Literature Review}\label{section: Related Work}

Our work contributes to advancing business decision-making through state-of-the-art deep RL, and it relates to the following three streams of the literature.

\textbf{AI for business decision-making.} 
The adoption of AI in business decision-making has accelerated in recent years, driven by advancements in machine learning (ML) and data availability. To name a few, \cite{liu2021time} develop a ML-based framework that integrates travel-time prediction with order-assignment optimization for last-mile delivery.
\cite{chan2022machine} present a ML-augmented approach to large-scale bilevel decision problems by approximating follower-level reactions.  \cite{hong2023learning} develop a generative metamodeling framework using quantile regression, which acts as a fast surrogate simulator to support real-time decision-making.
\cite{qi2023practical} develop an end-to-end deep learning model that directly learns replenishment strategies from data. 
\cite{ma2025machine} leverage ML models to predict passengers' check-in waiting times in order to improve resource allocation in terminal operations.
\citet{zheng2025preference} tackle dynamic assortment and inventory planning via preference learning.
\cite{huang2025orlm} explore training large language models to automate optimization modeling and solver generation, addressing practical challenges in data management and deployment.

While existing applications of AI have demonstrated effectiveness in various operational settings, our work advances this line of research by embedding structural features of business problems into the algorithmic design process. We introduce a coordination-centric RL framework that explicitly accounts for the functional roles and interdependencies within complex business systems. This design promotes interpretability, adaptability, and generalizability, laying a foundation for trustworthy and scalable RL-based solutions in complex, multi-agent environments.

\textbf{RL in operations and marketing.} 
RL has been widely recognized as a powerful framework for sequential decision-making under uncertainty, particularly in settings where traditional analytical methods may fall short (e.g., \citealt{chen2016dynamic,lin2022dynamic,lei2024joint,jiang2024intertemporal}).
Earlier RL applications in operations and marketing often relied on discretized state-action spaces or heavily simplified environments to ensure computational tractability. For instance, \cite{cao2023safe} develop a model-based RL algorithm for high-stakes applications such as personalized healthcare, which necessitates explicit estimation of the underlying transition dynamics and reward functions using general statistical models. 
The emergence of deep learning has further expanded RL’s applicability by enabling the handling of high-dimensional state spaces and unstructured data. \cite{gijsbrechts2022can} and \cite{oroojlooyjadid2022deep} apply deep RL in single-agent inventory settings.
\citet{liu2023dynamic} adopts a DQN-based approach for high-frequency coupon targeting in livestream shopping, capturing consumer dynamics and behavioral heterogeneity without restrictive assumptions. \citet{cohen2025dynamic} study dynamic pricing under fairness constraints and propose regret-optimal RL algorithms that balance exploration with structural policy considerations.

Despite these advancements, most existing approaches remain rooted in classical single-agent RL paradigms, which fail to account for organizational structures and cross-functional interdependencies prevalent in real-world firms. Though \citet{liu2022multi} implement a general-purpose multi-agent RL algorithm in supply chain contexts, their model assumes structurally similar agents and does not accommodate functional asymmetry. In contrast to prior model-based RL applications, we adopt a \emph{model-free policy-based} RL framework that avoids estimating transition dynamics and reward functions, thereby enhancing scalability and robustness in complex, uncertain business settings. 
Our approach is closely aligned with the recent paradigm shift in RL research that emphasizes heterogeneous agent structures and modular policy architectures.
Building on this perspective, we develop a novel multi-timescale multi-agent RL algorithm that can deal with interdependencies among functionally distinct departments and support effective cross-functional coordination.

\textbf{Deep RL as stochastic approximation (SA).} 
Modern deep RL methods are grounded in two foundational paradigms: value-based and policy-based approaches. Value-based methods, such as DQN and its variants \citep{mnih2015human, hessel2018rainbow}, approximate state–action value functions and select actions with the highest estimated returns. In contrast, policy-based methods like REINFORCE \citep{williams1992simple} directly optimize parameterized policies, and PPO has been a widely adopted RL baseline due to its strong training stability and  data efficiency \citep{schulman2017proximal}.
Building on these foundations, the actor–critic architecture integrates the strength of both value- and policy-based paradigms \citep{konda1999actor2}. 
In this setup, value function estimation is typically easier than policy optimization, which motivates updating the critic on a faster timescale than the actor \citep{konda1999actor}.

Distinct from existing literature on deep RL algorithm, our algorithm design is informed by managerial insights from structural characteristics specific to real-world coordination problems. While it preserves the two-timescale update between the actor and critic, its innovation lies in the principled identification of update rates across heterogeneous agents within the actor system. This design enables a coherent and scalable application of multi-timescale learning to intra-agent coordination, advancing the capabilities of multi-agent RL in complex enterprise environments.

\section{Theoretical Model and Insights}\label{Theoretical Model}

    Here we lay out a theoretical model of how product replenishment and recommendation contribute to a platform's revenue in a dynamic and interactive manner. Even within a relatively simplified framework, the theoretical analysis proves to be intricate and nuanced. Despite this, we are able to derive valuable managerial insights regarding how coordination can be achieved and implemented across these two distinct channels. In the next section, we introduce our RL-based approach that not only tackles the coordination problem in a convenient and unified way but also aligns with the general insights obtained in this section, thereby validating its effectiveness.

\begin{figure}[t]
        \centering
        \includegraphics[trim=0cm -0.5cm 0cm -0.5cm, clip, width=0.725\linewidth]{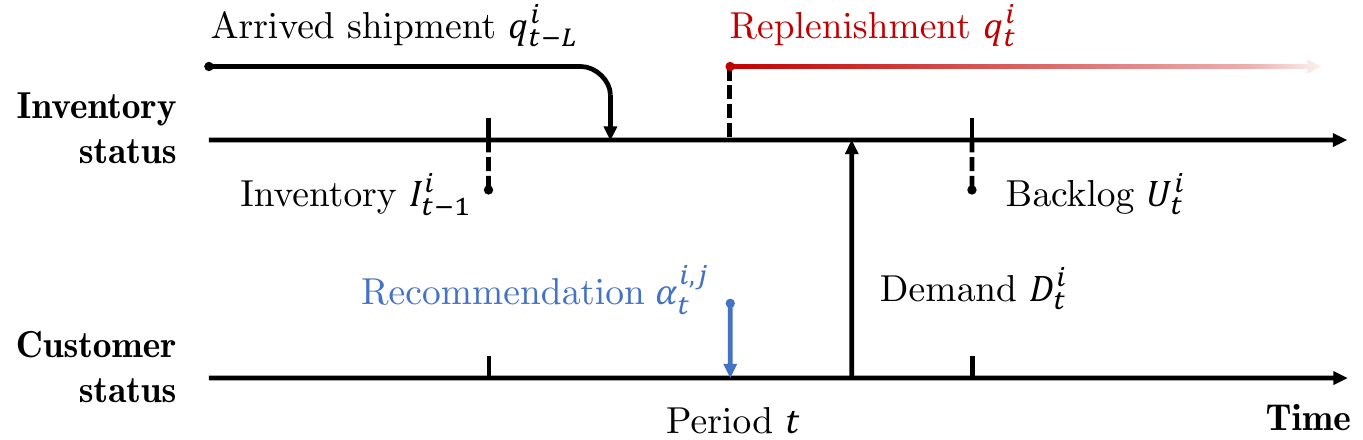}
        \caption{Illustration of the dynamic interaction between the inventory and recommendation systems.}
        \label{fig:timeline}
    \end{figure}

    We consider a platform that sells $N$ products to a fixed pool of $M$ customers with repeated purchases. We have excluded the introduction of new products and customers, assuming the products are essential or consumable items (e.g., toothpaste), to focus on inventory control and recommendation issues. For notational consistency, we use superscripts $i$ and $j$ to denote the $i$-th product and $j$-th customer, respectively, and subscripts $t$ to index time periods. 
    The dynamic interaction between the inventory and recommendation systems is illustrated in Figure \ref{fig:timeline}. Specifically, on the inventory management side, the platform reviews its product stock levels at the beginning of each period $t$ and determines the replenishment quantities $q_t^i$ for each product. These replenished products arrive at the inventory system after a fixed lead time of $L$ periods. Let $I_{t-1}^i$ represent the remaining inventory of product $i$ at the end of period $t-1$. Therefore, the available inventory for sale in period $t$ is given by $I_{t-1}^i+q_{t-L}^i$. Considering a \emph{backlog} setting, the realized sales $S_t^i$ for product $i$ in period $t$ is then determined by both the available inventory and the aggregate demand, which includes the previously unmet and current-period demand, i.e.,
    \begin{align}\label{sales_dynamic_system}
    S_t^i = D_t^i + U_{t-1}^i - \left[D_t^i + U_{t-1}^i - I_{t-1}^i - q_{t-L}^i\right]^+,
    \end{align}
    where $[x]^+:=\max\{0,x\}$, $D_t^i$ represents the demand for product $i$ during period $t$, and $U_t^i := \left[D_t^i + U_{t-1}^i - S_t^i\right]^+$ captures the backlog carried to period $t$. In each period, customer demand is influenced by the platform’s recommendation policy, which will be elaborated later. The remaining inventory for product $i$ at the end of period $t$ is thus updated as follows:
    \begin{align}\label{inventory_dynamic_system}
    I_t^i = \left[I_{t-1}^i + q_{t-L}^i - S_t^i\right]^+.
    \end{align}

    On the marketing side, the platform employs recommendation strategies to affect
    customers' purchasing behaviors, which can manifest through methods such as association-based recommendations and personalized page displays. To quantitatively model the impact of recommendations on purchasing willingness and ultimately, on their purchasing decisions, as shown in the second channel of Figure \ref{fig:timeline}, we assume that a customer’s purchasing willingness evolves according to the following dynamic process:
    \begin{align}\label{willingness_dynamics}
    R_t^{i,j} = \eta R_{t-1}^{i,j} + \mathcal{R}(\alpha_t^{i,j}, \eta R_{t-1}^{i,j}), 
    \end{align}
    where $\alpha_t^{i,j}\in[0,1]$ represents the personalized recommendation of product $i$ to customer $j$ at time $t$, controlled by the platform; $\eta \in (0,1)$ is a decay factor that captures the natural decline in willingness or attention over time; and $\mathcal{R}(\cdot,\cdot)$ is a general function that models the recommendation effect on the current-period purchasing willingness, which may depend on the customer’s previous willingness level. 
    This general dynamic model encompasses many special cases commonly used in the marketing literature. For instance, the advertising effect could exhibit diminishing returns in terms of stimulating customers’ purchasing desires (see, e.g., \citealt{rubel2011optimal,demirezen2016optimization}).
    Customers then make purchasing decisions at each time period based on their willingnesses. Let $d_{t}^{j} = (d_{t}^{1,j}, \ldots, d_{t}^{N,j})$ represent the demand vector of customer $j$ across $N$ products at time $t$. These individual demands are influenced by the \emph{relative} purchasing willingness across all products, rather than by individual absolute values, and are assumed to follow a general distribution $F(\cdot; R_t^{1,j}, \ldots, R_t^{N,j})$. For instance, we may consider a standard choice model where willingness levels are mapped to a probability vector using a softmax function, so that customers randomly select one product at each time step according to these probabilities. The aggregate demand for product $i$ at time $t$, denoted as $D_t^i$, is then the summation of individual demands:
    \begin{align}\label{demand_model}
        D_t^i = \sum_{j=1}^M d_{t}^{i,j}, \quad
        d_{t}^{j} \sim F(d_{t}^{j}; R_t^{1,j}, \ldots, R_t^{N,j}).
    \end{align}

    The platform seeks to jointly optimize inventory replenishment and personalized recommendation strategies to maximize aggregate revenue, after deducting operational expenses.
    Specifically, the profit function for product $i$ during period $t$ is denoted as $P_t^i = \mathcal{P}(S_t^i, q_t^i, I_t^i, U_t^i)$ to account for not only the revenue from product sales and the ordering cost but also penalizing potential mismatches between demand and inventory, such as overage and underage. As before, we use a general functional form to maintain flexibility, as the RL approach introduced in the next section allows us to adapt to different situations based on practical needs.  On the other hand, personalized recommendations typically involve extra operational expenses, which we denote as $C_t^{i,j} = \mathcal{C}(\alpha_t^{i,j})$ for any recommendation strategy set allocated among different combinations of customers and products. In summary, the platform's optimization problem can be formulated as a dynamic planning problem, where the replenishment quantities $(q_t^i)$ and personalized recommendation intensities $(\alpha_t^{i,j})$ are optimized over a horizon of $T$ periods:
    \begin{align}
        \max_{(q_t^i), (\alpha_t^{i,j})} \mathcal{L} :=  \mathbb{E}\left[\sum_{t=1}^{T}\iota^{t-1} \sum_{i=1}^{N} \left(P_t^i - \sum_{j=1}^{M} C_t^{i,j}\right)\right], \label{eq:obj}
    \end{align}
    where $\iota\in(0,1]$ is a discount factor that determines the present value of future rewards,
     and the expectation is taken with respect to the demand uncertainty introduced in \eqref{demand_model}. 
    This joint optimization framework highlights the integrated perspective on how to coordinate two distinct but closely related channels to achieve profit maximization. Nowadays, such coordination is not only essential for reducing operational costs and enhancing efficiency, but increasingly attainable through advanced ML tools, such as the RL approach discussed in the current paper.

    Apparently, the problem is readily complex enough to preclude explicit solutions, and classical methods in the inventory management and marketing literature fall short of providing a good \emph{global} solution in two channels. For this reason, our primary focus in the remainder of this section is to derive useful managerial insights by examining simpler special cases, which will help establish a basic understanding of how coordination works between inventory and marketing channels. These insights are crucial not only for guiding the design of the RL algorithm but also for validating its effectiveness by ensuring the consistency of its numerical behavior with our established understanding. To be more specific, we explore two orthogonal dimensions of possible coordination to overcome the model intractability. The first one is cross-product coordination, where, within the same time period, both inventory decisions and product recommendations need to be made in alignment with products' relative status. The second is intertemporal coordination, where demand can be smoothed over time through strategic recommendations, facilitating better inventory management. Together, these two dimensions illustrate how inventory management and recommendation systems can be coordinated across different products and time periods to manage supply-demand relationships, thereby enhancing overall efficiency and profitability.

\subsection{Cross-Product Synergy: Interdependent Inventory and Marketing Strategies}\label{Cross-Product Synergy: Interdependent Inventory and Marketing Strategies}

We first consider a simplified scenario to shed light on how inventory and marketing strategies closely interact with each other. Specifically, we reduce problem to a single-period model, abstracting away temporal considerations to focus on the coordination mechanism. To enhance readability, we omit time subscripts. The platform manages two products and can implement a recommendation strategy $\alpha^i$ for $i=1,2$ to influence customers' purchasing willingness, which is given by
\begin{align}\label{equation:willingness_special}
    R^{i} = R_{0}^{i} + (\overline{R}- R_{0}^{i}) \alpha^{i},
\end{align}
where $R^i_0$ represents the initial willingness 
and $\overline{R}$ denotes the upper bound on willingness. The recommendation effect exhibits diminishing returns when willingness approaches its upper limit, consistent with findings in the marketing literature (see, e.g., \citealt{rubel2011optimal}). Customer demand is then modeled as a Bernoulli random variable:
\begin{align}\label{logit_model_cross_product}
    D^i \sim \text{Bernoulli}(\gamma^{i}), \quad \text{where } \gamma^{i} = e^{R^{i}} \Big/\sum_{k=1}^2 e^{R^{k}},
\end{align}
indicating that customers are more likely to purchase products with relatively higher willingness levels \citep{lilien1992marketing}. Given any ordering decision $q^i$ made by the platform, the realized sales are then determined as 
$S^i =D^{i}-\left[D^{i}-q^{i}\right]^{+}.$

Analogous to the classical newsvendor model, we define the platform’s profit function to account not only for revenue generation but also for the costs associated with supply-demand mismatches. Specifically, we introduce unit underage and overage costs, denoted by $b$ and $h$, respectively. The resulting profit function for product $i$ is $P^i =p S^i-h I^i-b U^i$, where $p$ is the unit selling price, and the underage and overage amounts are given by $U^i =\left[D^{i}-S^i\right]^{+}$ and $I^i =\left[q^{i}-S^i\right]^{+}$. We further take the marketing expenses to be linear in the marketing effort for convenience, expressed as $C^i=r\alpha^i$, where $r$ is the unit marketing cost. Consistent with our general optimization framework stated previously, the platform seeks to jointly optimize replenishment and marketing strategies to maximize the expected aggregate profit of  
$\mathbb{E} [\sum_{i=1}^2 (P^i-C^i)].$ 

To this end, we aim to investigate the interplay between inventory and marketing strategies within this simplified model, maintaining analytical tractability while capturing the core trade-offs. We begin by characterizing the optimal ordering strategy under a fixed marketing plan. 

\begin{proposition}\label{prop:ordering_to_marketing}
For the considered simplified system, the optimal replenishment quantities $q^{i,*}$ are uniquely determined given each product's recommendation decision $\alpha^i$. As the recommendation intensity $\alpha^i$ for product $i$ increases, its optimal replenishment quantity also increases, while the replenishment for other products decreases.
\end{proposition}

Proposition \ref{prop:ordering_to_marketing} gives the intuitive reaction of replenishment strategies to marketing policies. Specifically, when the platform strengthens the recommendation for a particular product, the resulting increase in demand necessitates a corresponding rise in optimal inventory levels to accommodate it. Meanwhile, since the purchasing probabilities for other products diminish, the need for their replenishment decreases. In particular, the uniqueness of the optimal ordering decisions under a given marketing plan is a direct consequence of the convexity of the objective function with respect to inventory decisions. This property simplifies inventory optimization relative to the more complex recommendation problem, and it further motivates the adoption of a multi-timescale RL approach to solve for optimal solutions, as discussed in Section \ref{subsec:connecting_theory_to_algorithms}.

Before analyzing how marketing strategies respond to replenishment decisions, we first introduce several metrics to facilitate understanding and presentation. 
Given that recommendation decisions should be largely driven by cross-product effects, we focus on \emph{relative} measures to highlight comparisons across products. Without loss of generality, we present the analysis with respect to the first product. 
From \eqref{equation:willingness_special}, we know that marketing efficiency is inversely related to the gap between the original willingness and its upper bound, $\overline{R} - R_0^i$. Accordingly, we define the \emph{relative marketing efficiency} (RME) as $\operatorname{RME}:= (\overline{R} - R_0^1) - (\overline{R} - R_0^2)=R_0^2-R_0^1$, which measures the comparative ease of raising the purchasing willingness. A larger RME indicates that it is more convenient to boost the willingness for product 1. The second dimension, while more nuanced, captures the impact of purchasing willingness on revenue. We define the \emph{relative marketing profitability} (RMP) to be $\operatorname{RMP}:= (p+h+b) ([1-q^2]^+ - [1-q^1]^+)$, 
where each component reflects the platform revenue's sensitivity to changes in purchasing probability. A higher RMP suggests that increasing the likelihood of purchasing product 1 yields greater revenue gains, thereby making its promotion more profitable. We further define the cost-effectiveness coefficient $\zeta = r/(\overline{R} - R_0^1)$, which captures the cost required to raise willingness per unit. With these metrics, we can now characterize the optimal recommendation strategies as follows. 

\begin{proposition}\label{th:recommendation}
For the considered simplified system and given replenishment decisions, the optimal recommendation strategy is determined by the joint effects of relative marketing efficiency and profitability. It falls into two several distinct regimes:
\begin{itemize}
    \item \textbf{No recommendation when relative advantage is insufficient.} Either of the following conditions leads to no recommendation: (a) $0\leq \operatorname{RMP} \leq 4\zeta$; (b) $\operatorname{RME} \leq 0$ and $0\leq\frac{\operatorname{RMP} \exp({\operatorname{RME}}) }{(\exp({\operatorname{RME}}) + 1)^2}  \leq \zeta$.

    \item \textbf{Positive recommendation under significant relative advantage.} When  $\frac{\operatorname{RMP} \exp({\operatorname{RME}}) }{(\exp({\operatorname{RME}}) + 1)^2}  > \zeta$, the optimal recommendation $\alpha^{1,*}$ is positive and unique. No recommendation is made for the second product.
\end{itemize}

In both regimes, the optimal recommendation intensity $\alpha^{1,*}$ increases with both  $\operatorname{RME}$ and $\operatorname{RMP}$, and ultimately, with the relative replenishment quantity $q^1-q^2$.
\end{proposition}
\begin{figure}[h]
    \centering
    \includegraphics[width=0.796\linewidth]{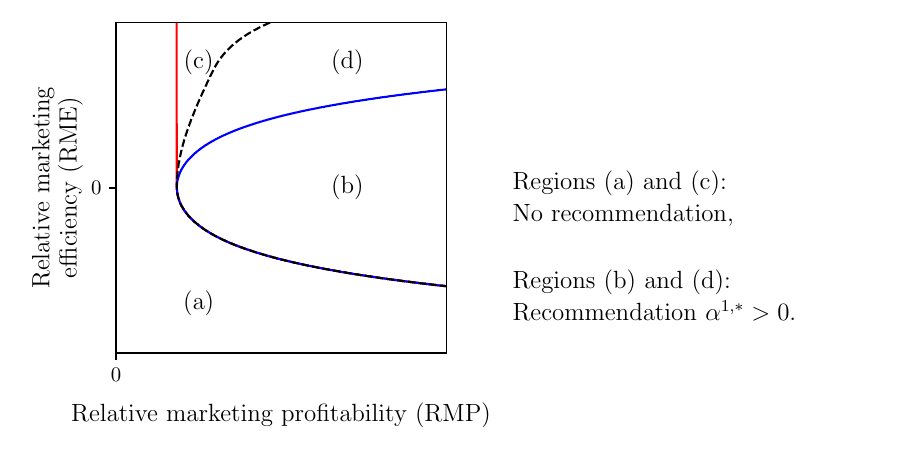}
    \caption{Recommendation decision regions depending on RME and RMP.}
    \label{fig:recom1}
\end{figure}

Due to the logit structure of the demand model, it is never optimal to recommend both products simultaneously. This is because competing recommendations fragment consumer attention and cannot simultaneously increase the demand across multiple products. In this regard, Proposition \ref{th:recommendation} uncovers the underlying principle guiding the selection of which product to promote. Specifically, relative profitability plays a decisive role: promotion occurs only when RMP is positive. Conversely, if RMP is negative, the promotional focus naturally shifts to the second product, and a symmetric conclusion can be drawn regarding its optimal recommendation strategy. Once a product is identified as relatively profitable to promote, the firm next evaluates its relative efficiency. A positive promotion decision is made only if RME exceeds a certain threshold, ensuring that the returns from increasing willingness outweigh the costs. 

The corresponding recommendation decision regimes are illustrated in Figure \ref{fig:recom1}. The analytically derived boundaries in Proposition \ref{th:recommendation} delineate two main regions: (a) a no-recommendation regime and (b) a positive-recommendation regime. They are shown as distinct areas in the figure. For the remaining areas, although a closed-form boundary is not available, the black dashed line effectively separates the recommendation and no-recommendation regimes. These results once again confirm that in a competitive environment, the firm prefers to allocate recommendation efforts to products with relatively high profitability and efficiency. It echoes the essential nature of the problem as one of \emph{resource allocation}, where marketing efforts are directed toward opportunities with the highest marginal return relative to cost. Furthermore, as a product becomes more profitable or its recommendation becomes more efficient, it is not only more likely to be recommended, but also to receive a higher promotion intensity.

Naturally, the recommendation strategy is closely intertwined with inventory decisions in a coordinated manner. On one hand, the recommendation boundaries illustrated in Figure \ref{fig:recom1} explicitly depend on both RMP and RME, which are functions of the ordering decisions for the two products. On the other hand, Proposition \ref{th:recommendation} reinforces that recommendation strategies should align with inventory decisions: when a larger order is placed for a product, the firm should correspondingly implement a stronger recommendation effort to help accelerate inventory turnover. This insight is consistent with the observations from Proposition \ref{prop:ordering_to_marketing}, further validating the importance of coordinated decision-making between two channels to enhance operational performance. However, even in this simplified setting, obtaining globally optimal inventory and recommendation strategies remains analytically intractable. This underscores the necessity of adopting an RL approach, which not only accommodates the complexity and interdependence of these two channels but also yields effective and practically implementable strategies, as we demonstrate in Section \ref{sec:multi_scale_RL}.

\subsection{Intertemporal Coordination: Strategical Demand Smoothing and Adaptive Ordering}\label{subsec:Intertemporal Coordination}

Coordination can occur not only across products but also across time periods. In this subsection, we turn our attention to the vertical dimension---intertemporal coordination---by investigating how replenishment and recommendation strategies dynamically adjust over time to achieve system-wide optimization. To facilitate analysis and isolate the effects of temporal coordination, we consider another simplified setting involving a single product over two periods. For notational clarity and consistency, we omit superscripts in this subsection.

Customer behavior is modeled as repeated binary purchase decisions across periods. Specifically, demand in each period follows a Bernoulli distribution:  $D_i \sim \text{Bernoulli}(\gamma_{t})$, where the purchase probability is given by $\gamma_{t} = 1 / (e^{-R_{t}}+1)$.
Here, $R_{t}$ denotes the customer's purchase willingness in period $t$, which follows the dynamics $R_t=\eta R_{t-1} + (\bar{R}- \eta R_{t-1})\alpha_t$, driven by the firm’s recommendation strategy and higher recommendation intensity leads to higher willingness. Each period results in either a purchase or no purchase. Under a backlog setting without lead time, the backlog quantity $U_t$, inventory level $I_t$, and realized sales $S_t$ obey the same dynamics introduced in the general framework; see, e.g., \eqref{sales_dynamic_system} and \eqref{inventory_dynamic_system}. As in Section \ref{Cross-Product Synergy: Interdependent Inventory and Marketing Strategies}, we adopt a linear recommendation cost $C_t=r\alpha_t$ and a profit function penalized by both underage and overage costs, given by $P_t =p S_t-h I_t-b U_t$. The firm’s objective is to maximize the expected cumulative profit net of recommendation costs over a two-period horizon, given by $\mathbb{E} [\sum_{t=1}^2 (P_t-C_t) ]$.

In this illustrative model, we are able to uncover several fundamental and insightful patterns of coordinated dynamic decision-making. These include \emph{demand smoothing}, achieved through strategic recommendations that align customer demand with inventory availability, and \emph{adaptive ordering}, whereby inventory decisions are dynamically adjusted to reflect evolving customer preferences. Notably, these patterns persist in the general setting and will be revisited in numerical experiments.

\begin{proposition}[Demand Smoothing]\label{prop:demand_smoothing}
Given a fixed rational aggregate replenishment budget $\sum_t q_t$, and assuming that $p\geq b e^{\overline{R}}-h$, increasing the inventory allocated to a specific period 
 (e.g., by raising $q_1$) 
leads to a higher corresponding ratio of optimal recommendation intensities (e.g., $\alpha_1^*/\alpha_2^*$).
\end{proposition} 

Proposition \ref{th:recommendation} has implicitly suggested that stronger recommendations are required when a product’s (relative) inventory level is higher. Building on this insight, Proposition \ref{prop:demand_smoothing} further demonstrates that while recommendation intensity increases in response to inventory pressure, firms must strategically allocate their recommendation budget over time to smooth demand in accordance with the inventory schedule. For example, if inventory pressure is disproportionately high in the early periods, it is necessary to deploy strong recommendations upfront, even when customers’ initial purchase willingness is low. Through this early intervention, and given the retention effect in customers’ purchase willingness, demand can be strategically shifted forward in time, helping to alleviate early-stage inventory pressure. This coordinated adjustment between recommendation timing and inventory levels is what we refer to as \emph{demand smoothing}. In fact, demand smoothing is particularly important in large-scale systems where thousands of products with differing inventory cycles coexist. Factors such as lead time, replenishment frequency, and capacity constraints often result in complex and asynchronous inventory dynamics. In such environments, demand smoothing serves as a key mechanism to align demand trajectories with fluctuating inventory patterns, thereby promoting operational flexibility and greater system-wide efficiency.

\begin{proposition}[Adaptive Ordering]\label{prop:adaptive_ordering}
Given a fixed aggregate purchase willingness $\sum_t R_t$ (shaped by the recommendation strategy), and assuming that $p\geq h + b^2/h$, an increase in willingness for a particular period 
leads to an increase in the corresponding ratio of replenishment quantities.
\end{proposition}

Purchase willingness directly decides the customer demand. When the temporal demand trend shifts, Proposition \ref{prop:adaptive_ordering} shows that the corresponding ordering policy should adjust accordingly. For instance, stronger early-period demand resulted from recommendations naturally calls for a more front-loaded replenishment strategy. The mild technical condition in Proposition \ref{prop:adaptive_ordering} merely requires that the selling price is sufficiently high relative to the underage and overage costs. 
The observed behavior is indeed different from classical, uncoordinated systems in which inventory decisions are made independently of marketing considerations. In a coordinated system, inventory planning is not just reactive to realized demand; rather, it actively collaborates with marketing strategies to achieve improved outcomes. This is possible because demand is endogenously shaped by the platform’s marketing efforts. In particular, due to the retention effect in purchase willingness, current-period marketing actions also influence willingness, and thus demand, in future periods. For this reason, we refer to this behavior as \emph{adaptive ordering}, which fundamentally differs from conventional interpretation as a passive reaction to demand forecasts. Instead, it reflects a proactive and coordinated adjustment of inventory in response to the marketing-induced demand dynamics.

\subsection{Connecting Theory to Algorithms}\label{subsec:connecting_theory_to_algorithms}

Thus far, we have established key insights into how a coordinated system integrating inventory and recommendation should operate. However, even under a significantly simplified setting, the theoretical analysis remains highly complex. In the general case where decisions must be made in a high-dimensional, dynamic environment, classical analytical methods fall short of producing tractable solutions. This motivates us to use RL techniques, which are well-suited to solving sequential decision problems in such environments.

In the following section, we introduce an RL-based framework that employs parameterized decision rules---policy functions that map observable system states to actions. These rules are updated iteratively through feedback from observed outcomes, allowing the system to refine its decision-making over time. Importantly, the theoretical insights developed earlier are far from redundant; rather, they play a crucial role in guiding algorithm design, offering structural understanding, and serving as benchmarks for evaluating the quality and reasonableness of the solutions produced by the RL algorithm. We elaborate on these connections below.

\begin{enumerate}
\item \textbf{A multi-agent perspective to facilitate learning.}
In coordinated decision settings, the dimensionality of the action space increases significantly compared to isolated inventory or recommendation optimization problems. This makes the learning task computationally intensive and complex. Inspired by our prior theoretical analysis, we adopt a modular decomposition that treats the inventory and recommendation components as two interdependent agents. This decomposition enables a form of conditional learning, where one agent’s policy is updated while treating the other’s as fixed, and vice versa. Although this approach may yield a local equilibrium rather than a global optimum, similar to the limitations of global search methods in high-dimensional spaces, it greatly reduces computational complexity and is naturally suited to the multi-timescale learning structure introduced later. Importantly, both modules are trained jointly toward a common profit objective, preserving coordination despite the decoupled learning process. This architecture also mirrors practical organizational structures, where separate operational teams manage inventory and recommendation functions. Thus, the proposed multi-agent framework not only improves scalability but also enhances real-world implementability without compromising coordination.

\item \textbf{A multi-timescale scheme for state-based policy learning.}
Our theoretical findings reveal distinct differences in the stability and complexity of two decision components. Inventory-related decisions tend to be more structurally stable and can adjust effectively once demand materializes. In contrast, recommendation policies involve more intricate feedback loops and are highly sensitive to parameter changes, which can trigger cascading behavioral responses. These structural differences suggest that applying a uniform learning rate across all components may hinder convergence and impair learning efficiency. A learning rate that is too small can significantly slow convergence for stable components; conversely, a rate that is too large may destabilize sensitive components and lead to oscillatory behavior.

To address this issue, we adopt a multi-timescale SA framework. Under this scheme, components with more stable dynamics (e.g., inventory decisions) are updated using larger step sizes, while more sensitive components (e.g., recommendation policies) are updated more conservatively. Although all decision components are updated at each iteration, the relative step sizes determine their respective responsiveness. In effect, larger step sizes act as proxies for faster adaptation, even if the update frequency remains the same.
This multi-timescale structure not only reflects the theoretical geometry of the problem but also improves convergence stability in practice. By decoupling update magnitudes, the algorithm mitigates mutual interference among components and maintains coherence during learning. It also aligns with real-world operational logic, where straightforward tasks are addressed more rapidly while complex decisions require greater caution and iteration.

 \item \textbf{General behavioral guidelines for validating RL solutions.} The structural insights developed in Sections \ref{Cross-Product Synergy: Interdependent Inventory and Marketing Strategies} and \ref{subsec:Intertemporal Coordination} offer high-level behavioral expectations that serve as useful guidelines for evaluating RL-generated solutions. Specifically, we expect to observe: (1) coordinated inventory and recommendation strategies, with promotional efforts concentrated on products exhibiting strong relative efficiency and profitability; and (2) intertemporal coordination patterns, such as demand smoothing and adaptive ordering, aligning demand trajectories with inventory dynamics.
 These general principles are critical for ensuring the practical reliability of RL–based operational strategies. They provide interpretability and help practitioners build confidence in the robustness and effectiveness of the learned policies. 
\end{enumerate}

\section{A Multi-Timescale RL Approach}\label{sec:multi_scale_RL}

In this section, we propose a multi-timescale, multi-agent RL algorithm to solve for coordinated strategies in high-dimensional, dynamic environments, building on the theoretical insights developed earlier. Section \ref{subsec4.1:multi-timescale_SA} introduces a multi-timescale SA framework tailored to our setting. In Section \ref{subsec4.2:cooperation_markov_game}, we reformulate the coordination problem as a Markov game, which supports the development of a multi-agent RL algorithm with structured multi-timescale updates for joint decision-making in Section \ref{subsec4.3:multi-timescale_multi_agent_RL}. Unlike conventional single-agent RL approaches, such as deep Q-learning (DQN; \citealt{mnih2015human,hessel2018rainbow}) and proximal policy optimization (PPO; \citealt{schulman2017proximal}), our multi-agent method adopts a modular architecture that decomposes the global reward and assigns differentiated credit to individual agents.
The integration of multi-timescale updates further enhances learning efficiency and yields strong performance within limited training horizons. Overall, the proposed algorithm captures the coordination mechanism revealed in theory and offers a scalable, adaptive solution for learning stable policies in complex environments.

\subsection{Multi-Timescale SA}\label{subsec4.1:multi-timescale_SA}

Analytical solutions to the joint optimization of inventory replenishment and recommendation are generally intractable, necessitating the use of iterative algorithms. Let $\theta$ denote the joint vector of decision policies. The iterative update procedure follows the form:
 $$\theta[n+1] = \theta[n] + \varepsilon[n] \cdot G(\theta[n], \xi[n]),$$ 
where $\theta[n]$ is the policy vector at iteration step $n$; $\varepsilon[n]$ is the step size, often referred to as the \emph{timescale} or learning rate; $\xi[n]$ is a random variable that captures the system stochasticity; and $G(\theta[n], \xi[n])$ represents a performance-improving direction based on the current policy and observation, typically the gradient of the expected profit function.

While this \emph{uniform} update scheme provides a unified approach to joint policy optimization, its direct application to complex, coupled systems with heterogeneous structures may lead to poor performance. As revealed in Section \ref{Theoretical Model}, inventory and recommendation decisions differ significantly in optimization complexity and structural properties. Applying a single learning rate to both can lead to instability or inefficient convergence, as the components may interfere with each other's learning process. To address this, we adopt a multi-timescale SA framework that assigns distinct step sizes to different components of the decision vector. This design allows each policy component to adapt at a rate aligned with its structural complexity, thereby improving both learning stability and convergence efficiency.  

\begin{figure}[t]
\centering 
\includegraphics[trim=0.3cm 0.7cm 0.3cm 0.7cm, width=0.6\linewidth]{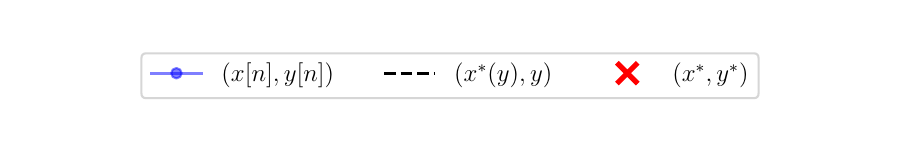}
\vspace{0.5em}\par
\subfigure[\parbox{\linewidth}{Trajectory of slow single-timescale updates.}
]{
\label{fig:stss}
\includegraphics[trim=0cm 0.5cm 0cm 0.5cm, width=0.36\linewidth]{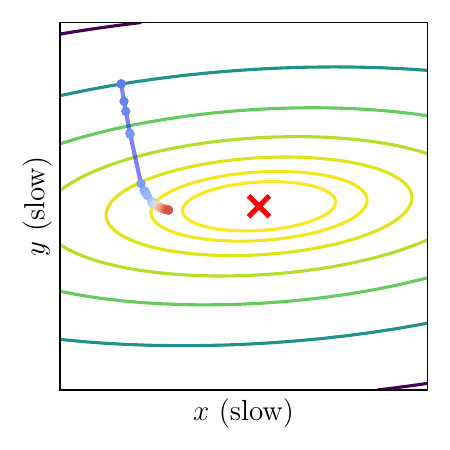}}
 \hspace{0.66cm}
\subfigure[\parbox{\linewidth}{Trajectory of fast single-timescale updates.}]{
\label{fig:stsf}
\includegraphics[trim=0cm 0.5cm 0cm 0.5cm, width=0.36\linewidth]{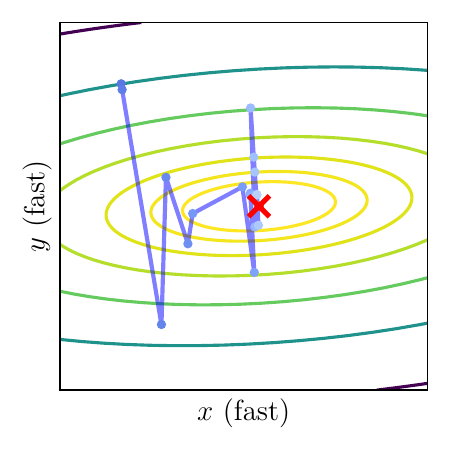}}
\vspace{0.5em}\par
\subfigure[\parbox{1.6\linewidth}{Trajectory of multi-timescale updates.}]{
\label{fig:mts}
\includegraphics[trim=0cm 0.5cm 0cm 0.5cm, width=0.36\linewidth]{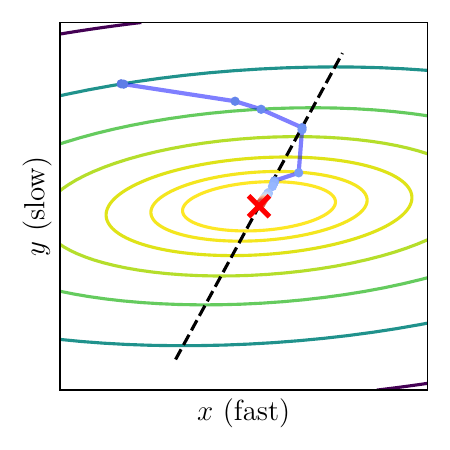}}
\hspace{0.66cm}
\subfigure[\parbox{\linewidth}{Multi-timescale updates for each variable.}]{
\label{fig:mts2}
\includegraphics[trim=0cm 0.5cm 0cm 0.5cm, width=0.36\linewidth]{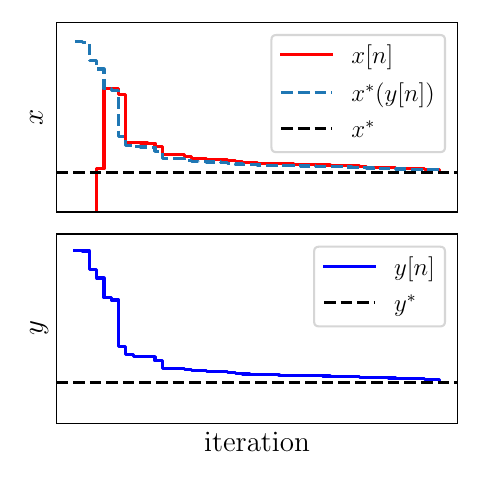}}
\caption{Joint policy optimization via stochastic approximation. $(x[n], y[n])$ denotes the joint policy at step $n$;
$(x^*, y^*)$ indicates the globally optimal policy; and $(x^*(y), y)$ represents the optimal fast-timescale response $x$ given a fixed slow-timescale policy $y$. Panels (a), (b), and (c) use the same number of iterations.
}
\label{fig:mtscompareplots}
\end{figure}

Figure \ref{fig:mtscompareplots} illustrates the rough idea of multi-timescale SA for a two-component coupled strategy $\theta:=(x,y)$, where the optimization complexity of $x$ is significantly lower than that of $y$. In Figures \ref{fig:stss} and \ref{fig:stsf}, a uniformly slow or fast update scheme is applied, both resulting in suboptimal performance. Specifically, slow updates lead to sluggish convergence due to limited responsiveness in Figure \ref{fig:stss}, whereas overly aggressive updates induce oscillations and possible coordination failure in Figure \ref{fig:stsf}. In high-dimensional settings, such instability in one component can further compromise learning in others due to coupling.
In contrast, Figures \ref{fig:mts} demonstrates that multi-timescale updates achieve smooth and efficient convergence to the optimal policy within the same number of iterations. In particular, Figure \ref{fig:mts2} indicates that the fast-timescale component effectively tracks its conditional optimum as the slower component evolves gradually.

In our problem, adopting a multi-timescale approach is motivated by the structural asymmetry identified in Section \ref{Theoretical Model}. Recommendation decisions are interdependent across products and promoting one item may affect the demand for others, making the optimization landscape more complex. On the contrary, inventory decisions are conditionally independent and more tractable given the demand distribution. As a result, we assign larger step sizes to inventory updates to leverage their lower complexity and smaller step sizes to recommendation updates to ensure stability amid cross-product interactions.

To illustrate it in our framework, we consider a single-period instance of the general model in Section \ref{Theoretical Model}. Let $\alpha$ and $q$ denote the collective recommendation and inventory decisions, respectively. In this context, we propose the following two-timescale SA algorithm:
\begin{align}
q[n+1] =&\ \min\{\max\{ q[n] + \varepsilon_{1}[n]\ Q[n],\ 0 \},\ \overline{q}\}  , \label{iter:1}\\
\alpha[n+1] =&\  \min\{\max\{ \alpha[n] + \varepsilon_{2}[n]\ A[n],\ 0 \},\ 1\}, \label{iter:2}
\end{align}
where the step sizes satisfy $\varepsilon_{2}[n]/ \varepsilon_{1}[n] \to 0$, indicating that the inventory decision $q$ is updated on a faster timescale; $\bar{q}$ denotes the capacity constraint; and $Q[n]$ and $A[n]$ represent the estimated gradients of the expected revenue function $\mathcal{L}$ with respect to the replenishment and recommendation decisions so that $\mathbb{E}[Q[n]] = \nabla_q \mathcal{L}(q[n],\alpha[n])$ and $\mathbb{E}[A[n]] = \nabla_{\alpha} \mathcal{L}(q[n],\alpha[n])$. Detailed expressions for $Q[n]$ and $A[n]$ are provided in Appendix \ref{appendix4.1}. The following technical assumption is required to establish the asymptotic convergence properties of the two-timescale algorithm.

\begin{assumption}\label{assumption:stepsize}
The step sizes $\{(\varepsilon_1[n],\varepsilon_2[n])\}_n$ satisfy the following conditions:
(a) $\varepsilon_i[n]>0$, $\sum\varepsilon_i[n]=\infty$, $\sum\varepsilon_i[n]^2<\infty$, for $i=1,2$; and (b) $\varepsilon_{2}[n] = o(\varepsilon_{1}[n])$.
\end{assumption}

\begin{theorem}\label{mtsconv2}
If $T=1$ and Assumptions \ref{assumption:stepsize}, \ref{assumption:Lip}-\ref{assumption:concave} hold, the sequence $\{(q[n],\alpha[n])\}_n$ generated by updates \eqref{iter:1}-\eqref{iter:2} converges almost surely to a stationary point of the coordination problem \eqref{eq:obj} with the constraint $q=q^*(\alpha)$. 
Furthermore, if the profit function $\mathcal{L}$ is strictly concave in recommendation decisions $\alpha$, these updates converge to the global optimum $(q^{*}, \alpha^{*})$ with probability one.
\end{theorem}

Theorem \ref{mtsconv2} formalizes the intuition shown in Figure \ref{fig:mtscompareplots}: multi-timescale updates enable the fast component (i.e., inventory policy) to rapidly track its conditional optimum given the slow component (i.e., recommendation decision). Once the best-response inventory mapping is stabilized, it effectively reduces the dimensionality of the search space, enabling convergence analysis regarding the expected profit function with the fast variable replaced by its conditional optimum. Importantly, unlike coordinate-wise alternating optimization, which requires fully solving one subproblem before updating the other, the multi-timescale scheme permits simultaneous updates of both components without sacrificing convergence guarantees. Furthermore, under stronger conditions such as strict concavity conditions, the algorithm is guaranteed to converge to the global optimum of the joint decision-making problem.

\subsection{Formulating Coordination as a Markov Game}\label{subsec4.2:cooperation_markov_game}

To make the coordinated optimization problem amenable to algorithmic learning, we formalize it as a Markov decision process (MDP), defined by the $5$-tuple $(\mathcal{S}, \mathcal{A}, p, r, \eta)$. Here, $\mathcal{S}$ and $\mathcal{A}$ denote the state and action spaces, representing all attainable system states and admissible action combinations, respectively; 
$p:\mathcal{S}\times\mathcal{A}\times\mathcal{S}\mapsto [0,1]$  is the state-transition probability function that captures the system dynamics; $r:\mathcal{S}\times\mathcal{A}\times\mathcal{S}\mapsto\mathbb{R}$ is the reward function that quantifies the immediate gain from taking an action; and $\iota\in(0,1]$ is the discount factor, commonly used in large-horizon problems to account for the time value of rewards.
Actions are selected according to a joint policy $\pi(\cdot|s;\theta)$, parameterized as a conditional probability distribution over the action space given the current state $s$. In practice, this policy is often implemented using a neural network coupled with stochastic sampling. Let $s_t$ and $a_t$ 
denote the state and action at the end of period $t\geq 0$.  The objective is to learn the optimal policy parameter $\theta$ that maximizes the expected cumulative discounted reward:
\begin{align}
    \max_{\theta} \mathcal{L}(\theta)=\max_{\theta} \mathbb{E}_{s_0\sim p(\cdot),a_1\sim \pi(\cdot|s_0;\theta),\cdots,s_T\sim p(\cdot|s_{T-1},a_{T})}\bigg[\sum_{t=1}^{T} \iota^{t-1} r(s_{t},a_t,s_{t-1})\bigg],\label{eq:rl_prob}
\end{align}
where $r(s_t,a_t,s_{t-1}) = \sum_{i=1}^N (P_t^i-\sum_{j=1}^MC_t^{i,j})$ denotes the per-period reward, and system dynamics follow the specification in Section \ref{Theoretical Model}. For notational simplicity, we abbreviate the underlying sources of stochasticity in the expectation operator as, e.g., $\mathbb{E}_{\substack{{s_{0:\infty\sim p}}\\{a_{1:\infty\sim\pi(\theta)}}}}[\cdot]$, in the remainder of the paper.

\begin{figure}[t]
\centering 
\includegraphics[width=0.96\linewidth]{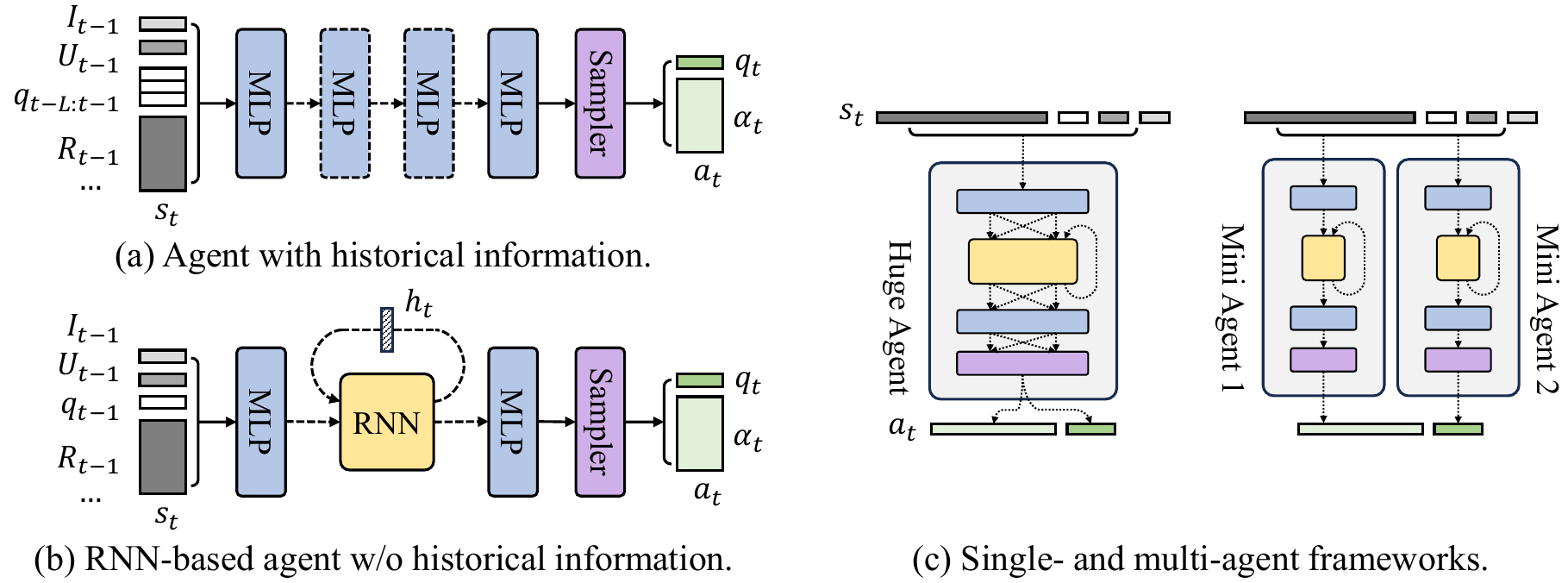}
\caption{Comparison of decision-making pipelines under different policy parameterizations.
}
\label{fig:nnstructure}
\end{figure}

In the cooperative replenishment and recommendation setting, a key modeling challenge lies in the specification of the system state. Due to lead times, replenishment decisions affect future inventory levels with delay, meaning that current observations alone may not sufficiently capture the system dynamics. Importantly, using only current observations as the state violates the Markov property: future system evolution depends not only on the current state and action but also on past decisions. This breaks a foundational assumption of the Bellman principle, thereby undermining the theoretical validity of RL and dynamic programming approaches.
For this reason, the state representation must incorporate relevant historical variables alongside current observations. Specifically, the state should include operational information such as on-hand inventories or shortages, in-transit replenishment quantities over the past $L$ periods, and the current purchase willingness relevant for recommendation decisions, as illustrated in Figure \ref{fig:nnstructure}(a). 
Alternatively, memory can be embedded directly within the policy network. Unlike value-based methods such as DQN, often limited to small-scale, discrete action spaces, policy-based approaches in complex settings typically combine a neural network (e.g., a multilayer perceptron, or MLP) with a stochastic sampler. The network maps input states to parameters of a probability distribution from which actions are drawn. Instead of expanding the state space, we may retain current observations as input and incorporate memory via a recurrent neural network (RNN), which compresses historical information into latent representations, as shown in Figure \ref{fig:nnstructure}(b). When lead times are moderate, RNN-based policies can perform comparably to MLP-based policies with explicit historical states, while offering reduced dimensionality and improved training efficiency.

In the earlier MDP formulation, the joint policy $\pi(\cdot|s;\theta)$ is represented by a monolithic neural network that maps the global state $s_t$ to all components of the decision variables. While such a single-agent RL framework is conceptually straightforward for joint optimization, it is often impractical in real-world applications where different functional decisions may not be centralized, and the number of trainable parameters becomes prohibitively large. 
To address this, we reformulate the problem as a \emph{cooperative Markov game}, introducing a structured multi-agent system comprising a set of mini agents indexed by $k = 1, \ldots, K$. Each agent $k$ has its own policy $\pi_k(\cdot \mid s_{t-1}; \theta_k)$, and the joint policy is formed as $\pi(\cdot \mid s_{t-1}; \theta) = \prod_{k} \pi_k(\cdot \mid s_{t-1}; \theta_k)$, where $\theta = (\theta_1,\ldots,\theta_K)$. The agents jointly generate the system action $a_t = (a_t^1,\ldots,a_t^K)$, and all share a global reward signal based on the overall system state and joint action, aligning their objectives.
Following the structural decomposition motivated by our theoretical analysis, we adopt a two-agent architecture, where one agent $\pi_q$ manages replenishment and the other $\pi_\alpha$ handles recommendation. This modular design can be further refined by assigning an agent to a small bundle of products, allowing fine-grained decision-making. Such decomposition remains effective as long as each mini-agent’s neural network is expressive enough to capture the relevant decision complexity. As shown in Figure~\ref{fig:nnstructure}(c), the multi-agent architecture (right) can significantly reduce the number of redundant neural connections compared to the single-agent setup (left), thereby lowering the dimensionality of the optimization problem and easing computational demands.

To support efficient joint learning under this multi-agent structure, we adopt the \emph{centralized training with decentralized execution} paradigm. During training, agents have access to global state and shared rewards, enabling coordinated learning. At execution, however, they operate independently without real-time communication, ensuring scalable and decentralized deployment. 
In practice, agents often have access only to partial state information due to organizational or informational constraints. To handle this, each agent can be equipped with an RNN module that encodes historical information into latent features, allowing decisions based on local observations while maintaining the same input-output structure. Such decentralization substantially reduces the number of trainable parameters and enhances scalability, particularly in large-scale systems.

\subsection{Multi-Timescale RL for Effective Coordination}\label{subsec4.3:multi-timescale_multi_agent_RL}

Improving RL efficiency is essential for solving the joint optimization problem, particularly as standard RL methods may struggle to learn stable and coordinated behaviors. This subsection illustrates how multi-timescale techniques can be strategically integrated into a multi-agent RL framework to enable efficient learning. We begin by revisiting policy improvement in the single-agent setting and then extend these insights to the multi-agent case, establishing asymptotic guarantees for the multi-timescale algorithm tailored to our problem structure.

In the single-agent case, define the state and state-action value functions under policy $\theta$ as 
$V(s;\theta):=\mathbb{E}_{\substack{{s_{1:\infty}\sim p}\\{a_{1:\infty}\sim\pi(\theta)}}}[\sum_{t=1}^\infty \iota^{t-1} r_t|s_0=s]$
and $Q(s,a;\theta):=\mathbb{E}_{\substack{{s_{1:\infty}\sim p}\\{a_{2:\infty}\sim\pi(\theta)}}}[\sum_{t=1}^\infty \iota^{t-1} r_t|s_0=s, a_1=a]$, where $r_t$ is the per-period reward defined in \eqref{eq:rl_prob}. The advantage function is defined as $A(s,a;\theta):=Q(s, a;\theta) - V(s;\theta)$, which evaluates the relative merit of action $a$ at state $s$ under policy $\theta$.
The performance improvement of a new policy $\tilde{\theta}$ over $\theta$ can be expressed as the expected aggregate advantage \citep[see, e.g.,][]{kakade2002approximately}:
\begin{equation}
    \begin{aligned}
    \Delta(\tilde\theta|\theta)&:= \mathcal{L}(\tilde\theta)-\mathcal{L}(\theta)= \mathbb{E}_{\substack{s_{0:\infty} \sim p\\a_{1:\infty} \sim \pi(\tilde\theta)}} \big[\sum_{t=1}^\infty \iota^{t-1} A(s_{t-1},a_t;\theta)\big].\label{eq:imp}
\end{aligned}
\end{equation}
This motivates choosing policy updates with $\Delta(\theta[n+1]| \theta[n])\geq 0$, ensuring monotonic improvement toward a stationary solution. Let $\rho(s;\theta):=\sum_{t=0}^\infty \iota^{t} P(s_t=s|\theta)$ denote the unnormalized discounted state-visitation distribution under policy $\theta$. Then, the policy improvement \eqref{eq:imp} can be rewritten as $\Delta(\tilde\theta|\theta) = \mathbb{E}_{\substack{{s\sim\rho(\tilde\theta)}\\{a\sim\pi(\tilde\theta)} }}[A(s,a;\theta)]=\mathbb{E}_{\substack{{s\sim\rho(\tilde\theta)}\\{a\sim\pi( \theta)} }}[\frac{\pi(a|s;\tilde\theta)}{\pi(a|s;\theta)}A(s,a;\theta)]$.
However, using off-policy data from $\theta$ to estimate the improvement under $\tilde \theta$ introduces distributional mismatch due to the change in $\rho$, making prior samples invalid and thus reducing sample efficiency. To tackle this issue, a surrogate objective is considered:
 $\tilde\Delta(\tilde\theta|\theta)  = \mathbb{E}_{\substack{{s\sim\rho( \theta)}\\{a\sim\pi( \theta)} }} [\frac{\pi(a|s;\tilde\theta)}{\pi(a|s;\theta)}A(s,a;\theta)]$, which enables efficient sample reuse by avoiding dependency of $\rho$ on $\tilde \theta$.
\cite{schulman2015trust} show that the gap between the true and surrogate improvements is bounded from above by
$B(\theta)\max_s\text{KL}(\pi(\cdot|s;\theta)\Vert\pi(\cdot|s;\tilde\theta))$, where $B(\theta):={4\eta\max_{s,a}\vert A(s,a;\theta)\vert}/{(1-\eta)^2}$ and $\text{KL}(\cdot\Vert\cdot)$ denotes the Kullback-Leibler divergence. This provides a principled constraint on how far the new policy can deviate from the current one. Consequently, the updated policy $\theta[n+1]$ can be obtained by solving the following surrogate problem:
\begin{align}
    \theta[n+1]=\arg\max_\theta \mathbb{E}_{\substack{{s\sim\rho( \theta[n])}\\{a\sim\pi(\theta[n])} }}\bigg[\frac{\pi(a|s;\theta)A(s,a;\theta[n])}{\pi(a|s;\theta[n])}\bigg] - B(\theta[n])\max_s\text{KL}(\pi(\cdot|s;\theta[n])\Vert\pi(\cdot|s;\theta)).\label{eq:trpo}
\end{align}
This formulation leverages importance sampling to reuse historical data and allows stable updates within a trust region, substantially improving sample efficiency in RL.

In the cooperative Markov game setting, we extend the idea behind \eqref{eq:trpo} to evaluate and improve each agent's policy within a multi-agent framework.
For each agent $z\in\{q,\alpha\}$, define the state-action value function $Q_z(s,a_z;\theta):= \mathbb{E}_{A_{z'}\sim\pi_{z'}(\theta_{z'})}[Q(s,a;\theta)|_{a_{z'}=A_{z'}}]$, which captures the expected value of agent $z$'s action, marginalizing over the other agent's behavior. The corresponding agent-specific advantage function is defined as $A_{z, z'}(s,a;\theta) := Q(s,a;\theta) - Q_{z'}(s,a_{z'};\theta)$, measuring the net benefit of agent $z$ taking action $a_z$ given that the other agent takes action $a_{z'}$. Analogous to the single-agent case, the partial advantage of $a_{z}$ is defined as $A_{z}(s,a_{z};\theta):=Q_z(s,a_{z};\theta)-V(s;\theta)$.
The multi-agent advantage decomposition theorem of  \cite{kuba2021trust} yields 
the surrogate improvement $\tilde\Delta(\tilde\theta|\theta)=\mathbb{E}_{\substack{{s\sim\rho( \theta )}\\{a\sim\pi(\theta )} }}\big[ \frac{\pi(a|s;\tilde\theta)}{\pi(a|s;\theta)} (A_z(s,a_{z};\theta)+A_{z',z}(s,a;\theta))\big]$, with the surrogate gap upper-bounded in an agent-wise manner because $\max_s\text{KL}(\pi(\cdot|s;\theta)\Vert\pi(\cdot|s;\tilde\theta))\leq \sum_{z\in\{q,\alpha\}} \max_s\text{KL}(\pi_z(\cdot|s;\theta_z)\Vert\pi_z(\cdot|s;\tilde\theta_z))$.  
This leads to the multi-agent surrogate objective:
\begin{equation}
\resizebox{\linewidth}{!}{$%
\begin{aligned}
\mathbb{E}_{\substack{{s\sim\rho( \theta[n])}\\{a\sim\pi(\theta[n])} }}\bigg[\frac{\pi(a|s; \theta)}{\pi(a|s;\theta[n])} &(A_z(s,a_{z};\theta[n])+A_{z',z}(s,a;\theta[n])) \bigg]- B(\theta[n])\sum_{z\in\{q,\alpha\}}\max_s\text{KL}(\pi_z(\cdot|s;\theta_z[n])\Vert\pi_z(\cdot|s;\theta_z)),
\end{aligned}\label{eq:matrpo}
$}%
\end{equation}
which motivates an extension of \eqref{eq:trpo} to the multi-agent setting that sequentially updates policies in a random order based on individual advantage estimates.
At each iteration, one agent $z\in\{q, \alpha\}$ is sampled uniformly at random, and policies are updated sequentially using the rules:
\begin{equation}\label{eq:matrpo2}
\resizebox{\linewidth}{!}{$%
 \begin{aligned}
     \theta_{z}[n+1]&=\arg\max_{\theta_z\in\Theta_z}\mathbb{E}_{\substack{{s\sim\rho( \theta[n])}\\{a\sim\pi(\theta[n])} }}\bigg[ \frac{\pi_z(a_z|s;\theta_z) A_z(s,a_z;\theta[n])}{\pi_z(a_z|s;\theta_z[n])}\bigg]- B(\theta[n]) \max_s\text{KL}(\pi_z(\cdot|s;\theta_z[n])\Vert\pi_z(\cdot|s;\theta_z)),\\
     \theta_{z'}[n+1]&=\arg\max_{\theta_{z'}\in\Theta_{z'}}\mathbb{E}_{\substack{{s\sim\rho( \theta[n])}\\{a\sim\pi(\theta[n])} }}\bigg[ \frac{\pi_{z'}(a_{z'}|s;\theta_{z'})}{\pi_{z'}(a_{z'}|s;\theta_{z'}[n])}  \frac{\pi_z(a_z|s;\theta_z[n+1])}{\pi_z(a_z|s;\theta_z[n])}A_{z',z}(s,a;\theta[n])\bigg]\\
     &\quad\quad\quad\quad\quad\quad\quad\quad\quad\quad\quad\quad\quad\quad\quad\quad\quad\quad\quad\quad- B(\theta[n]) \max_s\text{KL}(\pi_{z'}(\cdot|s;\theta_{z'}[n])\Vert\pi_{z'}(\cdot|s;\theta_{z'})),
\end{aligned}  
$}%
\end{equation}
where the importance ratio in agent $z$'s update is simpler than that of agent $z'$ due to stale sampling from the previously fixed agent, and $\Theta_z$ for $z\in\{q,\alpha\}$ denotes the parameter space which is a compact and convex subset in the Euclidean space.
To avoid explicit KL penalties and improve computational tractability, \citet{schulman2017proximal} propose clipping the importance ratio. Moreover, the local advantages in \eqref{eq:matrpo} can be replaced with reweighted global advantages (see Appendix \ref{appendix4.3algo} for details). 
Building on these, we propose a multi-agent variant that incorporates multi-timescale updates to improve stability and convergence. The resulting clipped update rules are
\begin{equation}
 \begin{aligned}
    \theta_{z}[n+1]&=\arg\max_{\theta_z\in\Theta_z} \mathcal{L}_z(\theta[n], \theta_z, \theta_{z'}[n])\text{ and } \theta_{z'}[n+1]=\arg\max_{\theta_{z'}\in \Theta_{z'}} \mathcal{L}_{z'}(\theta[n],  \theta_{z'},\theta_{z}[n+1]), \label{eq:happo0}
\end{aligned}    
\end{equation}
where the agent-wise objective is defined as
\begin{equation}\label{eq:L_z(theta_z_z')}
\resizebox{\linewidth}{!}{$%
 \begin{aligned}
    \mathcal{L}_z(\theta, \theta'_z, \theta'_{z'}) &:= \mathbb{E}_{\substack{{s\sim\rho( \theta)}\\{a\sim\pi(\theta)} }}\bigg[ \min\bigg(\frac{\pi_{z}(a_{z}|s;\theta'_{z})}{\pi_{z}(a_{z}|s;\theta_{z})}  \frac{\pi_{z'}(a_{z'}|s;\theta'_{z'})}{\pi_{z'}(a_{z'}|s;\theta_{z'})}A(s,a;\theta), 
     \Pi_{[1-\epsilon,1+\epsilon] }\big(\frac{\pi_{z}(a_{z}|s;\theta'_{z})}{\pi_{z}(a_{z}|s;\theta_{z})} \big)  \frac{\pi_{z'}(a_{z'}|s;\theta'_{z'})}{\pi_{z'}(a_{z'}|s;\theta_{z'})}A(s,a;\theta)\bigg)\bigg].
\end{aligned}    
$}%
\end{equation}
Here, $\Pi_{[x,y]}$ denotes the clip operator that bounds the importance ratio within the trust region $[x,y]$, and $\epsilon>0$ is a tolerance parameter.
More implementation details and derivations of this multi-timescale multi-agent RL algorithm are provided in Appendix \ref{appendix4.3algo}, with its full pseudocode presented in Algorithm~\ref{alg:marl}.

To highlight the core ideas underlying algorithmic convergence, we consider the setting where each policy update follows a single round of data collection. Extensions to multiple updates per sampling, as implemented in our framework, can be handled using techniques for Markovian noise in \citet{kushner2003stochastic} or \citet{borkar2024stochastic}.
The corresponding single-update-per-sample version of our multi-timescale multi-agent RL algorithm, jointly optimizing the replenishment and recommendation agents, is given by
\begin{align}
    \theta_q[n+1] &= \Pi_{\Theta_q}\big(\theta_q[n]+\varepsilon_1[n]\ G_q(\theta[n], \sigma[n]\theta_\alpha[n]+(1-\sigma[n])\theta_\alpha[n+1],\varphi[n])\big), \label{eq:drl1}
    \\
    \theta_{\alpha}[n+1] &= \Pi_{\Theta_\alpha}\big(\theta_{\alpha}[n]+\varepsilon_2[n]\ G_{\alpha}(\theta[n],\sigma[n]\theta_q[n+1]+(1-\sigma[n])\theta_q[n],\varphi[n])\big), \label{eq:drl2}
\end{align}
where $G_z(\theta, \theta'_{z'},\varphi)$ is an estimate of $\nabla_{\theta'_{z}}\mathcal{L}_z(\theta, \theta'_z, \theta'_{z'})|_{\theta'_z=\theta_{z}}$, $\sigma[n]\sim\text{Bernoulli(0.5)}$ randomly determines the update order, $\varphi[n]$ is a simplified notation that encapsulates all sources of randomness in the sampling process when evaluating \eqref{eq:L_z(theta_z_z')}, and $\Pi_\Theta$ denotes Euclidean projection onto the feasible set $\Theta$. The step sizes satisfy the timescale condition
$\varepsilon_{2}[n] = o(\varepsilon_{1}[n])$,  ensuring that two agents learn on different timescales.
Define the estimation error  for agent $z\in\{q,\alpha\}$ as $\mathcal{M}_Z[n]=G_{z}(\theta[n], \theta_{z'}[n],\varphi[n])-\nabla_{\theta_z} \mathcal{L}_z (\theta[n], \theta_z, \theta_{z'}[n])\vert_{\theta_z=\theta_z[n]}$.
We assume each agent updates in an asymptotically unbiased direction, formalized as follows:
\begin{assumption}\label{assumption:asymunbiased}
    For each agent $z\in\{q,\alpha\}$, the gradient estimation error is asymptotically negligible with bounded variance, i.e., it holds almost surely that $\mathbb{E}[\mathcal{M}_z[n] \vert \theta[0], \dots, \theta[n]] \to 0 $ as $ n \to \infty $, and there exists a constant $ K' > 0 $ such that for all $ n \geq 0 $, $\mathbb{E}[\|\mathcal{M}_z[n]\|^2 \vert \theta[0], \dots, \theta[n]] \leq K'$.
\end{assumption}

From \eqref{eq:L_z(theta_z_z')}, it follows that $\nabla_{\theta'_z} \mathcal{L}_z (\theta, \theta'_z, \theta_{z'})\vert_{\theta'_z=\theta_{z}}=\mathbb{E}_{\substack{{s\sim\rho( \theta )}\\{a\sim\pi(\theta )} }}\big[A(s,a;\theta) \nabla_{\theta_z} \log\pi_z(a_z|s;\theta_z) \big]$, which can be efficiently approximated via generalized advantage estimation (GAE; \citealt{schulman2015high}).  GAE combines Monte Carlo sampling with state-value function approximation, typically referred to as the critic in RL. As the critic improves, the bias in the gradient estimate diminishes. This is commonly achieved within an actor-critic framework \citep{konda1999actor}, where the critic is updated on a faster timescale than the agents.

The remainder of this subsection establishes the asymptotic behavior of the proposed multi-timescale RL algorithm. 
We begin by analyzing the faster timescale, which governs the replenishment agent’s updates. The following theorem characterizes its long-run behavior.
\begin{theorem}[Asymptotics for the Fast-Timescale Agent]\label{marlconv1}
Under Assumptions \ref{assumption:stepsize}-\ref{assumption:asymunbiased} and \ref{assumption:Lip2}-\ref{assumption:Lip3}, the sequence $\{\theta_\alpha[n]\}$ converges almost surely to a limit point $\overline\theta_\alpha\in\Theta_\alpha$, and the sequence $\{\theta_q[n]\}$ converges almost surely to a compact internally chain transitive invariant set $\mathcal{I}(\overline\theta_\alpha) \subset \Theta_q$ associated with the projected ODE:
\begin{align}
    \dot\theta_q(\xi) = \tilde\Pi_{\Theta_q}\big(\mathbb{E}_{\substack{s\sim\rho(\theta)\\ a\sim\pi(\theta)}}\left[ A(s,a;\theta) \nabla_{\theta_q} \log\pi_q(a_q|s;\theta_q) \right]\big|_{\theta=(\theta_q(\xi),\overline\theta_\alpha)} \big).\label{eq:marlode1}
\end{align}
For any fixed window length $\overline{\zeta}>0$, let $\langle \zeta\rangle' := \max\{n:\sum_{n'<n}\varepsilon_2[n']\le \zeta\}$ and define the Dirac‐valued path
$\mu^{[n]}_\zeta = \delta_{\theta_q[\langle \zeta[n]+\zeta \rangle']} $ over $\Theta_q$, for $\zeta\in[0,\overline{\zeta}]$ and $n\geq0$. 
Then, almost surely, there exists a subsequence of $\{\mu^{[n]}_\cdot \}$ that converges to $ \mu_\cdot^*\in \mathcal{P}(\Theta_q)^{[0, \overline{\zeta}]}$, where for each $\zeta\in[0,\overline\zeta]$, $\mu_\zeta^*$ belongs to the set of invariant probability measures $J(\overline\theta_\alpha)$ associated with the ODE (\ref{eq:marlode1}). Moreover, if $J(\overline\theta_\alpha)$ is a singleton, then $\{\mu^{[n]}_\cdot \}$ converges almost surely.
\end{theorem}

This result indicates that the replenishment agent, though subject to minor short-term fluctuations, exhibits empirical behavior that remains concentrated around stable response patterns induced by the quasi-static recommendation policy and governed by the associated projected ODE. When the ODE admits a unique globally asymptotically stable equilibrium, the agent converges almost surely to that point. Otherwise, if the ODE possesses multiple attractors or a continuum of equilibria, the agent’s trajectory remains confined to a well-defined, invariant subset of the parameter space. This subset is typically low-dimensional, forward-invariant, and dynamically stable, ensuring that the agent’s actions are consistent and predictable over time. We now proceed to analyze the slow component of the algorithm.

\begin{theorem}[Asymptotics for the Slow-Timescale Agent]\label{marlconv2}
Let $\tilde\theta_{\alpha}(\zeta)$ denote the piecewise‑constant interpolation of the sequence $\{(\sum_{n'<n}\varepsilon_2[n'],\theta_{\alpha}[n])\}$.  
Under the condition of Theorem \ref{marlconv1}, almost surely, as $\zeta\to\infty$, every limit point of $\tilde\theta_{\alpha}(\zeta+\cdot)$ in $C(\mathbb{R};\Theta_\alpha)$ is a trajectory 
 in an internally chain transitive invariant set 
of the differential inclusion
$\dot \theta_{\alpha}(\zeta)\in \tilde\Pi_{\Theta_\alpha}\big(H\bigl(\theta_{\alpha}(\zeta)\bigr)\big)$, where $H(\theta_\alpha):=\bigl\{\int_{\Theta_q} \mathbb{E}_{\substack{{s\sim\rho( \theta )}\\{a\sim\pi(\theta )} }}\big[A(s,a;\theta) \nabla_{\theta_\alpha} \log\pi_\alpha(a_\alpha|s;\theta_\alpha) \big]\,\mu(d\theta_q):\;\mu\in J(\theta_\alpha)\bigr\}$.
\end{theorem}

Theorem \ref{marlconv2} implies that even in nonconvex settings,  the slow-update agent evolves within a dynamically stable region aligned with the long-run behavior of the fast-update component. In favorable cases, such as when objective functions are appropriately regularized, the invariant set reduces to a singleton, yielding convergence to a unique solution. More generally, the algorithm stabilizes within a compact, well-structured subset of policy parameters where both agents act nearly optimally and exhibit resilience to unilateral deviations. This characterization encompasses not only strict equilibria but also neighboring stable configurations that the system revisits recurrently, offering a nuanced and practically relevant view of long-term coordination in dynamic, competitive settings.
Unlike classical multi-timescale SA theories, which typically rely on the almost sure convergence of the fast-timescale recursion to a unique fixed point before embedding it into the slower update, Theorem \ref{marlconv2} relaxes this stringent requirement by aggregating multiple possible limits probabilistically, providing theoretical support for complex dynamic coordination.

Finally, we reiterate that the proposed multi-timescale update mechanism resonates with emerging insights from both neuroscience and AI. Notably, Geoffrey Hinton, recipient of the Nobel Price in Physics and Turing Award, has suggested that different regions of the brain may operate on distinct learning rates, advocating for ML algorithms that mirror such biologically inspired architectures. Embedding this principle into the structure of policy updates is thus not only a practical design tailored to the coordination challenges at hand, but also reflects a broader methodological development aligned with current trends in AI research.

\section{Simulation Study}\label{sec:Numerical Experiments}

This section presents simulation experiments that evaluate the implementation and performance of the proposed multi-timescale multi-agent (MTMA) RL approach. 
In Section \ref{subsec5.1}, we demonstrate that multi-timescale updates outperform single-timescale counterparts in training efficiency, convergence stability, and overall performance. Moreover, the multi-agent architecture delivers significant improvements relative to the single-agent counterpart.
Section \ref{subsec5.2} visualizes agent behavior alongside the  state dynamics of the joint inventory-recommendation system, revealing strong alignment and effective policy adaptation. These patterns reinforce the managerial insights developed in Section~\ref{Theoretical Model} and provide validation of the proposed RL framework. In Section \ref{subsec5.3}, we quantify the benefits of coordination by comparing cooperative and isolated training paradigms. The results demonstrate notable performance improvements for both individual departments (agents) and the overall platform under cooperative learning. We further conduct extensive supplementary experiments in Section~\ref{Supplementary Materials to Numerical Experiments} to validate the robustness and generalizability of our algorithm, as well as to examine the behavioral plausibility of the trained RL agents.

The simulation study is conducted in an environment with $N=5$ products and $M=20$ customers. Each training episode spans a default horizon of $T=100$ steps.
Customer behavior follows the model specified in Section \ref{Theoretical Model}, with purchase willingness evolving as $R_t^{i,j} = \eta R_{t-1}^{i,j} + (\bar{R} - \eta R_{t-1}^{i,j}) \alpha_t^{i,j}$, reflecting diminishing marginal effects of recommendation. Purchase decisions are binary and determined by a softmax function over willingness scores: $\gamma_t^{i,j} = {\exp(R_t^{i,j})}/{\sum_{i'=1}^N \exp(R_t^{i',j})}$.
The firm’s per-period profit is given by
$P_t^i = p_{\text{out}} S_t^i - p_{\text{in}} q_t^i - h I_t^i - b U_t^i,$ where $p_{\text{out}}$ and $p_{\text{in}}$ denote selling and procurement prices, respectively. Recommendation cost is proportional to recommendation intensity.
At the beginning of each episode, both purchase willingness and inventory levels are randomly initialized.

All agents adopt a common architecture comprising a four-layer MLP, followed by two RNN layers and a final linear output layer. During training, actions are sampled from a Gaussian distribution centered at the network output, then passed through a hyperbolic tangent or rounding transformation to ensure feasibility. During evaluation, sampling is disabled and the mean is used as the deterministic action.
In the multi-agent setting, separate networks are used for the inventory and recommendation agents. The inventory agent employs hidden-layer width 128, while the recommendation agent, which handles more complex actions, uses wider layers with 384 neurons. In the single-agent baseline, both functions are merged into one agent with  width 512, equal to the sum of the two. We implement multi-timescale updates with diminishing step sizes of the form:
$$
\varepsilon_i[n] = \epsilon_i \left(\frac{0.1 N_I}{n + 0.1 N_I}\right)^{p_i}, \quad \text{for } i = 1, 2 \text{ and } n \geq 0,
$$
where $N_I$ is the total number of training iterations, and $(\epsilon_i, p_i)$ specifies the initial step size and decay rate. To meet the timescale conditions in Assumption \ref{assumption:stepsize}, the decay rates are chosen to satisfy $0.5<p_1<p_2\leq1$. In our implementation, we set $(\epsilon_1, p_1) = (10^{-3}, 0.75)$ for the fast-updating agent and $(\epsilon_2, p_2) = (2 \times 10^{-5}, 0.99)$ for the slow-updating agent.
We employ the centralized training with decentralized execution framework described in Section \ref{subsec4.2:cooperation_markov_game}, using a centralized critic to approximate the state-value function and compute advantage estimates. The critic shares the agent architecture but employs a wider hidden layer of width 512. Following the actor-critic design \citep{konda1999actor}, it is updated using an even faster step size, with parameters $(\epsilon_0, p_0) = (10^{-3}, 0.51)$. Notably, the critic is used exclusively during training and is discarded during execution, where each agent independently determines its action.
 
All simulations are conducted on a Linux server equipped with four NVIDIA RTX 4090 GPUs and two Intel\textsuperscript{\textregistered} Xeon\textsuperscript{\textregistered} Platinum 8352V CPUs (2.10GHz) with parallel training and evaluation.

\subsection{Efficiency of Multi-Timescale Multi-Agent RL}\label{subsec5.1}

\begin{figure}[htb]
\centering 
\subfigure[Multi- vs. single-timescale multi-agent RL.]{
\label{fig:mtcompare}
\includegraphics[trim=0cm 0.5cm 0cm 0.5cm, width=0.4496\linewidth]{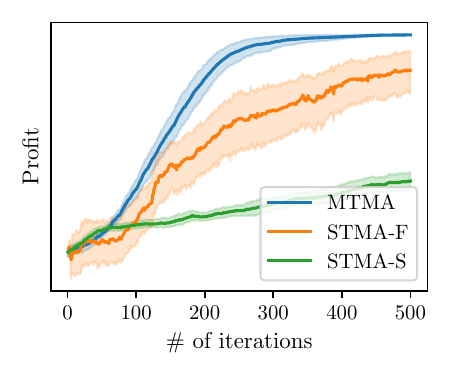}}
\hspace{1cm}
\subfigure[Multi- vs. single-agent RL.]{
\label{fig:macompare}
\includegraphics[trim=0cm 0.5cm 0cm 0.5cm, width=0.4496\linewidth]{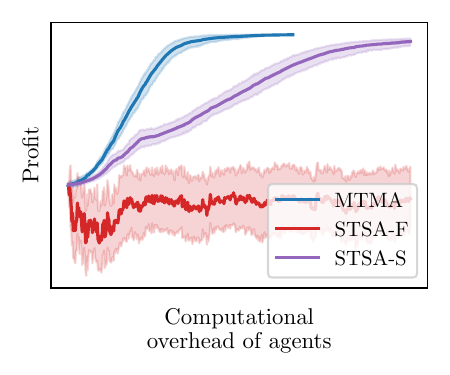}}
\caption{Learning curves under different algorithmic configurations. MTMA denotes the proposed multi-timescale multi-agent RL approach; STMA-F and STMA-S are single-timescale multi-agent baselines using shared fast or slow step sizes, respectively, matching those used in MTMA; STSA-F and STSA-S refer to single-agent baselines trained with the same fast or slow step sizes as in MTMA. Curves are averaged over 20 independent runs, with solid lines indicating the mean and shaded areas showing $95\%$ confidence intervals.
}
\label{fig:mtscompare}
\end{figure}

As discussed in Section~\ref {subsec:connecting_theory_to_algorithms}, the structural asymmetry between inventory control and recommendation tasks motivates the use of distinct learning rates for the respective agents. This design aligns the learning dynamics with the complexity of each task. Figure~\ref {fig:mtcompare} validates this claim by comparing learning curves in terms of episodic profit. On one hand, the proposed MTMA algorithm achieves rapid convergence within a moderate number of iterations, whereas single-timescale baselines require significantly more iterations to converge, especially the variant with slower updates.
This underscores the necessity of the proposed multi-timescale approach. 
On the other hand, the learning curves for MTMA exhibit noticeably narrower confidence intervals, indicating enhanced stability. In particular, aggressively applying fast uniform updates could leads to high variance and unstable performance, as observed in our numerical experiments. Overall, the multi-timescale design achieves both training efficiency and robustness.

To further evaluate the benefits of a multi-agent learning framework, we compare the multi-agent and single-agent architectures in Figure~\ref{fig:macompare}. 
As discussed in Section \ref{subsec4.2:cooperation_markov_game}, consolidating all decisions into a single agent may introduce substantial computational inefficiency. 
Due to the increased model size in the single-agent setting, applying a fast step size results in complete learning failure, with performance fluctuating around a low level and exhibiting high variance. This also highlights the necessity of adopting a multi-agent approach, without it, learning quality could deteriorate significantly. While the slow-update variant achieves convergence at last, it does so at a substantially higher computational cost, at least twice that of the multi-agent configuration.  Although this example involves only two decision channels, the computational gap would grow rapidly with more agents, further amplifying the advantages of the multi-agent design. These findings underscore that the proposed multi-agent architecture is essential for effectively addressing the coordination problem, enabling both computational scalability and superior performance.
To further substantiate our conclusions, we conduct a series of robustness checks on the algorithmic configurations, as detailed in Appendix~\ref{appendix5.2}.

\subsection{Behavioral Analysis and Validation of Trained Agents}\label{subsec5.2}

To validate the effectiveness and interpretability of the RL decisions, we visualize agent behavior details, assessing their consistency with the theoretical insights established in Section~\ref{Theoretical Model}. The agents analyzed here are drawn from the MTMA-trained model in prior experiments. To highlight behavioral regularities, we extend the episode length and discard early steps as a burn-in phase to mitigate potential cold-start effects.

\begin{figure}[htb]
\centering 
\includegraphics[trim=0cm 0.5cm 0cm 0.5cm, width=0.896\linewidth]{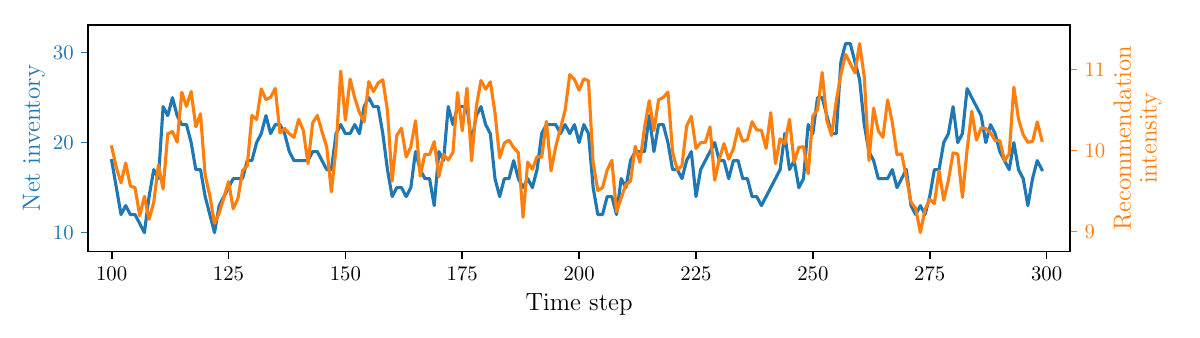}
\caption{Temporal dynamics of net inventory and recommendation intensity for a representative product.
}\label{fig:6a}
\end{figure}

We first examine the alignment between inventory decisions and recommendation intensity. Since replenishment choices are shaped by both current stock levels and anticipated demand, we use net inventory---defined as the sum of current replenishment, on-hand stock, and in-transit inventory, minus stockouts---as a proxy for effective inventory positioning.
Recommendation intensity is measured as the total recommendation effort aggregated across all 20 customers.
Figure~\ref{fig:6a} plots the temporal evolution of net inventory and recommendation intensity for a representative product. The two series exhibit highly synchronous pattens, suggesting that the system not only adjusts inventory levels in anticipation of future demand but also adapts its recommendation strategy in response to current inventory conditions.
This observation aligns with the analytical insights in Proposition \ref{prop:ordering_to_marketing} and partially in Proposition \ref{th:recommendation}, albeit under simplified model assumptions.

\begin{figure}[htb]
\centering 
\includegraphics[trim=0cm 0cm 0cm 1.5cm, clip, width=0.9\linewidth]{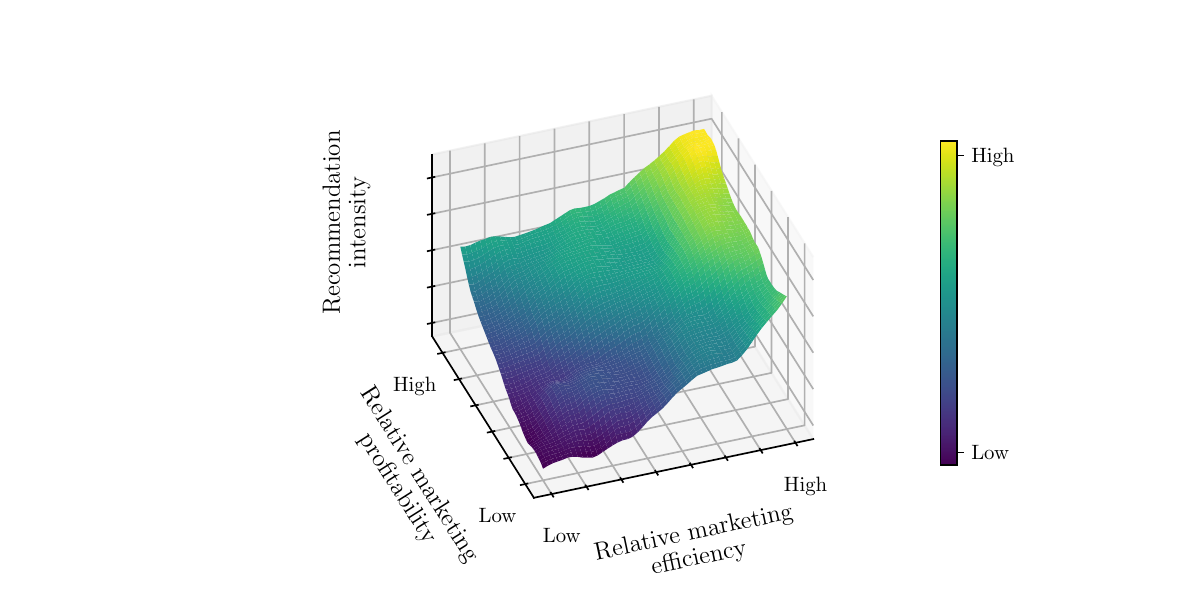}
\caption{Recommendation intensity as a function of relative marketing efficiency and profitability.
}\label{fig7}
\end{figure}

To better understand the recommendation behavior learned by the RL agent, we analyze the functional structure of the trained decision network, as theoretically analyzed in Proposition \ref{th:recommendation}. We first extend the definitions of RME and RMP in Proposition \ref{th:recommendation} to the more general setting considered here. In essence, these two metrics quantify, respectively, the comparative ease of increasing a customer’s purchase intention and the marginal contribution of such an increase to platform revenue, which is closely linked to inventory levels. Following this idea, we define the RME of recommending product $i$ to customer $j$ at time $t$ as $\operatorname{RME}_t^{i,j} =  (\bar{R} - R_t^{i,j} ) - \sum_{i'\neq i}  (\bar{R} - R_t^{i',j} )$, where a larger value implies greater relative potential for enhancing product $i$'s appeal. The RMP of product $i$ at time $t$ is defined as $\operatorname{RMP}_t^i =  {I_t^i} - {\sum_{i'\neq i} I_t^{i'}}$, which reflects its relative inventory exposure. Figure \ref{fig7} presents the RL agent's recommendation policy against RME and RMP. In general, the results are consistent with the guiding principles in Proposition \ref{th:recommendation}: recommendations intensify for products exhibiting both high relative efficiency and profitability. Notably, the agent will selectively target a subset of products for strong promotion, avoiding resource waste and confirming the rationality behind its learned strategy.

\begin{figure}[htb]
\centering 
\subfigure[Recommendation response to exogenous demand.]{
\includegraphics[trim=0cm 0.5cm 0cm 0.5cm, width=0.476\linewidth]{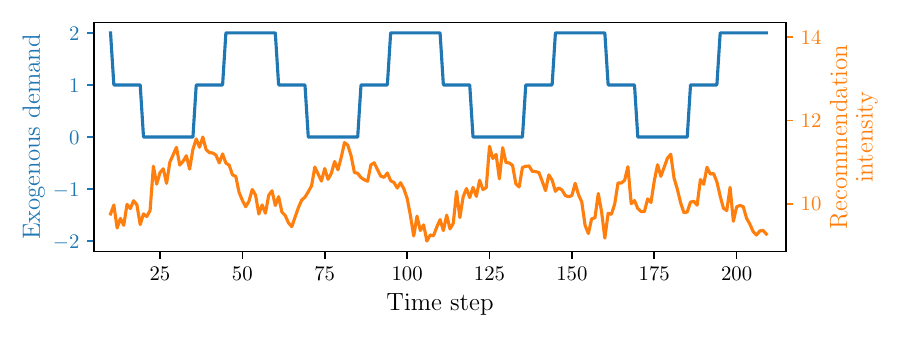}\label{fig8a}}
\hfill
\subfigure[Replenishment response to purchase willingness shocks.]{
\includegraphics[trim=0cm 0.5cm 0cm 0.5cm, width=0.496\linewidth]{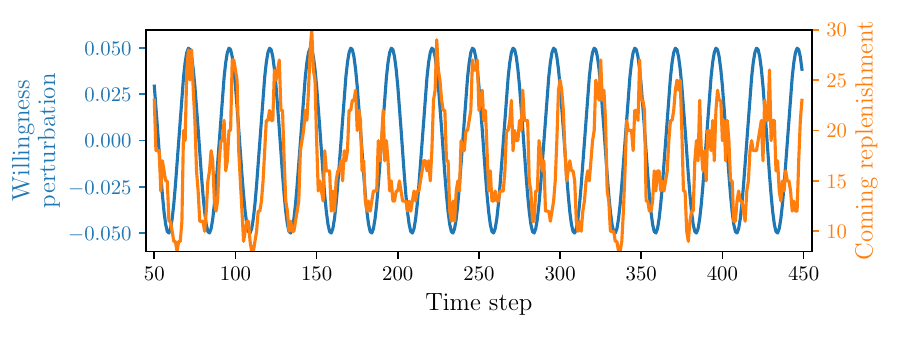}\label{fig8b}}
\caption{Behavioral validation of demand smoothing and adaptive ordering under exogenous perturbations.
}
\end{figure}

To verify the intuition from Propositions \ref{prop:demand_smoothing} and \ref{prop:adaptive_ordering}, we introduce exogenous perturbations to the system to investigate the agents' responses. 
In Figure~\ref{fig8a}, we apply rounded sinusoidal shocks to baseline customer demand, using a shared period but staggered phases across products. The resulting recommendation intensity (orange line) exhibits a distinct counter-cyclic pattern: following demand surges that raise inventory levels, the agent reduces recommendation effort; during demand troughs, recommendation intensity increases. This inverse behavior suggests that the agent actively smooths demand over time, consistent with the mechanism predicted in Proposition~\ref{prop:demand_smoothing}.
In a separate experiment shown in Figure~\ref{fig8b}, we impose similar periodic fluctuations on customer purchase willingness and track total order quantities placed during the lead time following each shock. The results show clear synchronization between ordering behavior and external shocks, indicating that the inventory agent adapts its actions in response to changes in purchase intent, as anticipated in Proposition~\ref{prop:adaptive_ordering}.

\subsection{Benefits of Cross-functional Coordination}\label{subsec5.3}

Finally, we evaluate the platform-level benefits enabled by our MTMA RL framework and highlight the importance of cross-channel coordination in reducing costs and enhancing revenues. Specifically, we compare a coordinated setting---where both agents jointly optimize the overall platform revenue---with a fully isolated setting that reflects common platform practices, where the replenishment and recommendation agents operate independently as separate teams. In the isolated configuration, each agent is evaluated based on its own departmental key performance indicator (KPI): the inventory agent minimizes inventory-related costs (including holding and stockout penalties), while the recommendation agent maximizes net sales revenue after accounting for recommendation expenditures.
To further dissect the value of coordination, we construct two intermediate benchmarks that isolate the marginal impact of each agent's cooperation. In the Isolated-Replenishment scenario, only the replenishment agent operates under its departmental objective, while the recommendation agent remains aligned with the platform-level revenue KPI. In contrast, the Isolated-Recommendation scenario features a self-interested recommendation agent, while the replenishment agent remains cooperative. For fair comparison, we initialize all variants from the fully cooperative model and selectively retrain individual agents using their respective departmental KPIs. For all settings involving non-cooperative behavior, we employ separate critics for each agent to prevent unintended value sharing during training. Each configuration is independently replicated over 20 simulation runs, and training continues for sufficiently many iterations to ensure convergence.

\begin{table}[htbp]
\centering
\caption{Performance comparison under varying degrees of agent cooperation.}
\label{tab:5.3}
\resizebox{\textwidth}{!}{  
\begin{tabular}{>{\centering\arraybackslash}m{4.5cm}||>{\centering\arraybackslash}m{3.5cm}>{\centering\arraybackslash}m{3.5cm}|>{\centering\arraybackslash}m{3.5cm}>{\centering\arraybackslash}m{3.5cm}} 
\toprule
Setting & \textbf{Cooperative} & Isolated & Isolated replenishment  & Isolated recommendation  \\
\midrule
Aver. Total Profit &  \textbf{584.32}  & 157.82 & 352.37 & 476.89      \\
95\% Interval & \textbf{(561.88, 606.75)}   & (-139.08, 454.71)  & (266.77, 437.98) & (418.87, 534.91)       \\
\midrule
Aver. Inventory Cost &  \textbf{289.34}  & 681.29  & 505.67 & 391.55     \\
95\% Interval & \textbf{(267.77, 310.91)}   & (416.09, 946.48)   & (426.19, 585.14) & (336.70, 446.41)     \\
\midrule
Aver. Marketing Revenue &  \textbf{873.65} & 839.10 & 858.04 & 868.44      \\
95\% Interval & \textbf{(872.19, 875.12)}  & (806.98, 871.22) & (851.44, 864.64)   & (865.10, 871.78)      \\
\bottomrule
\end{tabular}
}
\label{tab:example}
\end{table}

As shown in Table \ref{tab:5.3}, the coordinated mode consistently outperforms the independent-channel benchmarks across both individual KPIs and overall platform profit. While some improvement is expected, the magnitude of gain is notably substantial. Additionally, the narrower confidence intervals in the cooperative configurations suggest reduced performance volatility, indicating that coordination enhances operational stability. Given that our experimental setup involves a relatively small number of products and customers, we anticipate even greater relative benefits at scale. This reinforces the central motivation of the study and again demonstrates the effectiveness of the proposed MTMA RL framework. Furthermore, the decomposition of gains using partially independent settings reveals that coordination contributes meaningfully to both inventory cost reduction and marketing revenue enhancement. Although the cost savings appear more pronounced in our results, this asymmetry is likely attributable to features specific to the simulation environment. We also examine agent behaviors across different operational modes in Section \ref{appendix5.1}, and validate the robustness of the comparisons in Section \ref{appendix5.3}.

\section{Conclusions}\label{sec:conclusion}

This study contributes to both the theory and practice of cross-functional decision-making by proposing an RL framework tailored for complex organizational coordination. Focusing on the interplay between inventory management and product recommendation, we develop a theoretical model that captures the structural asymmetries and interdependencies between these functions. Guided by analytical insights into cross-sectional and temporal coordination mechanisms, we introduce a multi-agent, multi-timescale RL architecture that assigns distinct roles, learning dynamics, and update rates to departmental agents. The proposed framework not only enhances training efficiency and policy interpretability but also mirrors realistic organizational structures. Extensive numerical experiments confirm its performance advantages in terms of profit, learning stability, and behavioral alignment. More broadly, this work illustrates how modern RL techniques can be adapted to support scalable, modular, and interpretable decision-making in multi-departmental environments. Future directions include refining behavioral models, extending to multi-tier coordination, and validating deployment in operational platforms.



\setlength{\bibsep}{0pt}
{\small\bibliographystyle{informs2014} 
\bibliography{ref} 
}




\ECSwitch
\makeatletter
\renewcommand{\theHsection}{EC.\arabic{section}}
\renewcommand{\theHsubsection}{EC.\arabic{section}.\arabic{subsection}}
\renewcommand{\theHequation}{EC.\arabic{equation}}
\renewcommand{\theHtheorem}{EC.\arabic{theorem}}
\renewcommand{\theHlemma}{EC.\arabic{lemma}}
\renewcommand{\theHassumption}{EC.\arabic{assumption}}
\renewcommand{\theHproposition}{EC.\arabic{proposition}}
\renewcommand{\theHfigure}{EC.\arabic{figure}}
\renewcommand{\theHtable}{EC.\arabic{table}}
\renewcommand{\thealgorithm}{EC.\arabic{algorithm}}
\makeatother
\ECHead{\centering Online Appendix to\\ ``Coordinating Replenishment and Recommendation via Deep Reinforcement Learning"}

\section{Supplementary Materials to Section \ref{Theoretical Model}}

\proof{Proof of Proposition \ref{prop:ordering_to_marketing}.}

Recalling that $U^i = [D^{i}-q^{i} ]^{+}$ and $I^i = [q^{i}-D^{i} ]^{+}$, the expected profit for the $i$-th product can then be expressed as
\begin{align}
\mathbb{E}[P^i] 
=&\ (p+h) \mathbb{E}[D^i] - h q^i - (p+ h+ b)  \mathbb{E}[[D^{i}-q^{i} ]^{+}].
\label{eq:profit_mean}
\end{align}
To analyze the structural properties of the profit function, we consider its first-order derivative with respect to the replenishment quantity $q^i$:
\begin{align*}
\frac{d}{d q^i}\mathbb{E}[P^i] =&\  -h + (p+ h+ b) P(D^i\geq q^i).
\end{align*}
Since $p+ h+ b>0$ and the tail probability $\text{P}(D^i\geq q^i)$ is non-increasing in $q^i$,  
${d\mathcal{L}}/{d q^i}={d\mathbb{E}[P^i]}/{d q^i}$ is non-increasing in $q^i$ and is also independent of $q^{i'}$ for $i'\neq i$. Thus, the platform's total expected profit $\mathcal{L}$ is concave with respect to the replenishment vector $(q^1,q^2)$. Moreover, the optimal replenishment quantity admits the classical characterization $q^{i,*} = \sup\{q^i\in[0,\overline{q}]: \text{P}(D^i\geq q^i)\geq \frac{h}{p+ h+ b} \}$. 
With \eqref{equation:willingness_special} and \eqref{logit_model_cross_product}, we can 
differentiate $\gamma^i$ with respect to the recommendation intensities, 
\begin{align}
    \frac{d \gamma^{i}}{d\alpha^i} = (\overline{R}-R^i_0)\gamma^i (1-\gamma^i)> 0,\ \text{and }
    \frac{d \gamma^{i}}{d\alpha^{i'}} = -(\overline{R}-R^i_0) \gamma^i \gamma^{i'}< 0,\ \text{for }i'\neq I. \label{eq:ec1}
\end{align}
Therefore, increasing $\alpha^i$ raises the purchase probability $\gamma^i$, while increasing $\alpha^{i'}$ reduces it due to cross-product competition. 
Provided that the tail probability $\mathbb{P}(D^i \ge q^i)$ is increasing in $\gamma^i$, it follows that the optimal replenishment quantity $q^{i,*}$ is increasing in $\alpha^i$ and decreasing in $\alpha^{i'}$, completing the argument.
\Halmos
\endproof

\proof{Proof of Proposition \ref{th:recommendation}.}

By differentiating the profit expression in (\ref{eq:profit_mean}) with respect to the purchase probability $\gamma^i$, and exploiting the structure of Bernoulli demand, we obtain
\begin{align*}
\frac{d\mathbb{E}[P^i]}{d \gamma^{i}} = (p + h) - (p + h + b)\, [1 - q^i]^+.
\end{align*}
Moreover, since the profit of product $i$ is independent of $\gamma^{i'}$ for $i' \ne i$, it follows that ${d\mathbb{E}[P^i]}/{d \gamma^{i'}} = 0$.
Combining this with the sensitivity of the purchase probability derived in \eqref{eq:ec1}, the marginal impact of recommendation intensity $\alpha^i$ on the total expected
profit $\mathcal{L}$ can be expressed as
\begin{align}
\frac{d \mathcal{L}}{d \alpha^i} 
= (p + h + b)(\overline{R} - R_0^i)\, \big([1 - q^{i'}]^+ - [1 - q^i]^+\big) \gamma^i \gamma^{i'} - r. \label{ec:2}
\end{align}
If the replenishment quantity $q^i$ for product $i$ is larger than that of product $i'$, then the marginal benefit of recommending product $i$ is also higher, as indicated by the inequality $-[1 - q^{i'}]^+ \leq -[1 - q^i]^+$. In this case, we have ${d \mathcal{L}}/{d \alpha^{i'}} \leq 0$ and the optimal recommendation policy for product $i'$ should be $\alpha^{i',*} = 0$, implying that product $i'$ receives no recommendation.

Observe that the optimal policy never involves recommending both products simultaneously due to the demand model \eqref{logit_model_cross_product}. 
Without loss of generality, we consider the recommendation decision for the first product. For notational convenience, we recall the following definitions introduced in the main text:
$$
\operatorname{RME} := (\overline{R} - R_0^1) - (\overline{R} - R_0^2), \quad
\operatorname{RMP} := (p + h + b)\, \big([1 - q^2]^+ - [1 - q^1]^+\big), \quad
\zeta := \frac{r}{\overline{R} - R_0^1}.
$$
Let $\bar{\gamma}^1 := \gamma^1 \big|_{\alpha^2 = 0}$ denote the logit-based purchase probability for product 1 when product 2 receives no recommendation. Under the logit structure, $\bar{\gamma}^1$ lies within the following interval:
$$
\bar{\gamma}^1 \in [\gamma_b,\gamma_h]:=\left[\frac{1}{1 + \exp(\operatorname{RME})},\ \frac{1}{1 + \exp(R_0^2 - \overline{R})}\right].
$$
Substituting it into the gradient expression~\eqref{ec:2}, we obtain
\begin{align}
    \frac{d \mathcal{L}}{d \alpha^1} \bigg|_{\alpha^2 = 0} 
= \operatorname{RMP} \cdot (\overline{R} - R_0^1) \cdot \bar{\gamma}^1 (1 - \bar{\gamma}^1) - r.\label{ec:3}
\end{align}

We first consider the case where $\operatorname{RMP} \le 4\zeta$. Since $\bar{\gamma}^1 (1 - \bar{\gamma}^1)\leq 1/4$, the entire expression in~\eqref{ec:3} remains non-positive across the feasible range. This implies that any increase in $\alpha^1$ reduces expected profit, and hence the platform optimally sets $\alpha^{1,*}=0$, allocating no recommendation effort to product 1.
This result refines the prior analysis: not only must product 1 exhibit higher standalone profitability, but the relative margin must exceed a threshold to justify investment in recommendation.

When $\operatorname{RMP} > 4\zeta$, the first-order optimality condition yields two candidate roots for the conditional purchase probability $\bar{\gamma}^1$, given by
\begin{equation}\label{solution_gamma_pn}
    \bar{\gamma}^1_{\pm} := \frac{1}{2} \pm \frac{1}{2} \sqrt{1 - \frac{4\zeta}{\operatorname{RMP}}}.
\end{equation}
To determine whether a positive recommendation intensity should be assigned to product 1, we compare these two candidates within the feasible domain of $\bar{\gamma}^1$. Since $\gamma^1$ follows a logit form and $\alpha^{2,*} = 0$, this feasible domain for $\bar{\gamma}^1$ is bounded by
$$
\gamma_b := \frac{1}{1 + \exp(\operatorname{RME})}, \quad 
\gamma_h := \frac{1}{1 + \exp(R_0^2 - \overline{R})},
$$
where $\gamma_b$ and $\gamma_h$ represent the lower and upper bounds for $\bar{\gamma}^1$ based on the relative marketing efficiency and the full budget allocation to product 1.

We next examine three possible configurations of the feasible range for $\bar{\gamma}^1$, each yielding a unique maximizer of the overall profit. First, if both bounds $\gamma_b$ and $\gamma_h$ lie above $\bar{\gamma}^1_{+}$ (see \eqref{solution_gamma_pn}), i.e., $\bar{\gamma}^1_{+}<\gamma_b\leq\gamma_h$, then the profit function is strictly decreasing over the entire feasible interval. This occurs when
 $\operatorname{RMP}< \frac{\zeta (1+\exp(\operatorname{RME}))^2}{\exp(\operatorname{RME})}$ and $\operatorname{RME}<0$. 
 In this case, the optimal recommendation intensity is zero: $\alpha^{1,*} = 0$.
Second, if the whole feasible range lies between the two roots, i.e., $\bar{\gamma}^1_{-} < \gamma_b \leq \gamma_h < \bar{\gamma}^1_{+}$, then the profit function is strictly increasing across the interval. Increasing $\alpha^1$ always improves profit, and the optimum is achieved at the upper bound $\bar{\gamma}^1 = \gamma_h$.
Third, if the feasible interval intersects the region between the two roots, i.e., $\bar{\gamma}^1_{-} < \gamma_b < \bar{\gamma}^1_{+} < \gamma_h$, the profit function is single-peaked within the feasible range. The maximum is attained at the peak, $\bar{\gamma}^{1,*} = \bar{\gamma}^1_{+}$. 
In both of the latter two cases, a positive recommendation is made to product 1, while product 2 receives none. The optimal purchase probability can be uniformly expressed as $\bar{\gamma}^{1,*} = \min\{\gamma_h,\ \bar{\gamma}^1_{+}\}$, and the condition for recommending product 1 simplifies to $\operatorname{RMP}\geq \frac{\zeta (1+\exp(\operatorname{RME}))^2}{\exp(\operatorname{RME})}$.
Applying the inverse of the logit function, the optimal recommendation intensity admits the closed-form expression:
$$
\alpha^{1,*} = \min\left\{\frac{\zeta}{r} \left[ 2 \cdot \mathrm{artanh}\left( \sqrt{1 - \frac{4\zeta}{\operatorname{RMP}}} \right) + \operatorname{RME} \right], 1\right\}.
$$
This expression highlights how the optimal recommendation decision depends jointly on the relative marketing efficiency and profitability gap between the two products. Recommendation is warranted only when both factors are sufficiently favorable.

Under the condition that product 1 is recommended, the threshold inequality $\operatorname{RMP} \geq \frac{\zeta (1+\exp(\operatorname{RME}))^2}{\exp(\operatorname{RME})}> 4\zeta$ must hold. The sensitivity of the optimal recommendation intensity $\alpha^{1,*}$ with respect to the defined metrics RME and RMP can be expressed as
\begin{align*}
    \frac{\partial \alpha^{1,*}}{\partial \operatorname{RME}} = \frac{\zeta}{r}\cdot\mathbf{1}\{\alpha^{1,*}<1\} \geq 0,\quad \frac{\partial \alpha^{1,*}}{\partial \operatorname{RMP}} &= \frac{\zeta}{r} \cdot \frac{\mathbf{1}\{\alpha^{1,*}<1\}}{\sqrt{\operatorname{RMP}} \cdot \sqrt{\operatorname{RMP} - 4\zeta}} \geq 0.
\end{align*}
These expressions confirm that the optimal recommendation intensity increases monotonically with both $\operatorname{RME}$ and $\operatorname{RMP}$. 
\Halmos
\endproof

\proof{Proof of Proposition \ref{prop:demand_smoothing}.}

We define the information filtration up to period $t$ as $\mathcal{F}_t:=\{(q_{t'}, \alpha_{t'}, S_{t'}, I_{t'}, U_{t'}, R_{t'})\}_{{t'}=1}^t$. Hence, all decisions and state variables up to and including period $t$ are $\mathcal{F}_t$-measurable.
Under the classical backlog convention, material flow dynamics are governed by:
$S_t = D_t + U_{t-1} -U_t,\ 
    U_t = \big[D_t + U_{t-1} -I_{t-1}-q_t\big]^{+},\ 
    I_t = \big[I_{t-1}+q_t-D_t - U_{t-1}\big]^{+}.$
Analogous to the derivation of (\ref{eq:profit_mean}), we can express the expected profit in the first period as
\begin{align*}
\mathbb{E}[P_1] & =  (p+h) \mathbb{E}[D_1] - h q_1 - (p+ h+ b)  \mathbb{E}[[D_1-q_1 ]^{+}].
\end{align*}
For the second period, we compute the expected profit conditional on the first-period filtration $\mathcal{F}_1$. Substituting the above dynamics yields the following conditional expected profit:
\begin{align*}
    \mathbb{E}\big[ P_2 |\mathcal{F}_1\big] 
    = (p+h) (\mathbb{E}\big[ D_2 |\mathcal{F}_1\big] +U_1) - h I_1- h q_2 -(p+h+b) \mathbb{E}\big[ [D_2 + U_1 -I_1-q_2 ]^+|\mathcal{F}_1\big].
\end{align*}
Given that demand $D_t$ follows a Bernoulli distribution with success probability $\gamma_t$, the conditional expectations admit closed-form expressions. Summing across two periods, the total expected profit function $\mathcal{L}$ becomes:
\begin{align*}
\mathcal{L} =&
\gamma_1 \left(
(p + h) + h q_1 - b [1 - q_1]^+ - (p + h + b)[1 - q_1 - q_2]^+
\right) + \gamma_2 \left[
(p + h) - (p + h + b)[1 - q_1 - q_2]^+
\right] \\
&+ \gamma_1 \gamma_2 
(p + h + b)\left(2 [1 - q_1 - q_2]^+ - [2 - q_1 - q_2]^+\right)
 - h( 2 q_1 + q_2) - r(\alpha_1+\alpha_2).
\end{align*}
With the Bernoulli demand structure, the reasonable order quantities naturally lie in the set $\mathcal{Q}^-:=\{(q_1,q_2):q_1\in[0,1], q_2\in[0,2-q_1]\}$. Since the objective function $\mathcal{L}$ is a continuous piecewise linear function over this compact polyhedral domain, standard results from linear programming theory ensure that the maximum is attained at an extreme point of $\mathcal{Q}$, all of which correspond to integer-valued combinations of $(q_1,q_2)$. This insight allows us to restrict attention to the discrete set $\mathcal{Q}:=\mathcal{Q}^-\cap \mathbb{N}^2$, thereby simplifying the analysis. Specifically, we can obtain
\begin{align}
    \mathcal{L}=&\ \gamma_1\big(p+2h- (h+b)   \mathbf{1}_{\{ q_1= 0\}} - (p+h+b)  \mathbf{1}_{\{q_1 +q_2= 0\}}\big)+ \gamma_2 \big((p+h)-(p+h+b)\mathbf{1}_{\{q_1 +q_2 = 0\}}\big)\nonumber\\
    & + \gamma_1\gamma_2 \big(-(p+h+b)\mathbf{1}_{\{q_1 +q_2 = 1\}}\big) - 2h q_1 - h q_2 - r(\alpha_1+\alpha_2).\label{eq:orl_profit_multi_t}
\end{align}
This formulation removes the piecewise structure from $\mathcal{L}$, enabling direct analytical comparisons across replenishment and recommendation policies.
The subsequent analysis is based on this discrete representation.

Before analyzing the profit expression (\ref{eq:orl_profit_multi_t}), we first derive the partial derivatives of the purchasing probabilities $\gamma_t$ with respect to the recommendation intensities $\alpha_t$. 
By direct differentiation, we have
\begin{equation}\label{eq:diff}
\begin{aligned}
    \frac{\partial \gamma_t}{\partial \alpha_t} = \gamma_t (1 - \gamma_t)  (\overline{R} - \eta R_{t-1}),\ \text{for } t=1,2,\ \text{and}\ \quad
    \frac{\partial \gamma_2}{\partial \alpha_1} = \gamma_2 (1 - \gamma_2)  (\eta - \alpha_2 \eta)  (\overline{R} - \eta R_0).
\end{aligned}\end{equation}
Since each $R_t$ is upper bounded by $\overline{R}$ based on its dynamics, the right-hand sides in (\ref{eq:diff}) are non-negative for all $\alpha_t\in [0,1]$. The purchase probability $\gamma_t$ is monotonically increasing with respect to all relevant recommendation intensities
$\{\alpha_{t'}\}_{t'=1}^t$ within the feasible domain.

Within each subset of candidate solutions with a fixed total replenishment quantity $\sum_t q_t\in\{1,2\}$, we observe that the marginal effect of $\alpha_2$ on $\mathcal{L}$ is invariant to how the total order is distributed across the two periods. In contrast, the sensitivity of $\mathcal{L}$ with respect to $\alpha_1$ depends on the temporal allocation of replenishment and exhibits the following monotonicity:
\begin{align*}
    \frac{\partial \mathcal{L}}{\partial \alpha_1}\big|_{(q_1,q_2)=(0,1)}&\leq \frac{\partial \mathcal{L}}{\partial \alpha_1}\big|_{(q_1,q_2)=(1,0)},\quad 
    \frac{\partial \mathcal{L}}{\partial \alpha_1}\big|_{(q_1,q_2)=(0,2)}\leq \frac{\partial \mathcal{L}}{\partial \alpha_1}\big|_{(q_1,q_2)=(1,1)}.
\end{align*}
These inequalities indicate that, for a fixed total replenishment, front-loading orders (i.e., increasing $q_1$) enhances the marginal return of early-period recommendation $\alpha_1$. As a result, the optimal recommendation level increases with earlier replenishment: $\alpha^{*}_1\big|_{(q_1,q_2)=(1,0)}\geq\alpha^{*}_1\big|_{(q_1,q_2)=(0,1)}$ and $\alpha^{*}_1\big|_{(q_1,q_2)=(1,1)}\geq\alpha^{*}_1\big|_{(q_1,q_2)=(0,2)}$.
To examine how the optimal recommendation ratio $\alpha^*_1/\alpha^*_2$ adjusts, we examine the cross–partial derivative of $\mathcal{L}$. 
When $\sum_t q_t=2$, we have 
\begin{align*}
\frac{\partial^2 \mathcal{L}}{\partial \alpha_1 \partial \alpha_2}\bigg|_{q_1+q_2=2}
= (p+h)\,(\overline{R}-\eta R_0)\,
     \gamma_2 (1 - \gamma_2)\,
     \left[
        \eta (1 - \alpha_2)(1 - 2\gamma_2)(1 - \eta \alpha_1)
        - \eta
     \right].
\end{align*}
Since $\gamma_t\geq\frac{1}{2}$ by definition, we get $\frac{\partial^2 \mathcal{L}}{\partial \alpha_1 \partial \alpha_2}\big|_{q_1+q_2=2}\leq0$. Thus, $\alpha_1$ and $\alpha_2$ are substitutes in this regime. 
When $\sum_t q_t=1$, the cross–partial expands to
\begin{align*}
\frac{\partial^2 \mathcal{L}}{\partial \alpha_1 \partial \alpha_2}\bigg|_{q_1+q_2=1}
&=
\left[(p+h) - \gamma_1 (p+h+b)\right]
\, \gamma_2(1 - \gamma_2)
\, (\overline{R}-\eta R_0)
\left[
    \eta(1 - \alpha_2)(1 - 2\gamma_2)(1 - \eta \alpha_1)
    - \eta
\right] \\
&\quad
- (p+h+b)\,
\gamma_2(1 - \gamma_2)\,
(\overline{R}-\eta R_0)\,
\gamma_1(1 - \gamma_1)\,
(\bar{R} - \eta R_0).
\end{align*}
Note that $p\geq b e^{\overline{R}}-h$, we have $(p+h) - \gamma_1 (p+h+b)\geq0$ for all $\alpha_1\in[0,1]$, ensuring that the entire expression remains non-positive. Hence, marginal substitutability between $\alpha_1$ and $\alpha_2$ continues to hold in this case. Therefore, we can conclude that the optimal recommendation intensity ratio $\alpha_1^*/\alpha_2^*$ increases with $q_1$, i.e., $\alpha_1^*/\alpha_2^*\big|_{(q_1,q_2)=(1,0)}\geq \alpha_1^*/\alpha_2^*\big|_{(q_1,q_2)=(0,1)}$ and $\alpha_1^*/\alpha_2^*\big|_{(q_1,q_2)=(1,1)}\geq \alpha_1^*/\alpha_2^*\big|_{(q_1,q_2)=(0,2)}$.
\Halmos
\endproof

\proof{Proof of Proposition \ref{prop:adaptive_ordering}.}

Building on the previous argument in the proof of Proposition \ref{prop:demand_smoothing}, we first analyze the optimal replenishment decision using the intermediate variables $\gamma_t$, which help simplify the characterization. Note that $\mathcal{L}\big\vert_{q_1 +q_2= 2} -\mathcal{L}\big\vert_{q_1 +q_2= 1} = \gamma_1\gamma_2(p+h+b) -h$ and $\mathcal{L}_{(q_1,q_2)=(0,0)}  =\mathcal{L}_{(q_1,q_2)=(0,1)} + (p+h+b)(1-\gamma_2)(1-\gamma_1) - (p+b) =\mathcal{L}_{(q_1,q_2)=(1,0)} + (p+h+b)(1-\gamma_2)(1-\gamma_1) - (p+b) +  \gamma_1(h+b) -h$. These gaps between different replenishment decisions result in the following threshold-type conditions for optimal replenishment quantities. If the recommendation intensities $\alpha_t$ in both periods are sufficiently strong, leading to a high joint purchasing probability such that $\gamma_1\gamma_2\geq\frac{h}{p+h+b}$, then the optimal total replenishment quantity is two units: $q^*_1+q^*_2=2$. Otherwise, if the recommendations are weaker such that $\gamma_1\gamma_2<\frac{h}{p+h+b}$ and $(1-\gamma_1)(1-\gamma_2)\leq\frac{p+b}{p+h+b}$, then the platform optimally replenishes only one unit: $q^*_1+q^*_2=1$. If $(1-\gamma_1)(1-\gamma_2)>\frac{p+b}{p+h+b}$, it is no longer profitable to replenish any inventory.
Under the same total replenishment amount, we can further compute that
$\mathcal{L}\big\vert_{q_1=1,\ q_1+q_2= 1} -\mathcal{L}\big\vert_{q_1=0,\ q_1 +q_2= 1} = \mathcal{L}\big\vert_{q_1=1,\ q_1+q_2= 2} -\mathcal{L}\big\vert_{q_1=0,\ q_1 +q_2= 2} = \gamma_1(h+b) -h$, which implies the threshold condition under which early replenishment is preferred. Specifically, if the platform replenishes at least one unit, i.e., $(1-\gamma_1)(1-\gamma_2) \leq \frac{p+b}{p+h+b}$, and the recommendation intensity $\alpha_1$ at the first period is sufficiently strong such that $\gamma_1 > \frac{h}{h+b}$, then it is optimal to place the order in the first period, resulting in $q^*_1 = 1$.

Assuming the total replenishment quantity satisfies $\sum_t q_t > 0$, the key threshold condition  
$
\gamma_1 \gamma_2 = \frac{h}{p + h + b}
$
can be equivalently rewritten in terms of $R_1$ and $R_2$ as
$
(1 + e^{-R_1})(1 + e^{-R_2}) = \frac{p + h + b}{h}.
$
This equation defines a level curve in the $(R_1, R_2)$ plane that is tangent to the line $R_1 + R_2 = 2R^+$ at the symmetric point $(R_1, R_2) = (R^+, R^+)$, where
$
R^+ := -\log\left(\sqrt{\frac{p + h + b}{h}} - 1\right).
$
Notably, if $p \ge h + \frac{b^2}{h}$, this tangency point lies strictly to the left of the vertical line defined by the threshold condition $\gamma_1(R_1) = \frac{h}{h + b}$, which governs the initiation of first-period replenishment.
Likewise, a sufficient condition for strictly positive replenishment ($\sum_t q_t > 0$) can be derived by noting that whenever $R_1 + R_2 \ge 2R^-$, with
$
R^- := \log\left(\sqrt{\frac{p + h + b}{h + b}} - 1\right),
$, it follows that $(1 - \gamma_1)(1 - \gamma_2) \le \frac{h + b}{p + h + b}$, ensuring that at least one unit of inventory is ordered.

Combining these observations, consider any fixed total signal line $R_1 + R_2 = C$ with $C \ge 2R^+$. As we traverse this line from left to right---increasing $R_1$ while decreasing $R_2$---the optimal replenishment ratio $q_1^*/q_2^*$ evolves monotonically: it increases from 0 to 1 and ultimately to $+\infty$, or in some cases directly jumps from 0 to $+\infty$. In either case, the ratio is non-decreasing in $R_1$, implying a threshold-type structure in the optimal ordering policy. This completes the proof.
\Halmos
\endproof

\section{Supplementary Materials for Section \ref{sec:multi_scale_RL}}\label{appendix4}

\subsection{Convergence Analysis in Subsection \ref{subsec4.1:multi-timescale_SA}}\label{appendix4.1}

For the general single-period model with zero lead time, the expected overall profit is given by 
\begin{align*}
\mathcal{L}(q,\alpha) = \sum_{i=1}^N \mathbb{E}[P^i] - \sum_{i,j} C^{i,j} = \sum_{i=1}^N \mathbb{E}[\mathcal{P}(S^i, q^i, I^i, U^i)] - \sum_{i,j} \mathcal{C}(\alpha^{i,j}).
\end{align*}
Recall here that $S^i=D^i-[D^i-q^i]$, $U^i = [D^{i}-q^{i} ]^{+}$ and $I^i = [q^{i}-D^{i} ]^{+}$. For almost all realizations with $D^i\neq q^i$, the derivatives satisfy: $\frac{\partial S^i}{\partial q_i} = -\frac{\partial U^i}{\partial q_i} = \mathbf{1}\{D^i > q^i\}$ and $\frac{\partial I^i}{\partial q_i} =\mathbf{1}\{D^i < q^i\}$. Moreover, since $\frac{\partial}{\partial q^{i}}\mathbb{E}[P^{i'}] = 0$ for $i\neq i'$, only the $i$-th term contributes to the gradient with respect to $q^i$. Under standard integrability conditions justifying the interchange of differentiation and expectation, the gradient of the expected profit function with respect to the replenishment quantity $q^i$ is
\begin{align}
    \frac{\partial}{\partial q^i}\mathcal{L}(q,\alpha) = \mathbb{E}\bigg[\frac{\partial P^i}{\partial q^i} + \big(\frac{\partial P^i}{\partial S^i} - \frac{\partial P^i}{\partial U^i}\big) \mathbf{1}\{D^i> q^i\} +  \frac{\partial P^i}{\partial I^i} \mathbf{1}\{D^i< q^i\} \bigg].
\end{align}
To estimate the gradient with respect to recommendation intensity $\alpha^{i,j}$, we apply the likelihood ratio method. Let $f(d^j;R^{1,j},\cdots,R^{N,j})$ denote denote the joint density of customer $j$'s choices, parameterized by the recommendation scores $R^{i,j}$. Then,
\begin{align*}
    \frac{\partial}{\partial \alpha^{i,j}}\mathbb{E}\bigg[\sum_{i=1}^N  P^i\bigg] 
    &=  \int_{\mathbb{R}^{N}} \mathbb{E}\big[\sum_{i=1}^N  P^i\big| d^j\big]\ \frac{\partial}{\partial \alpha^{i,j}}  f(d^j;R^{1,j},\cdots,R^{N,j})\ d (d^j) \\
    &= \mathbb{E}\bigg[\big(\sum_{i=1}^N P^i\big) \frac{\partial }{\partial \alpha^{i,j}}  \ln f(d^j;R^{1,j},\cdots,R^{N,j})  \bigg], 
\end{align*}
where the first equality follows from the tower property of conditional expectation.
Thus, the gradient of the objective function with respect to the recommendation intensity $\alpha^{i,j}$ is given by
\begin{align}
    \frac{\partial}{\partial \alpha^{i,j}}\mathcal{L}(q,\alpha) = \mathbb{E}\bigg[\big(\sum_{i=1}^N P^i\big) \frac{\partial }{\partial R^{i,j}}  \ln f(d^j;R^{1,j},\cdots,R^{N,j}) \frac{\partial R^{i,j}}{\partial \alpha^{i,j}} - \frac{\partial C^{i,j}}{\partial \alpha^{i,j}} \bigg].
\end{align}
Accordingly, at each iteration $n$, the stochastic gradient estimates $Q[n]$ and $A[n]$, used for updating the decision variables in \eqref{iter:1}-\eqref{iter:2}, are obtained via direct Monte Carlo sampling of the above expectation expressions, where the randomness stems from the sampling of customer demand.

In the subsequent analysis, we characterize the asymptotic behavior of the proposed update rules in (\ref{iter:1})–(\ref{iter:2}) by by establishing their correspondence with the following system of coupled ODEs:
\begin{align}
    \dot{q}(\xi) &= \tilde{\Pi}_{[0,\overline{q}]}(\nabla_q\mathcal{L}(q(\xi),\alpha(\xi))) = \nabla_q\mathcal{L}(q(\xi),\alpha(\xi))) +p_q(\xi), \label{eq:ode1}\\
    \dot{\alpha}(\xi) &= \tilde{\Pi}_{[0,1]}(\nabla_{\alpha}\mathcal{L}(q(\xi),\alpha(\xi))), 
    \label{eq:ode2}
\end{align}
where $\tilde{\Pi}_{[\cdot,\cdot]}$ denotes the projected operator onto the corresponding feasible domain, and $p_q(\xi)$ is the realization of the projection. Given that the step sizes satisfy $\varepsilon_2[n]=o(\varepsilon_1[n])$, the recursion (\ref{iter:2}) evolves on a slower timescale than (\ref{iter:1}), allowing the slower dynamics to be treated as quasi-static when analyzing the faster updates. Consequently, the fast process (\ref{iter:1}) closely tracks  its instantaneous best response $q^*(\alpha[n])$ with respect to the current value of $\alpha[n]$. Exploiting this timescale separation, the coupled ODE system reduces to a single limiting ODE governing the evolution of the slow-timescale variable $\alpha$: 
\begin{align}
    \dot{\alpha}(\xi) = \tilde{\Pi}_{[0,1]}(\nabla_{\alpha}\mathcal{L}(q^*(\alpha(\xi)),\alpha(\xi))), \label{eq:ode3}
\end{align}
which serves as the foundation for the overall convergence analysis.
Denote the increasing family of $\sigma$-fields as $\mathcal{F}[n]:=(q[0],\alpha[0],\cdots,q[n],\alpha[n])$. We now introduce the following technical assumptions to support the convergence results.

\begin{assumption}\label{assumption:Lip}
    The gradients of the objective function with respect to decision variables are Lipschitz continuous. That is, there exists a constant $L>0$ such that for all $(q_i,\alpha_i)\in [0,\bar{q}]\times[0,1]$, $i=1,2$, and each $z\in\{q,\alpha\}$, $\Vert\nabla_{z}\mathcal{L}(q_1,\alpha_1) -\nabla_{z}\mathcal{L}(q_2,\alpha_2) \Vert \leq L (\Vert q_1 - q_2 \Vert + \Vert \alpha_1 - \alpha_2 \Vert)$.
\end{assumption}

\begin{assumption}\label{assumption:var}
    The stochastic gradient noise sequences form square-integrable martingale differences. Let $M_q[n]:=Q[n]-\nabla_{q}\mathcal{L}(q[n],\alpha[n])$ and $M_{\alpha}[n]:=A[n]-\nabla_{\alpha}\mathcal{L}(q[n],\alpha[n])$. Then, for each $z\in\{q,\alpha\}$, there exists a constant $K>0$ such that for all $n\geq 0$,
    $$\mathbb{E}[M_z[n+1]|\mathcal{F}[n]] = 0 \text{ and } \mathbb{E}[\Vert M_z[n+1]\Vert^2|\mathcal{F}[n]] \leq K, \text{ a.s.}$$
\end{assumption}

\begin{assumption}\label{assumption:concave}
    The objective function $\mathcal{L}(q,\alpha)$ is strictly concave in the replenishment quantities $q$ for any fixed $\alpha$.
\end{assumption}

Assumption \ref{assumption:Lip} ensures the Lipschitz continuity of the objective gradient over the joint parameter space, facilitating stable convergence analysis. Assumption \ref{assumption:var} imposes a bounded-variance martingale difference structure on the stochastic gradient noise, enabling effective asymptotic control of estimation errors. Assumption \ref{assumption:var} guarantees that, for any fixed value of the slow-timescale variable $\alpha$, the fast-timescale subproblem is strictly concave in $q$, thereby admitting a unique best response. This property is crucial for reducing the coupled system to a single limiting ODE that governs the evolution of $\alpha$. These conditions are mild and broadly applicable in practice. For instance, in standard linear-inventory-profit models with Lipschitz-continuous recommendation cost functions and independent customer demand distributions drawn from controlled continuous families, all of the above assumptions are readily satisfied. Assumption~\ref{assumption:concave} holds in many standard settings, including the linear profit structure discussed in Section~\ref{Cross-Product Synergy: Interdependent Inventory and Marketing Strategies}.

In cases where Assumption~\ref{assumption:Lip} does not hold, our analysis can be extended using the framework of differential inclusions as described in \citet{filippov2013differential_appendix}, which generalizes the notion of a limiting ODE to accommodate discontinuous dynamics. Similarly, if Assumption~\ref{assumption:var} fails, the convergence guarantees can still be recovered under the correlated noise framework outlined in Chapter 6 of \citet{kushner2003stochastic}. Finally, when the strict concavity condition in Assumption~\ref{assumption:concave} is not satisfied, we address the resulting complexities in our subsequent analysis of the multi-agent RL setting.

\begin{lemma}[Fast-timescale convergence to conditional optimum]\label{mtsconv1} Suppose $T=1$, and Assumptions \ref{assumption:stepsize}, \ref{assumption:Lip}-\ref{assumption:concave} hold. Then, the sequence $\{(q[n],\alpha[n])\}_n$ generated by update rules (\ref{iter:1})-(\ref{iter:2}) converges almost surely to the set $\{(q^{*}(\alpha),\alpha) : \alpha\in[0,1]^{N\times M}\}$, where $q^{*}(\cdot)$ denotes the optimal replenishment quantities given a recommendation strategy.
\end{lemma}

\proof{Proof of Lemma \ref{mtsconv1}.}
Rewriting the recursions (\ref{iter:1})-(\ref{iter:2}) as 
\begin{align}
q[n+1] =&\ q[n] + \varepsilon_{1}[n] \nabla_q\mathcal{L}(q[n],\alpha[n]) + \varepsilon_{1}[n]\ M_q[n] + \varepsilon_1[n]Q_p[n], \label{iter:11}\\
\alpha[n+1] =&\   \alpha[n] + \varepsilon_{1}[n] \frac{\varepsilon_{2}[n]}{\varepsilon_{1}[n]}\ \nabla_\alpha\mathcal{L}(q[n],\alpha[n]) + \varepsilon_{1}[n] \frac{\varepsilon_{2}[n]}{\varepsilon_{1}[n]}\ M_\alpha[n]+  \varepsilon_{1}[n] \frac{\varepsilon_{2}[n]}{\varepsilon_{1}[n]}\ A_p[n],  \label{iter:21}
\end{align}
where $\varepsilon_1[n]Q_p[n]:=q[n+1] - (q[n] + \varepsilon_{1}[n]\ Q[n])$ and $\varepsilon_2[n]A_p[n]:=A[n+1] - (A[n] + \varepsilon_{2}[n]\ A[n])$ are the minimal corrections by projection to keep parameters within the feasible region. 

\textbf{We first link the recursions to equicontinuous interpolation processes.}
Let $\xi[0]=0$, $\xi[n]=\sum_{n'=0}^{n-1}\varepsilon_1[n']$, and define $\langle\xi\rangle:=\max\{n:\xi[n]\leq\xi\}$. Construct the piecewise constant interpolation $q^{[0]}(\cdot)$ on $\mathbb{R}$ via $q^{[0]}(\xi)=q[0]$ for $\xi<0$, and $q^{[0]}(\xi)=q[n]$ for $\xi\in\big[\xi[n], \xi[n+1]\big)$.  Define the shifted processes $q^{[n]}(\xi)=q^{[0]}(\xi[n]+\xi)$. Similarly, we construct the shifted processes of (\ref{iter:21}) but using the time index of (\ref{iter:11}), i.e., $\alpha^{[0]}(\xi) = \alpha[0]$ for $\xi\in\big[\xi[n], \xi[n+1]\big)$, and $\alpha^{[n]}(\xi)=\alpha^{[0]}(\xi[n]+\xi)$.
By definition, we have from (\ref{iter:11}) that 
\begin{align}
q^{[n]}(\xi) = q[n] + \int_0^\xi \nabla_q\mathcal{L}(q^{[n]}(\xi'), \alpha^{[n]}(\xi')) d\xi' + M_q^{[n]}(\xi) + P_q^{[n]}(\xi) +\rho^{[n]}(\xi), \label{eq:continue_iter}
\end{align}
where we introduce $M_q^{[n]}(\xi) =\sum_{n'=n}^{\langle\xi[n]+\xi\rangle-1} \varepsilon_1[n'] M_q[n']$, $P_q^{[n]}(\xi) =\sum_{n'=n}^{\langle\xi[n]+\xi\rangle-1} \varepsilon_1[n'] Q_p[n']$, and $\rho^{[n]}(\xi)=\sum_{n'=n}^{\langle\xi[n]+\xi\rangle-1} \varepsilon_1[n'] \nabla_q\mathcal{L}(q[n'],\alpha[n']) - \int_0^\xi \nabla_q\mathcal{L}(q^{[n]}(\xi'), \alpha^{[n]}(\xi')) d\xi'$.
By Assumptions \ref{assumption:stepsize}(a) and \ref{assumption:var}, we have
$\sup_n \mathbb{E}\big[\Vert M_q^{[0]}(\xi[n])\Vert^2\big]
=\sup_n\mathbb{E}\big[\sum_{n'=0}^{n-1} \varepsilon_1[n']^2 \mathbb{E}[\Vert M_q[n']\Vert^2|\mathcal{F}[n']]\big] < \infty$. 
Thus, the sequence $\{M_q^{[0]}(\xi[n])\}_n$ forms a square-integrable martingale. By the Martingale Convergence Theorem \citep{durrett2019probability_appendix}, it converges a.s. Then, on any fixed bounded interval, $M_q^{[n]}(\xi) = M_q^{[0]}(\langle \xi[n]+t\rangle) - M_q^{[0]}(\xi[n])$ converges uniformly to zero as $n\rightarrow\infty$, a.s.
Since $\rho^{[n]}(\xi)$ is incurred when replacing the summation term by an integral, and $\nabla_q\mathcal{L}(q, \alpha)$ is Lipschitz continuous on bounded regions by Assumption \ref{assumption:Lip}, it follows that $\rho^{[n]}(\xi)=O(\varepsilon_1[n])$ also converges uniformly to zero as $n\rightarrow\infty$, a.s. Therefore, both $\{M_q^{[n]}(\cdot)\}$ and $\{\rho^{[n]}(\cdot)\}$ are equicontinuous in the extended sense on any bounded interval. Because the remaining terms in (\ref{eq:continue_iter}) other than $ P_q^{[n]}(\xi) $ vanish on bounded intervals, the sequence $ \{P_q^{[n]}(\cdot)\} $ inherits this equicontinuity.
 Along with the boundedness of $\{q[n]\}$ and $\nabla_q\mathcal{L}(q[n],\alpha[n])$, we can conclude that $\{q^{[n]}(\cdot)\}$ is equicontinuous in the extended sense.

Similarly, we have from (\ref{iter:21}) that $\alpha^{[n]}(\xi)=\alpha[n]+B^{[n]}(\xi) + M_\alpha^{[n]}(\xi) + P_\alpha^{[n]}(\xi)$, 
where $B^{[n]}(\xi)=\sum_{n'=n}^{\langle\xi[n]+\xi\rangle-1} \varepsilon_1[n'] \frac{\varepsilon_2[n']}{\varepsilon_1[n']}\nabla_{\alpha}\mathcal{L}(q[n'],\alpha[n'])$, $M_\alpha^{[n]}(\xi)=\sum_{n'=n}^{\langle\xi[n]+\xi\rangle-1} \varepsilon_1[n'] \frac{\varepsilon_2[n']}{\varepsilon_1[n']}M_{\alpha}[n]$, and $P_\alpha^{[n]}(\xi)=\sum_{n'=n}^{\langle\xi[n]+\xi\rangle-1} \varepsilon_1[n'] \frac{\varepsilon_2[n']}{\varepsilon_1[n']}A_p[n']$.
By Assumptions~\ref{assumption:stepsize}(b) and~\ref{assumption:Lip},  
$\frac{\varepsilon_2[n']}{\varepsilon_1[n']}\nabla_{\alpha}\mathcal{L}(q[n'],\alpha[n'])\rightarrow 0$ as $n\rightarrow \infty$,
and $\{B^{[n]}(\cdot)\}$ converges uniformly to zero a.s. and is equicontinuous on any bounded interval. Intuitively, the slow-timescale updates appear as a vanishing perturbation when analyzing the fast-timescale recursion.
Applying similar arguments to the martingale and projection terms yields analogous conclusions for $\{M_{\alpha}^{[n]}(\cdot)\}$ and $\{P_{\alpha}^{[n]}(\cdot)\}$, as in~\eqref{eq:continue_iter}. Thus, $\{\alpha^{[n]}(\cdot)\}$ is also equicontinuous in the extended sense.
It follows that the sequence $\{\alpha^{[n]}(\cdot)\}$ is equicontinuous in the extended sense.  
Moreover, because $B^{[n]}(\xi)$ and $M_{\alpha}^{[n]}(\xi)$ converge uniformly to zero, the same claim is true for $P_{\alpha}^{[n]}(\xi)$.

\textbf{Second, we characterize the limiting behavior of interpolation processes via ODEs.}
According to the Arzelà–Ascoli Theorem, the equicontinuity of the sequence $\{q^{[n]}(\cdot), \alpha^{[n]}(\cdot)\}$ implies that, along almost every
trajectory generated by the algorithm, there exists a convergent subsequence whose limit $\{q(\cdot), \alpha(\cdot)\}$ satisfies:
\begin{align*}
    q(\xi) = q(0)+\int_0^\xi \nabla_q\mathcal{L}(q(\xi'),\alpha(\xi'))d\xi' + P_q(\xi), \text{ and }\alpha(\xi)=\alpha(0),
\end{align*}
where $P_q(\xi)$ is the limiting projection term. 
This limit corresponds to the coupled ODE system given by (\ref{eq:ode1}) and $\dot\alpha(\xi)=0$, which further simplifies to a single ODE of the form $\dot{q}(\xi)  = \nabla_q\mathcal{L}(q(\xi),\bar{\alpha}) +p_q(\xi)$ for  some fixed $\bar{\alpha}\in[0,1]$.
By Assumption \ref{assumption:concave}, $q^*(\bar{\alpha})$ is the unique equilibrium point. Take $V(x)=\Vert x-q^*(\bar{\alpha})\Vert^2$ as the Lyapunov function, and the derivative is $V'(x) = 2(x-q^*(\bar{\alpha})) (\nabla_q\mathcal{L}(x,\bar{\alpha}) +p_q(\xi))$. Since $\mathcal{L}(q,\alpha)$ is strictly concave with respect to $q$ by Assumption \ref{assumption:concave}, $(q^*(\bar{\alpha})-x)\nabla_q\mathcal{L}(x,\bar{\alpha})> \mathcal{L}(q^*(\bar{\alpha}),\bar{\alpha}) - \mathcal{L}(x,\bar{\alpha}) > 0$ for any $x\neq q^*(\bar{\alpha})$. Note that $p_q(\xi)=-\nabla_q\mathcal{L}(q(\xi),\bar{\alpha})\cdot\mathbf{1}\{q(\xi)\in\{0,\overline{q}\}\}$
and $(q^*(\bar{\alpha})-x)p_q(\xi)\leq 0$ by definition. Thus, $q^*(\bar{\alpha})$ is global asymptotically stable by Lyapunov Stability Theorem \citep{liapounoff2016probleme_appendix}. Therefore, the single ODE converges to $q^*(\bar{\alpha})$.
Because the interpolation process $\{q^{[n]}(\cdot), \alpha^{[n]}(\cdot)\}$ has the same asymptotic behavior as $\{q[n],\alpha[n]\}$, it converges to $(q^{*}(\bar\alpha),\bar\alpha)$ with probability one.
\Halmos
\endproof

\proof{Proof of Theorem \ref{mtsconv2}.}
We first transform the recursion (\ref{iter:1}) as 
\begin{align}
\alpha[n+1] =&\   \alpha[n] + \varepsilon_{2}[n]\ \nabla_\alpha\mathcal{L}(q^*(\alpha[n]),\alpha[n]) + \varepsilon_{2}[n]\ \tilde{B}[n] + \varepsilon_{2}[n]\ M_\alpha[n]+  \varepsilon_{2}[n]  \ A_p[n],  \label{iter:22}
\end{align}
where $\tilde{B}[n] = \nabla_\alpha\mathcal{L}(q[n],\alpha[n]) - \nabla_\alpha\mathcal{L}(q^*(\alpha[n]),\alpha[n])$.
Let $\zeta[0]=0$, $\zeta[n]=\sum_{n'=0}^{n-1}\varepsilon_2[n']$, and define $\langle\zeta\rangle':=\max\{n:\zeta[n]\leq\zeta\}$. Define the piecewise constant interpolation $\tilde{\alpha}^{[0]}(\cdot)$ on $\mathbb{R}$ by $\tilde{\alpha}^{[0]}(\zeta)=\tilde{\alpha}[0]$ for $\zeta<0$, and $\tilde{\alpha}^{[0]}(\zeta)=\tilde{\alpha}[n]$ for $\zeta\in\big[\zeta[n], \zeta[n+1]\big)$.  Denote the shifted process as $\tilde{\alpha}^{[n]}(\zeta)=\tilde{\alpha}^{[0]}(\zeta[n]+\zeta)$. 
By definition, we have from (\ref{iter:11}) that 
\begin{align}
\tilde{\alpha}^{[n]}(\zeta) =  \alpha[n] + \int_0^\zeta \nabla_\alpha\mathcal{L}(q^*(\tilde{\alpha}^{[n]}(\zeta')), \tilde{\alpha}^{[n]}(\zeta')) d\zeta'  + \tilde B ^{[n]}(\zeta) + \tilde M_\alpha^{[n]}(\zeta) + \tilde P_\alpha^{[n]}(\zeta) + \tilde\rho^{[n]}(\zeta), 
\end{align}
where we again denote $\tilde{B}^{[n]}(\zeta) =\sum_{n'=n}^{\langle\zeta[n]+\zeta\rangle'-1} \varepsilon_2[n']\tilde{B}[n']$, $\tilde M_\alpha^{[n]}(\zeta) =\sum_{n'=n}^{\langle\zeta[n]+\zeta\rangle'-1} \varepsilon_2[n'] M_\alpha[n']$, $\tilde{P}_\alpha^{[n]}(\zeta) =\sum_{n'=n}^{\langle\zeta[n]+\zeta\rangle'-1} \varepsilon_2[n'] A_p[n']$, and $\tilde\rho^{[n]}(\zeta)=\sum_{n'=n}^{\langle\zeta[n]+\zeta\rangle'-1} \varepsilon_2[n'] \nabla_\alpha\mathcal{L}(q^*(\alpha[n']),\alpha[n']) - \int_0^\zeta \nabla_\alpha\mathcal{L}(q^*(\alpha^{[n]}(\zeta')), \alpha^{[n]}(\zeta')) d\zeta'$.
Following the same arguments as in the proof of Lemma \ref{mtsconv1}, the sequences $\{\tilde M_\alpha^{[n]}(\cdot)\}$, $\{\tilde P_\alpha^{[n]}(\cdot)\}$, and $\{\tilde \rho ^{[n]}(\cdot)\}$ are equicontinuous over bounded intervals; $\tilde M_\alpha^{[n]}(\zeta)$ and $\tilde\rho^{[n]}(\zeta)$ converge to zero uniformly on each bounded interval as $n\rightarrow\infty$, a.s.
Notice that
$\Vert\tilde{B}[n]\Vert \leq \Vert\nabla_\alpha\mathcal{L}(q[n],\alpha[n]) - \nabla_\alpha\mathcal{L}(q^*(\bar\alpha),\bar\alpha)\Vert +\Vert \nabla_\alpha\mathcal{L}(q^*(\bar\alpha),\bar\alpha)-\nabla_\alpha\mathcal{L}(q^*(\alpha[n]),\alpha[n])\Vert$. 
With Assumption \ref{assumption:Lip} and Lemma \ref{mtsconv1}, the first term on the right-hand side vanishes as $n\rightarrow\infty$. Moreover, under Assumptions \ref{assumption:Lip} and \ref{assumption:concave}, the mapping $q^*(\cdot)$ is Lipschitz continuous; hence, the second term also vanishes  as $\{\alpha[n]\}$ converges to $\bar\alpha$ a.s.
We can conclude that $\{\tilde B^{[n]}(\cdot)\}$ is also equicontinuous and converges to zero uniformly on any bounded interval as $n \to \infty$, a.s. Therefore, $\{\tilde \alpha^{[n]}(\cdot)\}$ is equicontinuous. By the Arzelà–Ascoli theorem, the limit of its convergent subsequence satisfies the ODE~\eqref{eq:ode3}, which implies that the sequence $\{(q[n], \alpha[n])\}_n$ converges with probability one to a stationary point of the coordination problem~\eqref{eq:obj}, subject to $q = q^*(\alpha)$. 

If $\mathcal{L}(q, \alpha)$ is strictly concave in $\alpha$, then analogously to the second part of the proof of Lemma~\ref{mtsconv1}, the ODE~\eqref{eq:ode3} admits $(q^*,\alpha^*)$ as its unique globally asymptotically stable equilibrium, and $\{(q[n], \alpha[n])\}_n$ further converges to $(q^*,\alpha^*)$ a.s.
\Halmos
\endproof

\subsection{Algorithmic Details for Section \ref{subsec4.3:multi-timescale_multi_agent_RL}}\label{appendix4.3algo}

In this section, we provide the details regarding the multi-timescale multi-agent RL algorithm, along with supporting definitions. 
We begin by highlighting a key structural property: when an agent's policy remains unchanged, its private advantage becomes zero, i.e.,
\begin{equation*}
\resizebox{\linewidth}{!}{$%
\begin{aligned}
&\mathbb{E}_{\substack{a\sim\pi(\theta[n])\\s\sim\rho(\theta[n])}}\bigg[\frac{\pi(a|s;\theta)}{\pi(a|s;\theta[n])} A_{z',z}(s,a;\theta[n])\bigg]\bigg|_{\theta_{z'}=\theta_{z'}[n]}=\mathbb{E}_{\substack{z\sim\pi_z(\theta_z[n])\\s\sim\rho(\theta[n])}}\bigg[\frac{\pi_z(a_z|s;\theta_z)}{\pi_z(a_z|s;\theta_z[n])} \mathbb{E}_{z'\sim\pi_{z'}(\theta_{z'}[n])}\bigg[  A_{z',z}(s,a;\theta[n])\bigg]\bigg]  = 0.\\
&\mathbb{E}_{\substack{a\sim\pi(\theta[n])\\s\sim\rho(\theta[n])}}\bigg[\frac{\pi(a|s;\theta)}{\pi(a|s;\theta[n])} A_{z}(s,a_z;\theta[n])\bigg]\bigg|_{\theta_{z}=\theta_{z}[n]} =\mathbb{E}_{\substack{z'\sim\pi_{z'}(\theta_{z'}[n])\\s\sim\rho(\theta[n])}}\bigg[\frac{\pi_{z'}(a_{z'}|s;\theta_{z'})}{\pi_z(a_{z'}|s;\theta_{z'}[n])} \mathbb{E}_{z\sim\pi_{z}(\theta_{z}[n])}\bigg[  A_{z}(s,a_z;\theta[n])\bigg]\bigg]  = 0.
\end{aligned}
$}%
\end{equation*}
Note that the agent-specific advantage is zero when the corresponding agent is not updated.  
Therefore, the sequential optimization of the multi-agent surrogate objective~(\ref{eq:matrpo})  
is equivalent to the recursive algorithm~(\ref{eq:matrpo2}).  
We next derive how local advantages can be replaced with reweighted global advantages.  
By the definition of the advantage function, the global advantage can be decomposed into individual agent advantages as follows \citep{kuba2021trust}:
\begin{equation}\label{eq:advdecop}
\begin{aligned}
A(s,a;\theta) &= Q(s,a;\theta) - Q_{z'}(s,a_{z'};\theta) + Q_{z'}(s,a_{z'};\theta) - V(s;\theta) \\ &=  A_{z}(s,a_z;\theta) +A_{z',z}(s,a;\theta).
\end{aligned}
\end{equation}
This decomposition explicitly represents how the global advantage can be attributed individually to each agent, facilitating clearer analysis and optimization.
Therefore, we have 
\begin{align*}
    &\mathbb{E}_{\substack{{s\sim\rho( \theta[n])}\\{a\sim\pi(\theta[n])} }}\bigg[ \frac{\pi_{z'}(a_{z'}|s;\theta_{z'})}{\pi_{z'}(a_{z'}|s;\theta_{z'}[n])}  \frac{\pi_z(a_z|s;\theta_z )}{\pi_z(a_z|s;\theta_z[n])}A_{z',z}(s,a;\theta[n])\bigg]\\
    =\ & \mathbb{E}_{\substack{{s\sim\rho( \theta[n])}\\{a\sim\pi(\theta[n])} }}\bigg[ \frac{\pi_{z'}(a_{z'}|s;\theta_{z'})}{\pi_{z'}(a_{z'}|s;\theta_{z'}[n])}  \frac{\pi_z(a_z|s;\theta_z )}{\pi_z(a_z|s;\theta_z[n])} A(s,a;\theta[n])-   \frac{\pi_z(a_z|s;\theta_z )}{\pi_z(a_z|s;\theta_z[n])} 
    A_{z}(s,a_z;\theta[n]) \bigg]\\
    =\ & \mathbb{E}_{\substack{{s\sim\rho( \theta[n])}\\{a\sim\pi(\theta[n])} }}\bigg[ \bigg(\frac{\pi_{z'}(a_{z'}|s;\theta_{z'})}{\pi_{z'}(a_{z'}|s;\theta_{z'}[n])} -1 \bigg)  \frac{\pi_z(a_z|s;\theta_z )}{\pi_z(a_z|s;\theta_z[n])} A(s,a;\theta[n])\bigg],
\end{align*}
where the first equality follows from Equation~(\ref{eq:advdecop}) by omitting irrelevant variables, and the second one follows from the definition of $A_{z}(s,a_z;\theta[n])$. Since the term inside the bracket does not contribute to the gradient with respect to $ \theta_{z'} $, it can be safely omitted from the optimization. Similarly, we can apply the same derivation to the term $\mathbb{E}_{\substack{{s\sim\rho( \theta[n])}\\{a\sim\pi(\theta[n])} }}\big[ \frac{\pi_z(a_z|s;\theta_z) A_z(s,a_z;\theta[n])}{\pi_z(a_z|s;\theta_z[n])}\big] $, leading to the following update rule:
\begin{equation*}
\resizebox{\linewidth}{!}{$%
 \begin{aligned}
     \theta_{z}[n+1]&=\arg\max_{\theta_z\in\Theta_z}\mathbb{E}_{\substack{{s\sim\rho( \theta[n])}\\{a\sim\pi(\theta[n])} }}\bigg[ \frac{\pi_z(a_z|s;\theta_z) A(s,a;\theta[n])}{\pi_z(a_z|s;\theta_z[n])}\bigg]- B(\theta[n]) \max_s\text{KL}(\pi_z(\cdot|s;\theta_z[n])\Vert\pi_z(\cdot|s;\theta_z)),\\
     \theta_{z'}[n+1]&=\arg\max_{\theta_{z'}\in\Theta_{z'}}\mathbb{E}_{\substack{{s\sim\rho( \theta[n])}\\{a\sim\pi(\theta[n])} }}\bigg[ \frac{\pi_{z'}(a_{z'}|s;\theta_{z'})}{\pi_{z'}(a_{z'}|s;\theta_{z'}[n])}  \frac{\pi_z(a_z|s;\theta_z[n+1])}{\pi_z(a_z|s;\theta_z[n])}A(s,a;\theta[n])\bigg]\\
     &\quad\quad\quad\quad\quad\quad\quad\quad\quad\quad\quad\quad\quad\quad\quad\quad\quad\quad\quad\quad- B(\theta[n]) \max_s\text{KL}(\pi_{z'}(\cdot|s;\theta_{z'}[n])\Vert\pi_{z'}(\cdot|s;\theta_{z'})).
\end{aligned}  
$}%
\end{equation*}
Since exactly evaluating the maximum KL divergence is computationally prohibitive, we adopt a PPO-style clipping operator as a practical approximation, robustly regulating policy updates. This clipping mechanism constrains policy updates within a trust region to effectively stabilize training. Incorporating this PPO-based clipping yields the final update rule presented in recursions~(\ref{eq:happo0}), which serves as the foundation for joint policy optimization procedure in Algorithm~\ref{alg:marl}. In the case of $N_B=1$, Lines 10 and 12 provide the explicit computation of the gradient estimator $G_z$.

To compute the global advantage $A(s_{t-1}, a_t; \theta)$ efficiently, we adopt the GAE method \citep{schulman2015high}, which combines temporal-difference (TD) residuals using an exponentially weighted average. Specifically, the advantage is estimated as
$$
\hat{A}(s_{t-1}, a_t) := \sum_{l=1}^\infty (\iota \lambda)^l \delta_{t+l},
$$
where $\delta_t := r_t + \iota v(s_{t}; \phi) - v(s_{t-1}; \phi)$ denotes the one-step TD error, and $v(s; \phi)$ is the critic that approximates the state-value function $V(s; \theta)$. The hyperparameter $\lambda \in [0,1]$ controls the bias-variance trade-off, interpolating between the high-bias, low-variance estimate of TD(0) and the low-bias, high-variance estimate of the Monte Carlo return.
Finally, to train the critic $v(s; \phi)$, we define a bootstrapped target return for each state as
$$
\hat{R}(s_{t-1}) := \hat{A}(s_{t-1}, a_t) + v(s_{t-1}; \phi),
$$
which serves as the regression target for minimizing the squared loss between the predicted and target values. The complete pseudocode of our proposed multi-timescale multi-agent RL algorithm is presented in Algorithm~\ref{alg:marl}.

\begin{algorithm}[htbp]
    \caption{Multi-timescale Multi-agent Reinforcement Learning}
    \label{alg:marl}
\begin{algorithmic}[1]
    \STATE \textbf{Input:} \# of iterations $N_I$, \# of mini batches $N_B$, batch size $B$, step sizes $\{\varepsilon_0[n],\varepsilon_1[n],\varepsilon_2[n]\}_{n=0}^{N_I-1}$, policy deviation tolerance $\epsilon$
    \STATE \textbf{Initialize:} Actor/agent policy parameters $\theta[0]=(\theta_q[0],\theta_\alpha[0])$, critic parameter $\phi[0]$, interaction history buffer $\mathcal{H}$
    \FOR{$n = 0$ to $N_I-1$}
        \STATE Collect interaction data with joint policy $\pi(\cdot|\cdot;\theta[n])$ and store by transitions $(s,a,s')$ into $\mathcal{H}$
        \STATE Draw a random shuffle $((z,\varepsilon_z[n]), (z',\varepsilon_{z'}[n]))$ of $((q, \varepsilon_1[n]), (\alpha, \varepsilon_2[n]))$
        \FOR{$m = 0$ to $N_B-1$}
        \STATE Set $\hat{\theta}[0]=\theta[n]$ and $\hat{\phi}[0]=\phi[n]$
        \STATE Sample a random batch of $B$ transitions $\{(s_b,a_b,s'_b)\}_{b=1}^B$ from $\mathcal{H}$
        \STATE Compute the advantage $\hat{A}_{z,b}=\hat{A}(s_b,a_b)$ based on critic $v(\cdot;\hat\phi[m])$ with GAE
        \STATE Update agent $\hat\theta_z[m]$ to $\hat\theta_{z}[m+1]$ with step size $\varepsilon_z[n]$ following the ascent direction
        $$\frac{1}{B}\sum_{b=1}^B \nabla_{\theta_z}\min\bigg(\frac{\pi_z(a_{z,b}|s_b;\theta_z) }{\pi_z(a_{z,b}|s_b;\theta_z[n])}\hat{A}_{z,b},\ \Pi_{[1-\epsilon,1+\epsilon] }\big(\frac{\pi_z(a_{z,b}|s_b;\theta_z)}{\pi_z(a_{z,b}|s_b;\theta_z[n])}\big)\hat{A}_{z,b}\bigg)\bigg|_{\theta_z=\hat{\theta}_{z}[m]}$$
        \STATE Compute the reweighted advantage $\hat{A}_{z',b}=\frac{\pi_z(a_{z,b}|s_b;\hat\theta_z[m+1]) }{\pi_z(a_{z,b}|s_b;\theta_z[n])}\hat{A}_{z,b}$
        \STATE Update agent $\hat\theta_{z'}[m]$ to $\hat\theta_{z'}[m+1]$  with step size $\varepsilon_{z'}[n]$ following the ascent direction
        $$\frac{1}{B}\sum_{b=1}^B \nabla_{\theta_{z'}}\min\bigg(\frac{\pi_{z'}(a_{z',b}|s_b;\theta_{z'}) }{\pi_{z'}(a_{z',b}|s_b;\theta_{z'}[n])}\hat{A}_{z',b},\ \Pi_{[1-\epsilon,1+\epsilon] }\big(\frac{\pi_{z'}(a_{z',b}|s_b;\theta_{z'})}{\pi_{z'}(a_{z',b}|s_b;\theta_{z'}[n])}\big)\hat{A}_{{z'},b}\bigg)\bigg|_{\theta_{z'}=\hat{\theta}_{z'}[m]}$$
        \STATE Update critic $\hat\phi[m]$ to $\hat\phi[m+1]$  with step size $\varepsilon_{\phi}[n]$ following the descent direction
        $$\frac{1}{B}\sum_{b=1}^B 2\big( v(s_b;\phi) - \hat{R}(s_b)\big)\nabla_{\theta_{\phi}}v(s_b;\phi)\bigg|_{\phi=\hat\phi[m]}$$
        \ENDFOR
        \STATE Set $\theta[n+1]=\hat\theta[N_B]$ and $\phi[n+1]=\hat\phi[N_B]$
    \ENDFOR

    \STATE \textbf{Output:} Trained agent policy $\theta[N_I]=(\theta_q[N_I],\theta_\alpha[N_I])$
\end{algorithmic}
\end{algorithm}

\subsection{Asymptotic Analysis in Section \ref{subsec4.3:multi-timescale_multi_agent_RL}}\label{appendix4.3theorem}

Before conducting the asymptotic analysis, we introduce two technical assumptions that regulate the behavior of neural network parameters under the multi-timescale SA dynamics.

\begin{assumption}\label{assumption:Lip2}
The gradient estimators are Lipschitz continuous with respect to the neural network parameters being updated, i.e., there exists a constant $ L' > 0 $ such that for all $ \theta, \theta_1, \theta_2 \in \Theta $, $ \varphi \in \Omega $, and $ z \in \{q, \alpha\} $, it holds that $ \| G_z(\theta, \theta_{1,z'}, \varphi) - G_z(\theta, \theta_{2,z'}, \varphi) \| \leq L' \| \theta_{1,z'} - \theta_{2,z'} \| $.
\end{assumption}

\begin{assumption}\label{assumption:Lip3}
    The gradients of the objective function with respect to neural network parameters are Lipschitz continuous. That is, there exists a constant $L''>0$ such that for all $\theta_i\in \Theta$, $i=1,2$, and $z\in\{q,\alpha\}$, $\Vert \nabla_{\theta_z} \mathcal{L}_z(\theta_1,\theta_z,\theta_{1,z'})\big|_{\theta_z=\theta_{1,z}}-\nabla_{\theta_z} \mathcal{L}_z(\theta_2,\theta_z,\theta_{2,z'})\big|_{\theta_z=\theta_{2,z}} \Vert \leq L'' \Vert \theta_1 - \theta_2 \Vert $.
\end{assumption}

Assumption \ref{assumption:Lip2} requires that the gradient estimators be Lipschitz continuous with respect to the network parameters currently being updated. This condition is critical for controlling additional error terms arising from the sequential update scheme, where different network components are updated in an alternating rather than fully synchronized manner. It ensures that the gradient estimates remain stable with respect to stale or lagged parameters of other modules.
Assumption~\ref{assumption:Lip3} imposes a standard Lipschitz continuity assumption on gradients, analogous to Assumption~\ref{assumption:Lip}, but is adapted to the parameterized neural network setting.

These assumptions are mild in practice and are often implicitly satisfied in standard deep RL models, particularly when smooth activation functions are used and the parameter space is regularized via projection, clipping, normalization, or other common techniques. Under such configurations, both assumptions naturally hold without requiring restrictive architectural constraints.

\proof{Proof of Theorem \ref{marlconv1}}
\textbf{We first establish the convergence of $\{\theta[n]\}$ generated by recursions (\ref{eq:drl1})–(\ref{eq:drl2}).}
Assume the sequence ${\theta[n]}$ lies within the compact convex region $\Theta$. In this case, the projection operator can be omitted; if not, it can still be handled in the same way as in Lemma \ref{mtsconv1}. We can then rewrite the recursions (\ref{eq:drl1})–(\ref{eq:drl2}) as follows:
\begin{align}
    \theta_q[n+1] &= \theta_q[n]+\varepsilon_1[n]\ \nabla_{\theta_q}\mathcal{L}_q(\theta[n],\theta_q,\theta_\alpha[n])\big|_{\theta_q=\theta_q[n]} + \varepsilon_1[n]\ \mathcal{M}_q[n] + \varepsilon_1[n]\ \mathcal{B}_q[n], \label{eq:upd1}
    \\
    \theta_{\alpha}[n+1] &= \theta_{\alpha}[n]+\varepsilon_2[n]\ \nabla_{\theta_\alpha}\mathcal{L}_\alpha(\theta[n],\theta_\alpha,\theta_q[n])\big|_{\theta_\alpha=\theta_\alpha[n]} + \varepsilon_2[n]\ \mathcal{M}_\alpha[n] + \varepsilon_2[n]\ \mathcal{B}_\alpha[n], \label{eq:upd2}
\end{align}
where we denote sequential update errors as $\mathcal{B}_q[n]= (1-\sigma[n])\big(G_q(\theta[n], \theta_\alpha[n+1],\varphi[n]) - G_q(\theta[n],\theta_{\alpha}[n],\varphi[n])\big)$ and $\mathcal{B}_\alpha[n]= \sigma[n]\big(G_\alpha(\theta[n], \theta_q[n+1],\varphi[n]) - G_\alpha(\theta[n],\theta_{q}[n],\varphi[n])\big)$.
We have
\begin{align*}
    \Vert \mathcal{B}_q[n] \Vert
    &\leq (1-\sigma[n])\Vert G_q(\theta[n], \theta_\alpha[n+1],\varphi[n]) - G_q(\theta[n],\theta_{\alpha}[n],\varphi[n])\Vert\\
    &\leq (1-\sigma[n])L'\Vert \theta_\alpha[n+1]-\theta_\alpha[n]\Vert \leq L'\Vert \varepsilon_2[n] G_\alpha(\theta[n], \theta_q[n],\varphi[n])\Vert\\
    &\leq L'\varepsilon_2[n] (\Vert \ \nabla_{\theta_\alpha}\mathcal{L}_q(\theta[n],\theta_\alpha,\theta_q[n])\big|_{\theta_\alpha=\theta_\alpha[n]}\Vert + \Vert \mathcal{M}_\alpha[n]\Vert),
\end{align*}
where the second inequality follows from Assumption \ref{assumption:Lip2}, and the third one is obtained after excluding the case $\sigma[n]=1$. Since $\nabla_{\theta_q}\mathcal{L}_q(\theta,\theta_\alpha',\theta_q)\big|_{\theta'_\alpha=\theta_\alpha}$ is Lipschitz continuous on bounded regions and $\Vert \mathcal{M}_\alpha[n]\Vert$ is bounded by Assumption \ref{assumption:asymunbiased}, $\mathcal{B}_q[n]\to0$ as $n\to\infty$ a.s. The same conclusion applies to the sequence $\{\mathcal{B}_\alpha[n]\}$. Under Assumption \ref{assumption:asymunbiased}, the gradient estimation error $\mathcal{M}_z[n]$ can be decomposed into a square-integrable martingale-difference noise term plus a vanishing bias term.
The above analysis shows that, apart from the first two terms on the right-hand side of recursions (\ref{eq:upd1}) and (\ref{eq:upd2}), all remaining terms can be treated as  negligible using the techniques in Lemma \ref{mtsconv1} without affecting the convergence, i.e., for fixed $\overline{\xi}>0$,
\begin{align}
    \lim_{n\to \infty} \sup_{\xi\in[0,\overline\xi]} \sum_{n'=n}^{\langle \xi[n]+\xi \rangle-1}\varepsilon_1[n']\mathcal{M}_q[n']=0 \text{ and } \lim_{n\to \infty} \sup_{\xi\in[0,\overline\xi]} \sum_{n'=n}^{\langle \xi[n]+\xi \rangle-1}\varepsilon_1[n']\mathcal{B}_q[n']=0, \text{ with probability one.}\label{eq:vanish}
\end{align}

Therefore, for each agent $z\in\{q,\alpha\}$, we can again define the piecewise constant interpolation $\theta_z^{[0]}(\cdot)$ on $\mathbb{R}$ by $\theta_z^{[0]}(\xi)=\theta_z[0]$ for $\xi<0$, and $\theta_z^{[0]}(\xi)=\theta_z[n]$ for $\xi\in\big[\xi[n], \xi[n+1]\big)$. Denote the shifted processes as $\theta_z^{[n]}(\xi)=\theta_z^{[0]}(\xi[n]+\xi)$. By the same argument as in Lemma \ref{mtsconv1}, we arrive at the conclusion that the limit $\theta(\cdot)$ of any convergent subsequence of $\{\theta^{[n]}(\cdot)\}$ satisfies 
$$
\theta_q(\xi)=\theta_q(0)+\int_0^\xi \nabla_{\theta_q}\mathcal{L}_q(\theta(\xi'),\theta_q,\theta_\alpha(\xi'))\big|_{\theta_q=\theta_q(\xi')} d\xi', \text{ and }\theta_\alpha(\xi)=\theta_\alpha(0). 
$$
Note that 
$\mathbb{E}_{\substack{s\sim\rho(\theta)\\ a\sim\pi(\theta)}}\left[ A(s,a;\theta) \nabla_{\theta_q} \log\pi_q(a_q|s;\theta_q) \right]\big|_{\theta=(x,\overline\theta_\alpha)} = \nabla_{\theta_q}\mathcal{L}_q((x,\overline\theta_\alpha),\theta_q,\overline\theta_\alpha)\big|_{\theta_q=x} $.
Therefore, almost surely, $\{\theta_\alpha[n]\}$ converges to $\overline\theta_\alpha\in\Theta_\alpha$, and
$\{\theta_q[n]\}$ converges to a compact connected, internally chain transitive invariant set $\mathcal{I}(\overline\theta_\alpha)\subset\Theta_q$ of ODE (\ref{eq:marlode1}).

\textbf{Next, we prove tightness of the occupation‐measure sequence.}
The convergence of sequences generated by recursions further implies that the sequence $\{\mu^{[n]}_\zeta\}$ is asymptotically tight in
$\mathcal P(\mathcal I(\overline\theta_\alpha))$ for any $\zeta\in[0,\overline\zeta]$. Since $\mathcal I(\overline\theta_\alpha)$ is compact in a Polish space and $P(\mathcal I(\overline\theta_\alpha))$ is compact under the weak topology, the product space $\mathcal{M}_\mathcal{I}:= \mathcal P(\mathcal I(\overline\theta_\alpha))^{[0,\overline\zeta]}$ is also compact by Tychonoff's Theorem \citep{kelley2017general_appendix}. Thus, the family $\{\mu^{[n]}_\cdot\}$ is tight in $\mathcal{M}_\mathcal{I}$. By Prokhorov’s Theorem \citep{billingsley2013convergence_appendix}, there exists a subsequence (again indexed by $n$ for notation simplicity) such that almost surely,
$\mu_\cdot^{[n]}\to\mu^*_\cdot\in \mathcal{M}_\mathcal{I}$, where $\mu^*_\cdot$ is a limiting measure-valued trajectory mapping $[0,\overline\zeta]$ to $\mathcal P(\mathcal I(\overline\theta_\alpha))$.

\textbf{Finally, we identify the limiting invariant measure of the occupation-measure sequence.}
For any test function $f\in C^{1}(\Theta_q)$, we have by Taylor expansion
\begin{align*}
    \lim_{n\to \infty}  f(\theta_q[\langle \zeta[n]&+\overline\zeta\rangle'])-f(\theta_q[n])=  \lim_{n\to \infty}   \sum_{n'=n}^{\langle \zeta[n]+\overline\zeta\rangle'-1} \bigg(\varepsilon_2[n'] \nabla_{\theta_q}^\top f(\theta_q[n'])  (\theta[n'+1]-\theta[n'])  +O(\varepsilon_2^2[n'])\bigg) \\
    =& \lim_{n\to \infty}  \bigg(\sum_{n'=n}^{\langle \zeta[n]+\overline\zeta\rangle'-1} \varepsilon_2[n'] \nabla_{\theta_q}^\top f(\theta_q[n'])  \nabla_{\theta_q}\mathcal{L}_q(\theta[n'],\theta_q,\theta_\alpha[n'])\big|_{\theta_q=\theta_q[n']}  \\
    & + \sum_{n'=n}^{\langle \zeta[n]+\overline\zeta\rangle'-1} \varepsilon_2[n'] \nabla_{\theta_q}^\top f(\theta_q[n']) \big(\mathcal{M}_q[n']+\mathcal{B}_q[n']\big) +O(\sum_{n'=n}^{\langle \zeta[n]+\overline\zeta\rangle'-1} \varepsilon_2^2[n'])\bigg)\\
    = &\lim_{n\to \infty}   \sum_{n'=n}^{\langle \zeta[n]+\overline\zeta\rangle'-1} \varepsilon_2[n'] \nabla_{\theta_q}^\top f(\theta_q[n'])  \nabla_{\theta_q}\mathcal{L}_q(\theta[n'],\theta_q,\theta_\alpha[n'])\big|_{\theta_q=\theta_q[n']}, \text{ a.s.,}
\end{align*}
where the terms in the third line all vanish mainly by Assumption \ref{assumption:stepsize} in the same manner of deriving (\ref{eq:vanish}), but with an extra uniformly bounded term $\nabla_{\theta_q} f(\theta_q[n'])$.
Note that $ \lim_{n\to \infty}  f(\theta_q[n'])-f(\theta_q[n])=0$ a.s. by the convergence of $\{\theta_q[n]\}$.
Replacing $\theta_\alpha[n']$ with the limit $\overline\theta_\alpha$, we obtain
\begin{equation*}
\resizebox{\linewidth}{!}{$%
\begin{aligned}
    \lim_{n\to \infty}&  f(\theta_q[\langle \zeta[n]+\overline\zeta\rangle'])-f(\theta_q[n]) = \lim_{n\to \infty} \int_0^{\overline\zeta} \int_{\Theta_q} \nabla_{\theta_q}^\top f(x)  \nabla_{\theta_q}\mathcal{L}_q((x,\overline\theta_\alpha),\theta_q,\overline\theta_\alpha)\big|_{\theta_q=x} d\mu^{[n]}_\zeta(x) d\zeta + \lim_{n\to \infty} \rho_{\overline\zeta}[n] \\
    &+\lim_{n\to \infty}   \sum_{n'=n}^{\langle \zeta[n]+\overline\zeta\rangle'-1} \varepsilon_2[n'] \nabla_{\theta_q}^\top f(\theta_q[n']) \big( \nabla_{\theta_q}\mathcal{L}_q(\theta[n'],\theta_q,\theta_\alpha[n'])\big|_{\theta_q=\theta_q[n']} - \nabla_{\theta_q}\mathcal{L}_q((\theta_q[n'],\overline\theta_\alpha),\theta_q,\overline\theta_\alpha)\big|_{\theta_q=\theta_q[n']}\big)\\
    =& \int_0^{\overline\zeta} \int_{\Theta_q} \nabla_{\theta_q}^\top f(x)  \nabla_{\theta_q}\mathcal{L}_q((x,\overline\theta_\alpha),\theta_q,\overline\theta_\alpha)\big|_{\theta_q=x} d\mu^{*}_\zeta(x) d\zeta = 0, \text{ with probability one,}
\end{aligned}
$}%
\end{equation*}
where $\rho_{\overline\zeta}[n]=O(\varepsilon_2[n])$ arises from replacing the discrete sum with an integral, and the last two terms on the right-hand side of the first equality vanish due to the convergence of $\{\theta_\alpha[n]\}$ and Assumptions \ref{assumption:stepsize} and \ref{assumption:Lip3}.
Since $\overline\zeta>0$ is chosen arbitrarily, we obtain the Echeverría–Khasminskiĭ Equation \citep{ethier2009markov_appendix}, i.e.,
$\int_{\Theta_q} \nabla_{\theta_q}^\top f(x)  \nabla_{\theta_q}\mathcal{L}_q((x,\overline\theta_\alpha),\theta_q,\overline\theta_\alpha)\big|_{\theta_q=x} d\mu^{*}_\zeta(x)=0$,
for any $f\in C^{1}(\Theta_q)$ and $\zeta\in[0,\overline\zeta]$, which implies that $\mu^{*}_\zeta\in J(\overline\theta_\alpha)$. Furthermore, if $J(\overline\theta_\alpha)$ is a singleton, then the almost sure convergence extends to the full sequence $\{\mu^{[n]}_\cdot \}$. 
\Halmos
\endproof

\proof{Proof of Theorem \ref{marlconv2}.}
We again replace the projection operation with a boundedness assumption of the recursions within $\Theta$ to avoid the redundant description throughout the proof. And note that $\nabla_{\tilde\theta_\alpha}\mathcal{L}_\alpha(\theta, \tilde\theta_\alpha, \theta_q)\vert_{\tilde\theta_\alpha=\theta_\alpha} = \mathbb{E}_{\substack{{s\sim\rho( \theta )}\\{a\sim\pi(\theta )} }}\big[A(s,a;\theta) \nabla_{\theta_\alpha} \log\pi_\alpha(a_\alpha|s;\theta_\alpha) \big]$.
For given $\overline\zeta>0$, we define the windowed ODE for each time offset $\zeta\geq0$ by
$$\theta_\alpha^\zeta(t) = \tilde{\theta}_\alpha(\zeta)+\int_\zeta^{t}\int_{\Theta_q}    \nabla_{\theta_\alpha}\mathcal{L}_\alpha((x,\theta_\alpha^\zeta(t')), \theta_\alpha, x)\big|_{\theta_\alpha=\theta_\alpha^\zeta(t')} d\mu^{[\langle \zeta\rangle']}_{t'-\zeta}(x) dt', \text{ for } t\in[\zeta, \zeta+\overline\zeta].$$
Recall the definition of $\tilde\theta_\alpha(\cdot)$, we have $\tilde\theta_\alpha(t)=\theta_\alpha[\langle t \rangle']$. Thus, we have from recursion (\ref{eq:upd2}) that
\begin{align*}
    \tilde\theta_\alpha(t) - \theta_\alpha^{\zeta}(t) &= \theta_\alpha[\langle t \rangle'] -\theta_\alpha[\langle \zeta \rangle'] - \int_\zeta^{t}\int_{\Theta_q}    \nabla_{\theta_\alpha}\mathcal{L}_\alpha((x,\theta_\alpha^\zeta(t')), \theta_\alpha, x)\big|_{\theta_\alpha=\theta_\alpha^\zeta(t')} d\mu^{[\langle \zeta\rangle']}_{t'-\zeta}(x) dt'\\
    &= \sum_{n=\langle \zeta \rangle'}^{\langle t \rangle'-1} \varepsilon_2[n'] \big( \mathcal{M}_\alpha[n] + \mathcal{B}_\alpha[n]\big) + \tilde\rho_{\zeta,t},
\end{align*}
for $t\in[\zeta,\zeta+\overline\zeta]$, where $\tilde\rho_{\zeta,t}=O(\varepsilon_2[\langle \zeta\rangle'])$ denotes the error from replacing the discrete sum with an integral, which shrinks to zero as $\zeta\to\infty$.
Similar to (\ref{eq:vanish}), the second-to-last term in the second line vanishes as $\zeta\to\infty$. Thus, we have 
$\lim_{\zeta\to\infty} \sup_{t\in[\zeta,\zeta+\overline{\zeta}]} \Vert \tilde\theta_\alpha(t) - \theta_\alpha^{\zeta}(t) \Vert = 0$, a.s.
In other words, the interpolation tracks the windowed ODE arbitrarily well as $\zeta\to\infty$.

It can be verified from the proof of Theorem~\ref{marlconv1} that $J(\theta_\alpha)$ is nonempty, compact and convex, and upper–semicontinuous for $\theta_\alpha\in\Theta_\alpha$. Hence, $H(\theta_\alpha)$ also takes nonempty compact convex values and is upper–semicontinuous (i.e. it has closed graph).
Along any convergent subsequence in $\zeta$, the limiting occupation measure of $\{\mu^{[\langle\zeta\rangle']}_{t-\zeta}\}$ lies in $J(\theta_\alpha(t))$ and $H(\theta_\alpha(t))$ is defined as the set of all such invariant-measure averages with closed graph and continuous dependence, and we further obtain
$\lim_{\zeta\to\infty} \sup_{t\in[\zeta,\zeta+\overline{\zeta}]} \text{dist}( \tilde\theta_\alpha(t), H(\theta_\alpha(t)))  = 0$, a.s.
Therefore, it follows Theorem 4.3 in \cite{benaim2005stochastic_appendix} that every limit point of the shifted path
$\tilde\theta_\alpha(\zeta+\cdot)$ in
$C(\mathbb{R}; \Theta_\alpha)$ is an absolutely continuous
solution of the inclusion $\dot\theta_\alpha(\zeta)\;\in\;
\tilde\Pi_{\Theta_\alpha}\!\bigl(H(\theta_\alpha(\zeta))\bigr)$,
and its trajectory lies in a compact, internally chain-transitive, and invariant set of this inclusion.
\Halmos
\endproof

In practical implementations of PPO, where multiple updates are performed per sample, the stochastic noise in the gradient estimator $G_z$ is no longer independent and instead exhibits Markovian dependence. For a rigorous treatment of such noise, we refer the reader to \citet{borkar2024stochastic}, where an auxiliary occupation measure is introduced to model the Markovian structure. This dependence is systematically accounted for and shown to average out in the limiting analysis.

\section{Supplementary Materials for Section \ref{sec:Numerical Experiments}}
\label{Supplementary Materials to Numerical Experiments}

This section presents supplementary analyses that extend the simulation results in Section~\ref{sec:Numerical Experiments}. Beyond the primary performance metrics, we begin by examining the behavioral characteristics of the trained RL agents through visualizations of decision correlations under varying coordination regimes. We then evaluate the robustness of our algorithmic design by systematically varying learning rates, decay schedules, and network architectures, thereby underscoring the stability conferred by the multi-timescale learning mechanism. Lastly, we assess the algorithm’s generalizability under alternative modeling assumptions, including changes to demand distributions, cost structures, system scales, and inventory fulfillment policies. Collectively, these extended experiments provide additional support for the effectiveness and adaptability of the proposed approach across diverse operational settings.

\subsection{Visualization of RL Agents' Behavioral Patterns}\label{appendix5.1}

To complement the quantitative results in Section~\ref{subsec5.2}, we present a series of correlation-based visualizations that shed further light on agent behavior under different coordination regimes. In particular, we analyze the behavior of agents trained under the four coordination configurations evaluated in Section~\ref{subsec5.3}: (i)  cooperative, (ii) fully isolated, (iii) isolated replenishment, and (iv) isolated recommendation. For each coordination setting, we conduct 20 independent training experiments and evaluate the performance of the resulting RL agents over 100 episodes each, yielding a total of 200,000 interaction steps across all scenarios.

\begin{figure}[ht!]
\centering 
\subfigure[Cooperative channels.]{
\label{fig:cov1}
\includegraphics[trim=0cm 0.5cm 0cm 0.5cm, width=0.9\linewidth]{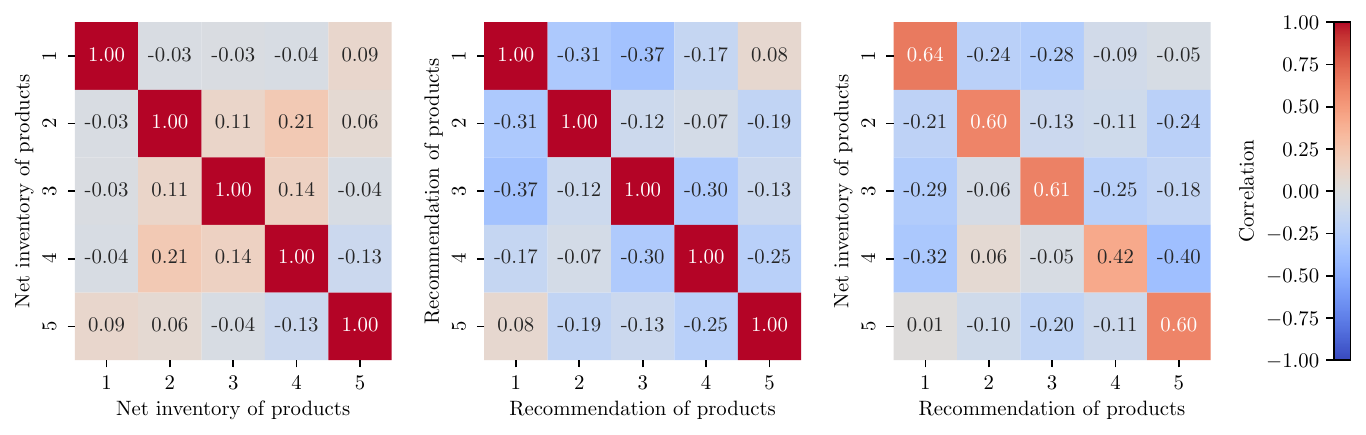}}
\subfigure[Fully isolated channels.]{
\label{fig:cov2}
\includegraphics[trim=0cm 0.5cm 0cm 0.5cm, width=0.9\linewidth]{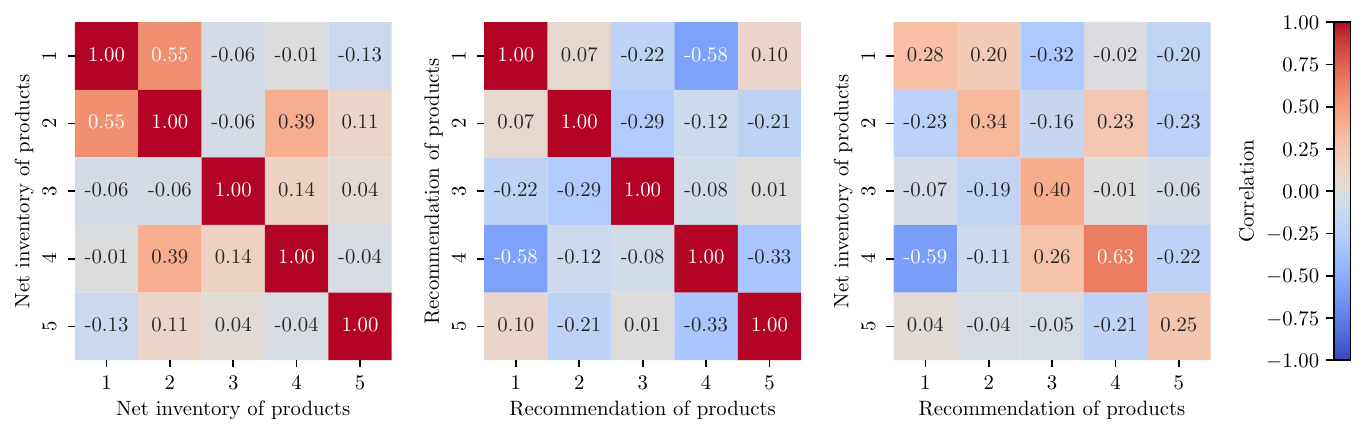}}
\subfigure[Isolated replenishment.]{
\label{fig:cov3}
\includegraphics[trim=0cm 0.5cm 0cm 0.5cm, width=0.9\linewidth]{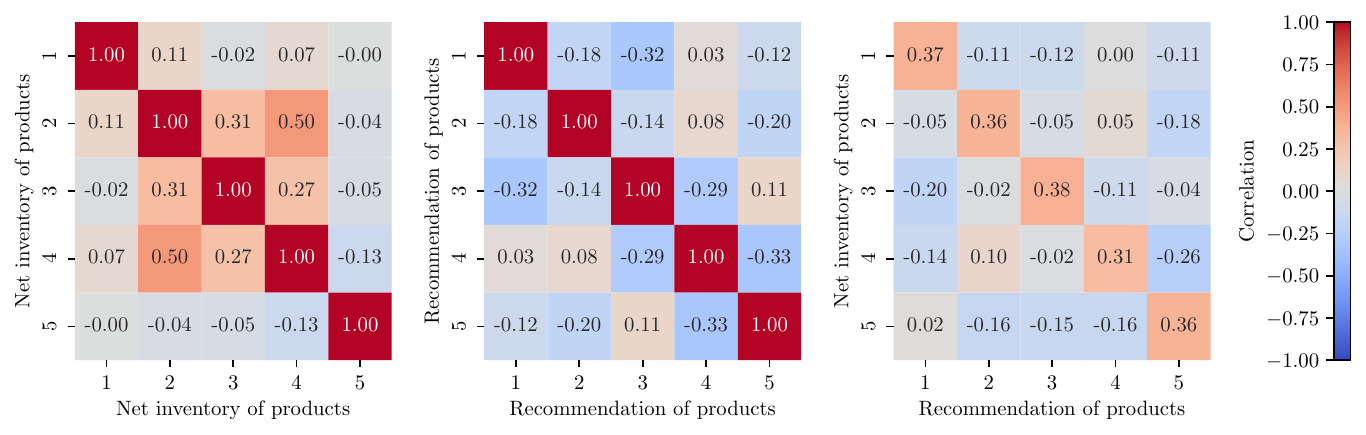}}
\subfigure[Isolated recommendation.]{
\label{fig:cov4}
\includegraphics[trim=0cm 0.5cm 0cm 0.5cm, width=0.9\linewidth]{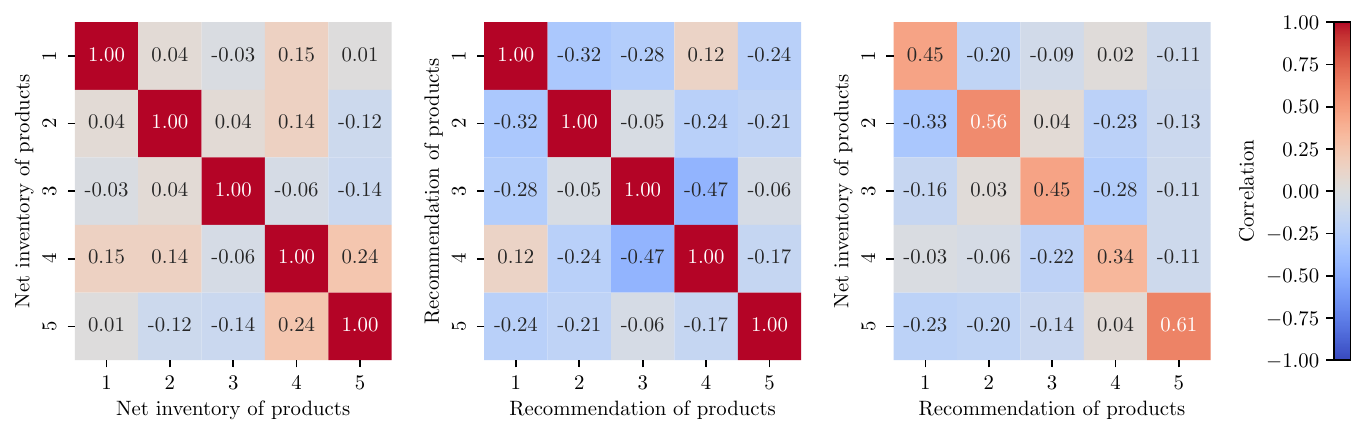}}
\caption{Correlation structure between net inventory levels and recommendation intensities. In each panel, the left figure displays pairwise correlations within inventory variables, the middle figure shows correlations within recommendation intensity variables, and the right figure illustrates cross-correlations between inventory and recommendation intensity.}
\label{fig:cov}
\end{figure}

We focus on two key behavioral variables: net inventory levels and aggregated recommendation intensities. For each setting, we compute pairwise correlation matrices among these variables. To facilitate interpretation, and taking advantage of the matrices' symmetry, Figure~\ref{fig:cov} presents heatmaps 
of the within- and between-domain correlations. Each panel corresponds to one of the four coordination mechanisms. These visualizations provide insight into how different coordination architectures shape statistical dependencies within and across functional decision domains, highlighting emergent patterns of alignment, independence, or interference in multi-agent learning dynamics.

We begin by examining Figure~\ref{fig:cov}\subref{fig:cov1}, which reveals that correlations among inventory variables are generally much weaker than those among recommendation intensities, with most inventory correlations clustering near zero. This observation aligns with the design intuition behind our algorithm: inventory decisions are typically more localized and independent, whereas recommendation strategies are inherently coupled due to shared exposure constraints and competing product visibility. Notably, recommendation intensities across different products exhibit predominantly negative correlations, indicating a learned trade-off in attention allocation, i.e., increasing the recommendation for one product often coincides with a reduction in others.
In contrast, the positive correlations observed between inventory levels and their respective recommendation intensities highlight a strong degree of cross-functional synergy. Moreover, negative correlations between the inventory of one product and the recommendation intensity of another suggest that agents have internalized substitution effects and adjusted their behaviors accordingly.

Comparing Figure~\ref{fig:cov}\subref{fig:cov1} with~\subref{fig:cov2}, we observe that disabling cooperative training causes agents to concentrate more narrowly on their own functional domains---operations or marketing---resulting in stronger within-group correlations (inventory-inventory or recommendation-recommendation). Simultaneously, cross-functional correlations become more erratic and attenuated, indicating a breakdown in coordination between decision layers. Figures~\ref{fig:cov}\subref{fig:cov3} and~\subref{fig:cov4}, which correspond to partially isolated training scenarios, further reinforce these patterns. When one functional unit is excluded from joint training, it compensates by relying more heavily on internal adjustments to address external requirements. This shift is reflected in marked changes in the magnitude and structure of the relevant correlation coefficients.

 \clearpage

\begin{figure}[ht!]
    \centering
    \includegraphics[width=0.4\linewidth]{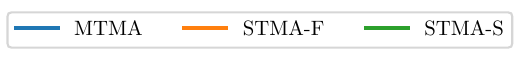}
    \includegraphics[width=0.95\linewidth]{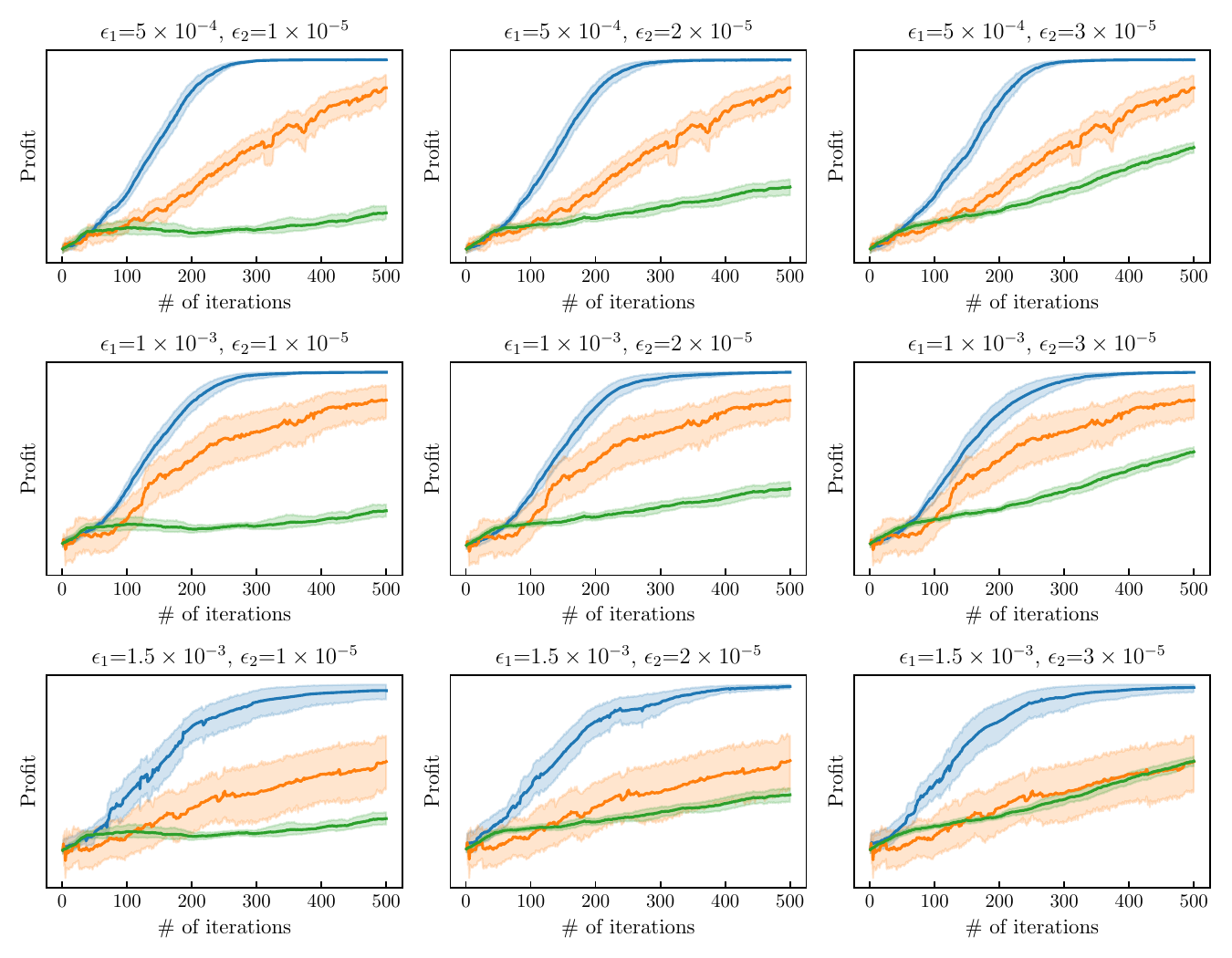}
    \caption{Learning curves of multi-timescale and single-timescale multi-agent reinforcement learning under varying initial learning rates, with decay exponents fixed at $(p_1, p_2)=(0.75, 0.99)$. All plotting configurations are consistent with those in Figure~\ref{fig:mtcompare}.}
    \label{fig:initlr_mt}
\end{figure}

\subsection{Robustness Evaluation on Algorithmic Configuration}\label{appendix5.2}

This subsection provides a detailed analysis of the robustness of the proposed multi-timescale multi-agent reinforcement learning algorithm by systematically varying key algorithmic parameters. The critic, responsible for state-value estimation, is updated using fixed parameters $(\epsilon_0, p_0) = (1\times10^{-3}, 0.51)$. All learning curves are averaged over 20 independent runs; in the presenting figures, solid lines indicate the mean, and shaded areas represent 95\% confidence intervals.

\begin{figure}[tb]
    \centering
    \includegraphics[width=0.4\linewidth]{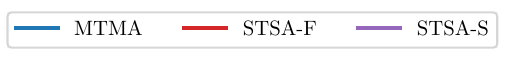}
    \includegraphics[width=0.95\linewidth]{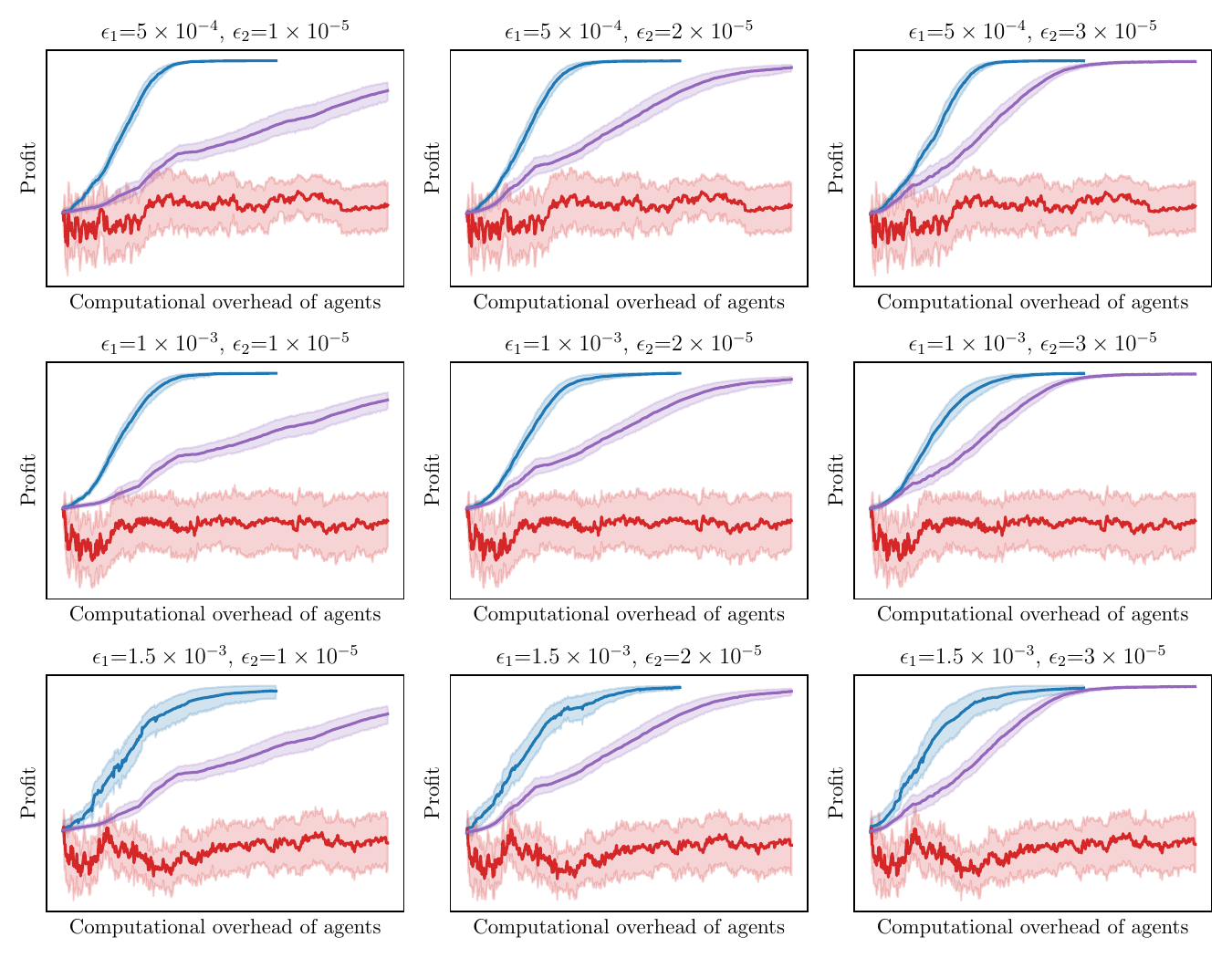}
    \caption{Learning curves of multi-agent and single-agent RL under varying initial learning rates, with the decay exponents fixed at $(p_1, p_2) = (0.75, 0.99)$. Plotting settings follow those in Figure~\ref{fig:macompare}.}
    \label{fig:initlr_ma}
\end{figure}

\textbf{Initial learning rate (step size).} We first investigate the impact of the initial learning rate on convergence speed and stability. In parallel to Figure \ref{fig:mtcompare}, Figures~\ref{fig:initlr_mt} and~\ref{fig:initlr_ma} present comparisons between multi- and single-timescale multi-agent RL, as well as between multi-agent and single-agent settings, under varying step sizes. The decay exponents are fixed at $(p_1, p_2)=(0.75, 0.99)$ throughout. we test performance under 50\% reductions and increases in the initial learning rate: $\epsilon_1 \in\{5\times10^{-4}, 1\times10^{-3}, 1.5\times10^{-3}\}$ and $ \epsilon_2  \in\{ 1\times10^{-5}, 2\times10^{-5}, 3\times10^{-5}\}$. Consistent with the observation in Section~\ref{subsec5.1}, our algorithm exhibits superior convergence speed and stability with minimal tuning overhead.

\begin{figure}[tb]
    \centering
    \includegraphics[width=0.4\linewidth]{legend_aver.pdf}
    \includegraphics[width=0.95\linewidth]{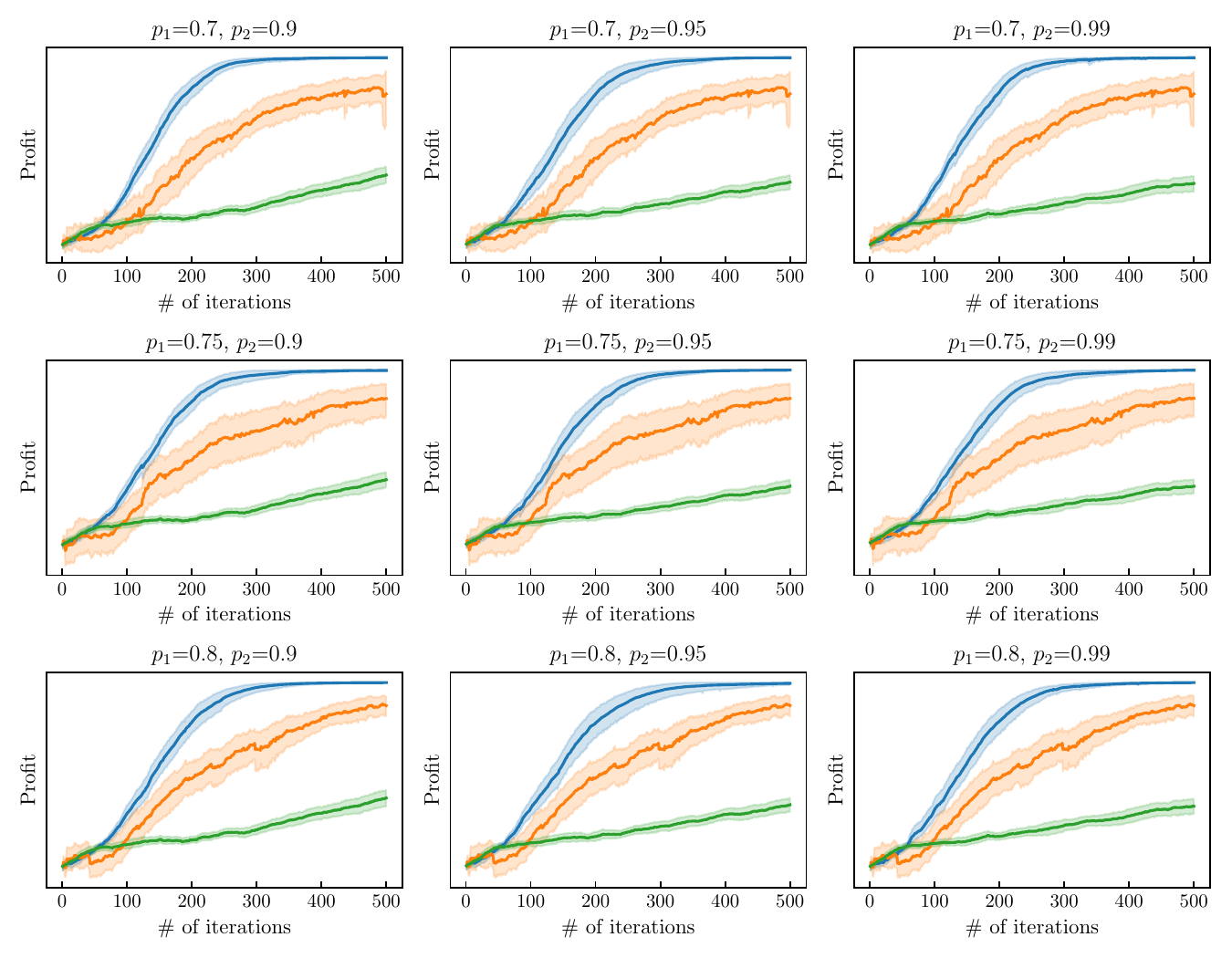}
    \caption{Learning curves of multi-timescale and single-timescale multi-agent RL under varying learning rate decay schedules, with initial learning rates fixed at $(\epsilon_1, \epsilon_2) = (1 \times 10^{-3},\ 2 \times 10^{-5})$. Plotting settings follow those in Figure~\ref{fig:macompare}.}
    \label{fig:decay_mt}
\end{figure}

\begin{figure}[tb]
    \centering
    \includegraphics[width=0.4\linewidth]{legend_aver_scaled.pdf}
    \includegraphics[width=0.95\linewidth]{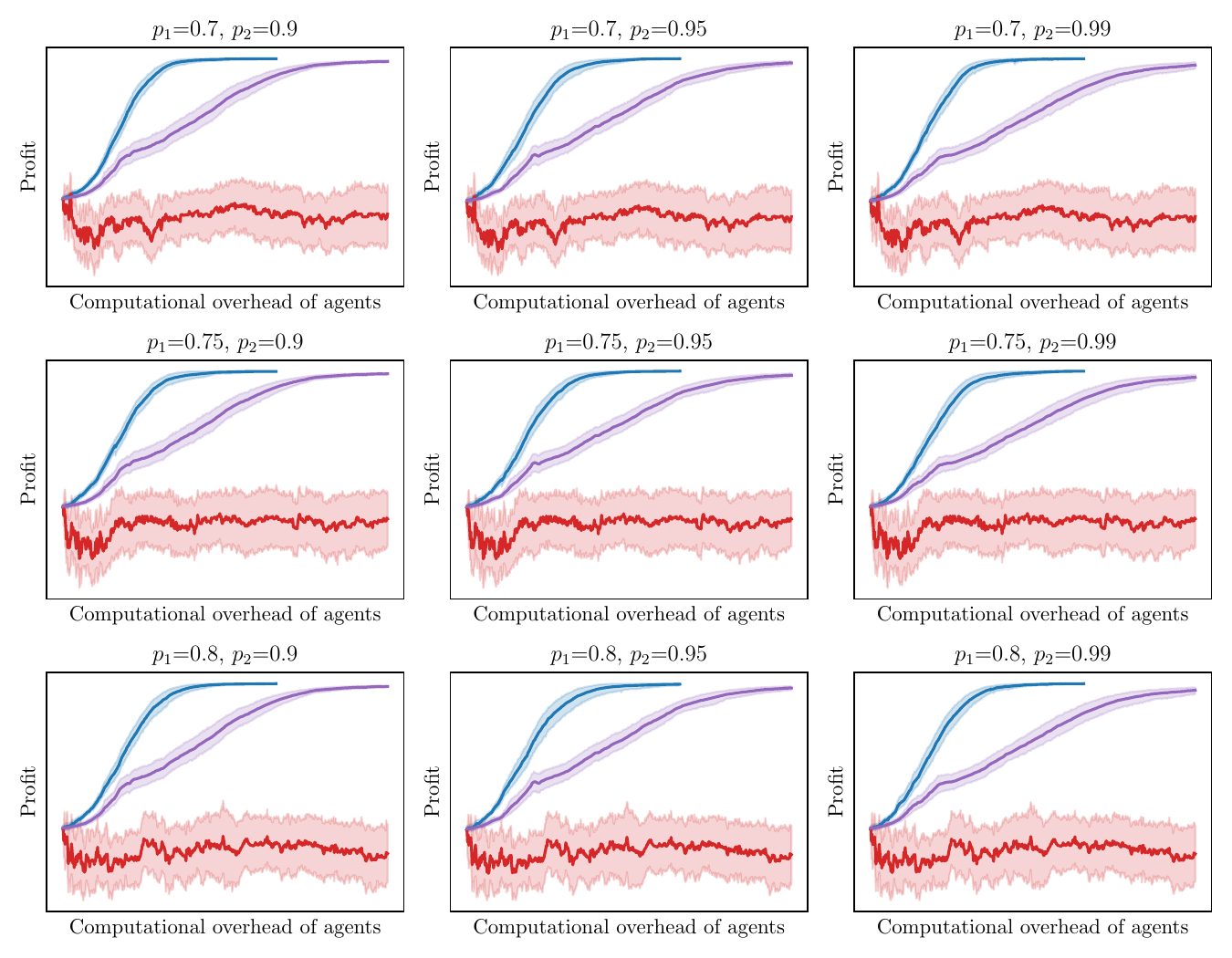}
    \caption{Learning curves of multi-agent and single-agent RL under varying learning rate decay schedules, with initial learning rates fixed at $(\epsilon_1, \epsilon_2) = (1 \times 10^{-3},\ 2 \times 10^{-5})$. Plotting settings follow those used in Figure~\ref{fig:macompare}.}
    \label{fig:decay_ma}
\end{figure}

\textbf{Learning rate decay exponent.} We next examine how variations in the learning rate decay schedule affect training performance. Figures~\ref{fig:decay_mt} and~\ref{fig:decay_ma} present learning curves under different decay exponents, with initial learning rates fixed at $(\epsilon_1, \epsilon_2)=(1\times10^{-3}, 2\times10^{-5})$. Consistent with Assumption~\ref{assumption:stepsize}, the decay exponents must not exceed 1, and the two learning rates must adhere to a prescribed timescale separation. Accordingly, we test $p_1 \in \{0.70, 0.75, 0.80\}$ and $p_2 \in \{0.90, 0.95, 0.99\}$. Across all configurations, the proposed algorithm maintains strong convergence speed and stability, reaffirming its robustness and low sensitivity to decay scheduling.

\begin{figure}[htb]
\centering 
\subfigure[Multi- vs. single-timescale multi-agent RL with half-width hidden layers.]{
\includegraphics[trim=0cm 0.5cm 0cm 0.2cm, width=0.4\linewidth]{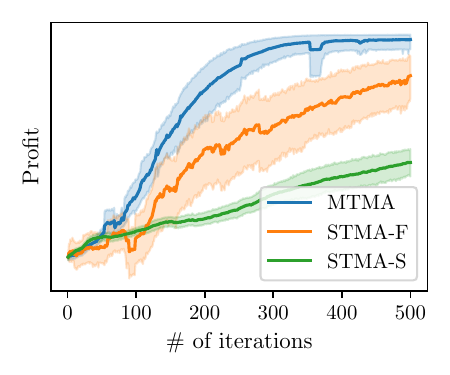}}
\hspace{1cm}
\subfigure[Multi- vs. single-agent RL with half-width hidden layers.]{
\includegraphics[trim=0cm 0.5cm 0cm 0.2cm, width=0.4\linewidth]{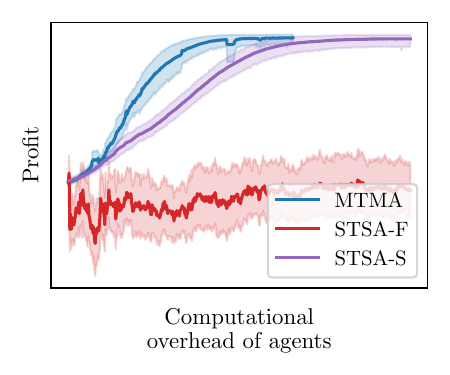}}\\
\subfigure[Multi- vs. single-timescale multi-agent RL with double-width hidden layers.]{
\includegraphics[trim=0cm 0.5cm 0cm 0.2cm, width=0.4\linewidth]{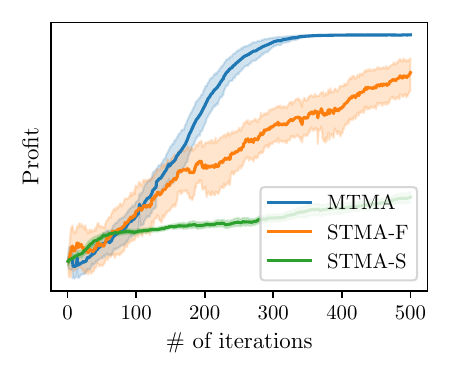}}
\hspace{1cm}
\subfigure[Multi- vs. single-agent RL with double-width hidden layers.]{
\includegraphics[trim=0cm 0.5cm 0cm 0.2cm, width=0.4\linewidth]{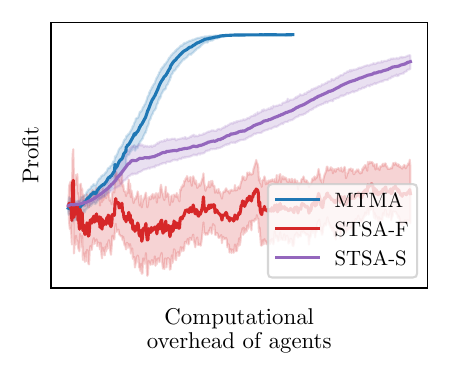}}
\caption{Learning curves of RL algorithms under varying neural network scales for agent policies. The initial learning rate is inversely scaled with network width---set higher for smaller networks and lower for larger ones. All plotting settings are consistent with those in Figure~\ref{fig:mtscompare}.}
\label{fig:netsize}
\end{figure}

\textbf{Neural network size.} Finally, we assess the sensitivity of learning performance to the size of neural network hidden layers used by both agents and the critic. Figure~\ref{fig:netsize} presents learning curves under two configurations: a half-width setting and a double-width setting. In the half-width configuration, the hidden layer dimensions are 64 for the inventory agent, 192 for the recommendation agent, and 256 for the critic. The double-width configuration scales these to 256, 768, and 1024, respectively. To account for the change in capacity, initial learning rates are inversely scaled with network width: $(\epsilon_1, \epsilon_2)=(2\times10^{-3}, 4\times10^{-5})$ for half-width, and $(5\times10^{-4}, 1\times10^{-5})$ for double-width networks. Across these configurations, the proposed algorithm exhibits stable and efficient convergence. Notably, these learning rates are not finely tuned but selected heuristically to enable controlled comparisons. Even when using more conservative defaults, e.g., $(\epsilon_1, \epsilon_2)=(1\times10^{-3}, 2\times10^{-5})$, the half-width setting maintains reliable convergence behavior.

Taken together, these findings highlight the effectiveness and robustness of our multi-timescale multi-agent RL algorithm. Specifically, the asymmetric agent system architecture and the separation of fast and slow update scales consistently deliver strong performance with minimal tuning effort.

\subsection{Robustness Evaluation of Modeling Assumptions}\label{appendix5.3}

This subsection examines the robustness and generalizability of the modeling framework introduced in Section~\ref{Theoretical Model} by testing its performance under a variety of structural and operational variations. The ablation studies serve two purposes: (i) to evaluate whether the proposed algorithm consistently delivers strong performance across diverse settings, and (ii) to assess the persistent value of agent coordination under these variations. We systematically investigate how key performance metrics---total profit, inventory cost, and marketing revenue---respond to changes in demand uncertainty, model parameters, system scale, and fulfillment policies. Each table reports average outcomes along with 95\% confidence intervals, computed over 20 independent simulation runs.

The main experiments in Section~\ref{sec:Numerical Experiments} assume customer purchasing behavior follows a Bernoulli distribution parameterized by softmax-transformed willingness scores. To evaluate generalizability, we extend the comparison of coordination mechanisms to alternative demand models, including exponential, multinomial, and Poisson distributions. For exponential demand, agents are allowed to make continuous replenishment decisions. All algorithmic hyperparameters are held the same with those used in Section \ref{subsec5.3}. Results summarized in Table~\ref{tab:5.3_ablation1} show that the cooperative setting consistently achieves the highest total profit and the lowest inventory cost across all demand types. These findings confirm that the proposed coordination framework remains effective under varied forms of demand uncertainty.

\begin{table}[htbp]
\centering
\caption{Performance comparison across coordination mechanisms under alternative customer demand models.}
\label{tab:5.3_ablation1}
\resizebox{\textwidth}{!}{  
\begin{tabular}{>{\centering\arraybackslash}m{4.5cm}||>{\centering\arraybackslash}m{3.5cm}>{\centering\arraybackslash}m{3.5cm}|>{\centering\arraybackslash}m{3.5cm}>{\centering\arraybackslash}m{3.5cm}} 
\toprule
Setting & \textbf{Cooperative} & Isolated & Isolated replenishment  & Isolated recommendation  \\
\midrule
\multicolumn{5}{c}{Exponential distribution} \\
\midrule
Aver. Total Profit &  \textbf{638.52}  &  451.71 &  416.50 & 467.99   \\
95\% Interval & \textbf{(617.04, 659.99)}   & (355.74,  547.68)  & (251.32, 581.69) & (336.01, 599.97)       \\
\midrule
Aver. Inventory Cost &  \textbf{237.04}  & 411.28  & 444.40  &  400.34  \\
95\% Interval & \textbf{(216.41, 257.66)}   & (322.99, 499.58)  & (291.44, 597.36) & (273.45, 527.22)       \\
\midrule
Aver. Marketing Revenue &  \textbf{875.55}  & 862.99  &  860.90 &  868.33  \\
95\% Interval & \textbf{(874.38, 876.73)}   & (854.99,  870.99)  & (848.42, 873.38) & (863.16, 873.49)       \\
\midrule
\multicolumn{5}{c}{Multinomial distribution} \\
\midrule
Aver. Total Profit &  \textbf{603.93}  & 361.08  &  284.33 &  480.93  \\
95\% Interval & \textbf{(593.27, 614.59)}   & (296.52, 425.64)  & (53.76, 514.90) & (443.85, 518.00)       \\
\midrule
Aver. Inventory Cost &  \textbf{270.30}  & 498.02  & 561.07  &  386.75  \\
95\% Interval & \textbf{(260.15, 280.44)}   & (437.66, 558.38)  & (356.10, 766.03) & (352.32, 421.19)       \\
\midrule
Aver. Marketing Revenue &  \textbf{874.23}  & 859.10  &  845.40 &  867.68  \\
95\% Interval & \textbf{(873.61, 874.84)}   & (854.35,  863.85)  & (819.46, 871.34) & (864.94, 870.42)       \\
\midrule
\multicolumn{5}{c}{Poisson distribution} \\
\midrule
Aver. Total Profit &  \textbf{514.97}  &  260.55 & 355.31  &  326.85  \\
95\% Interval & \textbf{(487.89, 542.05)}   & (147.16, 373.94)  & (268.59, 442.04) & (201.59, 452.11)       \\
\midrule
Aver. Inventory Cost &  \textbf{355.83}  & 591.55  & 500.63  &  534.13  \\
95\% Interval & \textbf{(330.12, 381.55)}   & (486.67, 696.43)  & (425.09, 576.17) & (414.92, 653.34)       \\
\midrule
Aver. Marketing Revenue &  \textbf{870.80}  &  852.10 &  855.94 &  860.98  \\
95\% Interval & \textbf{(868.79, 872.81)}   & (843.30, 860.90)  & (843.57, 868.31) & (854.74, 867.22)       \\
\bottomrule
\end{tabular}
}
\end{table}

\textbf{Cost parameters.} We next investigate the impact of unit cost structures on coordination outcomes. Noting that proportionally adjusting both inventory and recommendation costs yields an equivalent optimal solution, we focus on varying the unit recommendation cost while holding all other pricing parameters constant. This is mathematically equivalent to an inverse adjustment of unit inventory prices. The baseline cost used in Section \ref{subsec5.3} is $c = 0.025$; here, examine alternative values $c = 0.01$ and $c = 0.05$. As reported in Table~\ref{tab:5.3_ablation2}, the cooperative mechanism continues to consistently outperform all other coordination settings. Moreover, we observe that when recommendation becomes more cost-effective, the system increasingly leverages marketing to drive revenue, which in turn leads to higher inventory-related costs.

\begin{table}[htbp]
\centering
\caption{Performance comparison across coordination mechanisms under varying unit recommendation costs.}
\label{tab:5.3_ablation2}
\resizebox{\textwidth}{!}{  
\begin{tabular}{>{\centering\arraybackslash}m{4.5cm}||>{\centering\arraybackslash}m{3.5cm}>{\centering\arraybackslash}m{3.5cm}|>{\centering\arraybackslash}m{3.5cm}>{\centering\arraybackslash}m{3.5cm}} 
\toprule
Setting & \textbf{Cooperative} & Isolated & Isolated replenishment  & Isolated recommendation  \\
\midrule
\multicolumn{5}{c}{High recommendation cost ($c=0.05$)} \\
\midrule
Aver. Total Profit &  \textbf{481.85}  &  166.11 &  269.99 &  347.70  \\
95\% Interval & \textbf{(465.64, 498.05)}   & (65.64, 266.57)  & (183.40, 356.58) & (287.06, 408.35)       \\
\midrule
Aver. Inventory Cost &  \textbf{271.16}  & 565.29  &  467.77 &  396.78  \\
95\% Interval & \textbf{(255.46, 286.86)}   & (471.08, 659.50)  & (388.61, 546.92) & (338.92, 454.65)       \\
\midrule
Aver. Marketing Revenue &  \textbf{753.01}  & 731.40 &   737.75 &  744.48  \\
95\% Interval & \textbf{(751.69, 754.33)}   & (724.62, 738.18))  & (729.80, 745.71) & (741.34, 747.63)       \\
\midrule
\multicolumn{5}{c}{Low recommendation cost ($c=0.01$)} \\
\midrule
Aver. Total Profit &  \textbf{645.46}  & 232.38  & 433.73  &  530.53  \\
95\% Interval & \textbf{(627.27, 663.65)}   & (85.90, 378.86)  & (340.18, 527.28) & (496.98, 564.0)       \\
\midrule
Aver. Inventory Cost &  \textbf{301.90}  & 687.11  & 499.70  & 411.24   \\
95\% Interval & \textbf{(284.24, 319.57)}   & (552.64, 821.58)  & (412.95, 586.45) & (379.34, 443.14)       \\
\midrule
Aver. Marketing Revenue &  \textbf{947.36}  & 919.49  & 933.43  &  941.77 \\
95\% Interval & \textbf{(946.37, 948.35)}   & (906.48, 932.49)  & (926.17,  940.69) & (939.77, 943.78)       \\
\bottomrule
\end{tabular}
}
\end{table}

\textbf{System scale.} To assess the scalability of the proposed framework, we evaluate performance under expanded system configurations by altering the numbers of products and customers. Specifically, we consider two scenarios: (i) doubling the number of customers (\textit{N}=5, \textit{M}=40), and (ii) doubling the number of products (\textit{N}=10, \textit{M}=20). Given the resulting increase in the dimensionality of both the state and action spaces, we adopt a more conservative learning rate of $(\epsilon_1, \epsilon_2) = (5\times10^{-4}, 1\times10^{-5})$ and extend the training horizon in the product-doubling scenario to ensure convergence. Table~\ref{tab:5.3_ablation3} summarizes the outcomes. In both cases, the cooperative coordination mechanism continues to yield superior performance, highlighting the robustness and scalability of the proposed approach in more complex settings.

\begin{table}[htbp]
\centering
\caption{Performance comparison across coordination strategies under varying numbers of products and customers. (For the isolated replenishment setting with double the number of products, results reflect the average across runs after excluding two instances with diverging training behavior.)}
\label{tab:5.3_ablation3}
\resizebox{\textwidth}{!}{  
\begin{tabular}{>{\centering\arraybackslash}m{4.5cm}||>{\centering\arraybackslash}m{3.6cm}>{\centering\arraybackslash}m{3.5cm}|>{\centering\arraybackslash}m{3.5cm}>{\centering\arraybackslash}m{3.5cm}} 
\toprule
Setting & \textbf{Cooperative} & Isolated & Isolated replenishment  & Isolated recommendation  \\
\midrule
\multicolumn{5}{c}{5 products $\times$ 40 customers} \\
\midrule
Aver. Total Profit &  \textbf{1209.27}  & 569.11  & 403.28  &  835.82 \\
95\% Interval & \textbf{(1166.73, 1251.81)}   & (499.02, 639.20)  & (308.93, 497.64) & (774.37, 897.27)       \\
\midrule
Aver. Inventory Cost &  \textbf{423.97}  & 1136.37  & 1288.50  & 876.63  \\
95\% Interval & \textbf{(401.28, 446.65)}   & (1070.71, 1202.04)  & (1201.93, 1375.07) & (668.04, 800.82)       \\
\midrule
Aver. Marketing Revenue &  \textbf{1742.00}  & 1705.48  & 1691.78  &  1712.45 \\
95\% Interval & \textbf{(1735.63, 1748.38)}   & (1700.14, 1710.81)  & (1682.36, 1701.20) & (1702.87, 1722.02)       \\
\midrule
\multicolumn{5}{c}{10 products $\times$ 20 customers} \\
\midrule
Aver. Total Profit &  \textbf{461.96}  &  -435.87 & -$333.07^*$  & 160.61  \\
95\% Interval & \textbf{(444.34, 479.59)}   & (-1194.49, 322.75)  & (-735.95, 69.81) & (114.47, 206.75)       \\
\midrule
Aver. Inventory Cost &  \textbf{287.47}  & 1116.46  & 1014.32  & 572.68  \\
95\% Interval & \textbf{(271.96, 302.97)}   & (431.07, 1801.85)  & (655.46, 1373.19) & (531.74, 613.61)       \\
\midrule
Aver. Marketing Revenue &  \textbf{749.43}  &  680.59 &  681.25 & 733.29  \\
95\% Interval & \textbf{(747.02, 751.84)}   & (606.97, 754.21)  & (636.88, 725.63) & (727.83, 738.74)       \\
\bottomrule
\end{tabular}
}
\end{table}

\textbf{Fulfillment model.} While the main experiments in this paper assume a backlogging inventory model, we further evaluate the proposed framework under a lost-sales setting, which introduces more stringent constraints on inventory management. To ensure training stability under this setting, we apply a uniformly reduced learning rate across all components: $(\epsilon_0, \epsilon_1, \epsilon_2) = (5\times10^{-4}, 5\times10^{-4}, 1\times10^{-5})$, while keeping all other configurations unchanged. As reported in Table~\ref{tab:5.3_ablation4}, the cooperative coordination mechanism continues to deliver the best performance, confirming the robustness and adaptability of the proposed framework under alternative fulfillment models.

\begin{table}[htbp]
\centering
\caption{Performance comparison across agent coordination strategies under the lost-sales fulfillment setting.}
\label{tab:5.3_ablation4}
\resizebox{\textwidth}{!}{  
\begin{tabular}{>{\centering\arraybackslash}m{4.5cm}||>{\centering\arraybackslash}m{3.5cm}>{\centering\arraybackslash}m{3.5cm}|>{\centering\arraybackslash}m{3.5cm}>{\centering\arraybackslash}m{3.5cm}} 
\toprule
Setting & \textbf{Cooperative} & Isolated & Isolated replenishment  & Isolated recommendation  \\
\midrule
Aver. Total Profit &  \textbf{573.06}  &  394.17 & 352.72  &  370.82 \\
95\% Interval & \textbf{(556.50, 589.63)}   & (318.75, 469.59)  & (246.73, 458.70) & (323.69, 417.94)       \\
\midrule
Aver. Inventory Cost &  \textbf{232.85}  & 368.76  & 377.87  &  430.52 \\
95\% Interval & \textbf{(216.49, 249.22)}   & (294.19, 443.34)  & (262.33, 493.42) & (384.97, 476.06)       \\
\midrule
Aver. Marketing Revenue &  \textbf{805.92}  & 762.93  & 730.59  & 801.33  \\
95\% Interval & \textbf{(800.44, 811.39)}   & (751.62, 774.24)  & (712.41, 748.77) & (795.61, 807.06)       \\
\bottomrule
\end{tabular}
}
\end{table}

In summary, this comprehensive set of robustness evaluations highlights the key strengths of the proposed multi-timescale multi-agent reinforcement learning framework. The algorithm consistently generalizes well across a wide range of structural and operational variations, and it reliably benefits from coordinated decision-making. These findings underscore the robustness and adaptability of our approach in complex, uncertain operational environments.



\end{document}